\documentclass[letterpaper,11 pt]{article} 
\usepackage{algpseudocode}
\usepackage{algorithm}
\usepackage{mathtools}
\usepackage{amsthm}
\usepackage{amsmath,amsfonts,amssymb,color}
\usepackage{mathtools}
\usepackage{dsfont}
\usepackage{epsfig}
\usepackage{epstopdf}
\usepackage{caption}
\usepackage{subcaption}
\usepackage{dsfont}
\usepackage[dvipsnames]{xcolor}
\definecolor{mygreen}{rgb}{0.0, 0.5, 0.0}
\definecolor{winered}{rgb}{0.8,0,0}
\definecolor{myblue}{rgb}{0,0,0.8}
\usepackage{tikz}
\usepackage{empheq}

\usepackage{array}
\usepackage{multirow}
\usepackage{tikz}
\usepackage{relsize}
\usetikzlibrary{shapes,arrows}
\usepackage[hidelinks]{hyperref}
\hypersetup{
    colorlinks=true,
    linkcolor={winered},
    citecolor={mygreen}
}
\usepackage{cite}
\usepackage{tcolorbox}
\tcbuselibrary{theorems}
\DeclarePairedDelimiter\ceil{\lceil}{\rceil}

\DeclareMathOperator*{\argmax}{\arg\!\max}
\DeclareMathOperator*{\argmin}{\arg\!\min}
\DeclarePairedDelimiterX{\norm}[1]{\lVert}{\rVert}{#1}
\newcommand{\mc}{\mathcal}

\newtheorem{theorem}{Theorem}
\newtheorem{lemma}{Lemma}

\newtheorem{corollary}{Corollary}
\newtheorem{remark}{Remark}
\newtheorem{assumption}{Assumption}

\newtcbtheorem[number within=section]{mytheo}{Theorem}%
{colback=blue!5,colframe=blue!35!black,fonttitle=\bfseries}{theo}
\newtcbtheorem[number within=section]{mylem}{Lemma}%
{colback=blue!5,colframe=blue!35!black,fonttitle=\bfseries}{lem}

\usepackage[margin=1in]{geometry}
\title{\LARGE Collaborative Linear Bandits with Adversarial Agents: Near-Optimal Regret Bounds}
\author{Aritra Mitra*, Arman Adibi*, George J. Pappas, and Hamed Hassani
\thanks{The first two authors contributed equally. The authors are with the Department of Electrical and Systems Engineering, University of Pennsylvania. Email: {\tt \{amitra20, aadibi, hassani, pappasg\}@seas.upenn.edu}. This work was supported by NSF Award 1837253, NSF CAREER award CIF 1943064, and the Air Force Office
of Scientific Research Young Investigator Program (AFOSR-YIP) under award FA9550-20-1-0111.}}
\date{}
\begin{document}
\maketitle
\begin{abstract}
     We consider a linear stochastic bandit problem involving $M$ agents that can collaborate via a central server to minimize regret. A fraction $\alpha$ of these agents are adversarial and can act arbitrarily, leading to the following tension: while collaboration can potentially reduce regret, it can also disrupt the process of learning due to adversaries. In this work, we provide a fundamental understanding of this tension by designing new algorithms that balance the exploration-exploitation trade-off via carefully constructed robust confidence intervals. We also complement our algorithms with tight analyses. First, we develop a robust collaborative phased elimination algorithm that achieves $\tilde{O}\left(\alpha+ 1/\sqrt{M}\right) \sqrt{dT}$ regret for each good agent; here, $d$ is the model-dimension and $T$ is the horizon. For small $\alpha$, our result thus reveals a clear benefit of collaboration despite adversaries. Using an information-theoretic argument, we then prove a matching lower bound, thereby providing the first set of tight, near-optimal regret bounds for collaborative linear bandits with adversaries. Furthermore, by leveraging recent advances in high-dimensional robust statistics, we significantly extend our algorithmic ideas and results to (i) the generalized linear bandit model that allows for non-linear observation maps; and (ii) the contextual bandit setting that allows for time-varying feature vectors. 
\end{abstract}
\thispagestyle{empty} 
\pagestyle{empty}

\section{Introduction}
In a classical online learning or sequential decision-making problem, an agent (learner) interacts with an \textit{unknown} environment by taking certain actions; as feedback, it observes rewards corresponding to these actions. Through repeated interactions with the environment, the typical goal of the agent is to take actions that maximize the expected sum of rewards over a time-horizon. To achieve this goal, the agent needs to occasionally \textit{explore}  actions that improve its understanding of the rewarding generating process. However, such exploratory actions need not maximize the current reward, leading to an \textit{exploration-exploitation dilemma}. Motivated by applications in large-scale web recommendation systems, distributed robotics, and federated learning, we consider a setting involving multiple agents facing such a dilemma while interacting with a \textit{common} uncertain environment. In particular, we ground our study within the linear stochastic bandit formalism \cite{dani,abbasi}. 

In our model, $M$ agents interact with the same linear bandit characterized by a $d$-dimensional unknown parameter $\theta_*$, and a finite set of $K$ arms (actions). These agents can collaborate via a central server to improve performance, as measured by cumulative regret. As examples, consider (i) a team of robots exploring actions (arms) in a common environment and interacting with a central controller; and (ii) a group of people exploring restaurants (arms) and writing reviews for a web recommendation server. In each of the above examples, there is a clear reason to collaborate: by exchanging information, each agent can reduce its uncertainty about the arms faster than when it acts alone, and thereby incur lesser regret. However, the situation becomes murkier and more delicate when, aligning with reality, certain agents \textit{misbehave}: What if certain robots get attacked or certain people deliberately write spam reviews? Can we still expect to benefit from exchanging (potentially corrupted) information? 

The above questions are motivated by the fact that \textit{security} is a primary challenge in modern large-scale computing systems, where individual components are susceptible to attacks. Responding to this challenge, a growing body of work has started to investigate the design and analysis of distributed algorithms that are provably robust to a small fraction of adversarial agents. Notably, this body of work has focused primarily on empirical risk minimization/stochastic optimization \cite{chensu,blanchard,yin,draco,alistarhbyz,xie,lirsa,sundaram,ghosh1,ghosh2,ghosh3,suByz,Kan,nirupam,karimireddy,adibi}, with the eventual goal of solving  supervised learning problems. However, when it comes to  multi-agent sequential decision-making  under uncertainty (e.g., bandits and reinforcement learning), our understanding of adversarial robustness is quite limited. The goal of this paper is to bridge the above gap by providing crisp, rigorous answers to the following questions. 
\vspace{2.5mm}
\par \textit{In a multi-agent linear stochastic bandit problem, can we hope for benefits of collaboration when a fraction $\alpha$ of the agents are adversarial? If so, what are the fundamental limits of such benefits?}
\vspace{0.1mm}

\par As far as we are aware, the answers to these questions have thus far remained elusive,  motivating our current study. The main technical hurdle we must overcome is to delicately balance the exploration-exploitation trade-off in the presence of both statistical uncertainties due to the environment, and \textit{worst-case} adversarial behavior. Importantly, the above trade-off - intrinsic to sequential decision-making - is absent in static optimization problems. Thus, the ideas used to guarantee robustness for distributed optimization do not apply to our problem, making our task quite non-trivial. In what follows, we summarize the main contributions of this paper. 

\subsection{Our Contributions} We contribute to a principled study of several canonical \textit{structured} linear bandit settings with adversarial agents. Our specific contributions are as follows.
\begin{itemize}
\item \textbf{Robust Collaborative Linear Bandit Algorithm.} We propose \texttt{RCLB} - a phased elimination algorithm that relies on distributed exploration, and balances the exploration-exploitation dilemma in the presence of adversaries via carefully constructed \textit{robust confidence intervals}. We prove that \texttt{RCLB} guarantees $\tilde{O}\left( \left(\alpha + {1}/{\sqrt{M}} \right) \sqrt{dT} \right)$ regret for each good agent; see Theorem \ref{thm:RCPLB}. This result is both novel, and significant in that it reveals that for small values of $\alpha$, each good agent can considerably improve upon the optimal single-agent regret bound of $\tilde{O}(\sqrt{dT})$ via collaboration. Simply put, our result conveys a clear and important message: \textit{there is ample reason to collaborate despite adversaries}. Notably, when $\alpha=0$, the regret bound of \texttt{RCLB} is minimax-optimal in all relevant parameters: the model-dimension $d$, the horizon $T$, and the number of agents $M$. 

\item \textbf{Fundamental Limits.} At this stage, it is natural to ask: \textit{Is the additive $\alpha \sqrt{T}$ term in Theorem \ref{thm:RCPLB} simply an artifact of our analysis?} In Theorem \ref{thm:lower_bnd}, we establish a fundamental lower bound, revealing that such a term is in fact \textit{unavoidable}; it is the price one \textit{must} pay due to adversarial corruptions. A key implication of this result is that \textit{our work is the first to provide tight, near-optimal regret bounds for collaborative linear bandits with adversaries.} Interestingly, our results complement those of a similar flavor for distributed statistical learning with adversaries \cite{yin}; see Section \ref{sec:results_RCPLB} for further discussion on this topic. The proof of Theorem \ref{thm:lower_bnd} relies on a novel connection between the information-theoretic arguments in \cite{bubeck}, and ideas from the robust mean estimation literature  \cite{chenrobust,lai}. As such, our proof technique may be relevant for proving lower bounds in  related settings. 

In our next set of contributions, we significantly extend our algorithmic ideas and results to more general bandit models. 

\item \textbf{Generalized Linear Bandit Setting.} In Theorem \ref{thm:GLM}, we prove that one can achieve bounds akin to that in Theorem \ref{thm:RCPLB} for the generalized linear bandit model (GLM) \cite{GLM1,GLM2} that accounts for non-linear observation maps. To achieve this result, we propose a variant of \texttt{RCLB} that leverages very recently developed tools from high-dimensional robust Gaussian mean estimation \cite{Dalal}. Deriving robust confidence intervals for this setting requires some work: we exploit regularity properties of the non-linear observation model in conjunction with high-probability error bounds from \cite{Dalal} for this purpose. As far as we are aware, Theorem  \ref{thm:GLM} is the first result to establish adversarial-robustness for GLMs, allowing our framework to be applicable to a broad class of problems (e.g., logistic and probit regression models). 

\item \textbf{Contextual Bandit Setting.} Finally, we turn our attention to the contextual bandit setting where the feature vectors of the arms can change over time. This setting is practically quite relevant as web recommendation systems are often modeled as contextual bandits \cite{licontext}. The main challenge here arises from the need to simultaneously contend with time-varying optimal arms and adversaries. To handle this scenario, we develop a robust variant of the \texttt{SUPLINREL} algorithm \cite{auer2} that guarantees a near-optimal regret bound identical to that of Theorem \ref{thm:RCPLB}; see Theorem \ref{thm:contextual}. 
\end{itemize}

Overall, via the proposal of new robust algorithms complemented with tight analyses, our work takes an important step towards multi-agent sequential decision-making under uncertainty, in the presence of adversaries. For a summary of our main results, please see Table~\ref{table:results}. 

\begin{table}[t]
\centering
\begin{tabular}{|| c | c | c ||} 
 \hline
 Setting & Algorithm & Regret of each good agent \\
 \hline
 \hline
 Linear bandit model in \S\ref{sec:model}  & Algorithm  \ref{algo:RCPLB} & $\tilde{O}\left( \left(\alpha + {1}/{\sqrt{M}} \right) \sqrt{dT} \right)$ [Theorem \ref{thm:RCPLB}]\\ 
 \hline
Generalized linear bandit model in \S\ref{sec:gen_lin} & Algorithm \ref{algo:RC-GLM} & $\tilde{O}\left( \left(\alpha + \sqrt{d/{M}} \right) \sqrt{dT} \right)$ [Theorem \ref{thm:GLM}]\\
\hline
Contextual bandit model in \S\ref{sec:context} & Algorithm \ref{algo:SupLin} & $\tilde{O}\left( \left(\alpha + {1}/{\sqrt{M}} \right) \sqrt{dT} \right)$ [Theorem \ref{thm:contextual}] \\
\hline
\end{tabular}
\vspace{3.5mm}
\caption{Summary of the settings we study, the algorithms we propose for each setting, and the corresponding main results. Here, $\alpha$ is the corruption fraction, $M$ is the number of agents, $d$ is the model-dimension, and $T$ is the horizon. For each of the considered settings, the known single-agent minimax optimal regret bound without adversaries is $\tilde{\Theta}(\sqrt{dT})$.}
\vspace{-2mm}
\label{table:results}
\end{table}

\subsection{Related Work} We now provide a detailed discussion of relevant work.
\begin{itemize}
\item \textbf{Reward Corruption Attacks in Stochastic Bandits.} In the single-agent setting, there is a rich body of work that studies the effect of reward-corruption in stochastic bandits, both for the unstructured multi-armed bandit problem \cite{junadv,liuadv,lykouris,guptaadv}, and also for structured linear bandits \cite{bogunovic1,garcelon,bogunovic2}. In these works, an adversary can modify the true stochastic reward/feedback on certain rounds; a corruption budget $C$ captures the total corruption injected by the adversary over the horizon $T$. The attack model we study is fundamentally different: the adversaries in our setting can inject corruptions of \textit{arbitrary} magnitude in \textit{all} rounds, i.e., there are no budget constraints. As such, the algorithmic techniques in \cite{junadv,liuadv,lykouris,guptaadv,bogunovic1,garcelon,bogunovic2} do not apply to our model. 

Continuing with this point, we note that in \cite{guptaadv}, the authors proved an \textit{algorithm-independent} lower bound of $\Omega(C)$ on the regret. This lower bound suggests that for the reward-corruption attack model, when the attacker's budget $C$ scales linearly with the horizon $T$, there is no hope for achieving sub-linear regret. In \cite{kapoor}, the authors studied a reward-corruption model closely related to those in \cite{lykouris,guptaadv,bogunovic1}, where in each round, with probability $\eta$ (independently of the other rounds), the attacker can bias the reward seen by the learner. Similar to the lower bound in \cite{guptaadv}, the authors in \cite{kapoor} proved a lower bound of $\Omega(\eta T)$ on the regret for their model. In sharp contrast to the fundamental limits established in \cite{guptaadv,kapoor}, for our setup, as long as the corruption fraction $\alpha$ is strictly less than half, we prove that with high-probability it is in fact possible to achieve sub-linear regret. The key is that for our setting, the server can leverage ``clean" information from the good agents in every round; of course, the identities of such good agents are not known to the server. We finally note that beyond the task of minimizing cumulative regret, the impact of fixed-budget reward-contamination has also been explored for the problem of  best-arm identification in \cite{zhongBAI}. 

\item \textbf{Multi-Agent Bandits.} There is a growing literature that
studies multi-agent multi-armed bandit problems in the absence of adversaries, both over peer-to-peer networks, and also for the server-client architecture model  \cite{dMAB1,kalathil,kar,landgren1,landgren2,shahrampour,dMAB2,kolla,dMAB3,sankararaman,martinez,dubeyICML20,dubeyNIPS20,lalitha20,chawla1,chawla2,ghoshbandits,agarwal21,FedMAB1,FedMAB3}. 
The main focus in these papers is the design of coordination protocols among the agents that balance communication-efficiency with performance. A few very recent works 
\cite{dubeyByz,vialrob1,vialrob2,mitrahet} also look at the effect of attacks, but for the simpler unstructured multi-armed bandit problem \cite{auer}. Accounting for adversarial agents in the structured linear bandit setting we consider here requires significantly different ideas that we develop in this paper. 

\item \textbf{Security in Distributed Optimization and Federated Learning.} As we mentioned earlier, several papers have studied the problem of accounting for adversarial agents in the context of supervised learning \cite{chensu,blanchard,yin,draco,alistarhbyz,xie,lirsa,sundaram,ghosh1,ghosh2,ghosh3,suByz,Kan,nirupam,karimireddy,adibi}. One of the primary applications of interest here is the emerging paradigm of federated learning \cite{konevcny,bonawitz,mcmahan}. Different from the sequential decision-making setting we investigate in our paper, the aforementioned works essentially abstract out the supervised learning task as a \textit{static} distributed optimization problem, and then apply some form of secure aggregation on either gradient vectors or parameter estimates. 

\item \textbf{Robust Statistics.} The algorithms that we develop in this paper borrow tools from the literature on robust statistics, pioneered by Huber \cite{huber,huber2}. We point the reader to 
\cite{chenrobust,lai,cheng,minsker,lugosi,Dalal}, and the references therein, to get a sense of some of the main results in this broad area of research. In a nutshell, given multiple samples of a random variable - with a small fraction of samples corrupted by an adversary - the essential goal of this line of work is to come up with statistically optimal and computationally efficient robust estimators of the mean of the random variable. Notably, unlike both the sequential bandit setting and the iterative optimization setting, the robust statistics literature focuses on one-shot estimation. In other words, the adversary gets to corrupt the batch of samples \textit{only once}, and the effect of such corruption does not compound over time or iterations. 

\end{itemize}
\newpage
\textbf{Notation}. Given two scalars $a$ and $b$, we use $a \vee b$ and $a \wedge b$ to represent $\max\{a,b\}$ and $\min\{a,b\}$, respectively. For any positive integer $n$, we use $[n]$ to denote the set of integers $\{1, \ldots, n\}$. Given a matrix $A$, we use $A'$ to denote the transpose of $A$. Given two symmetric positive semi-definite matrices $A$ and $B$, we use $B \preccurlyeq A$ to imply that $A-B$ is positive semi-definite. Unless otherwise mentioned, we will use $\Vert \cdot \Vert$ to represent the Euclidean norm.
\section{Problem Formulation}
\label{sec:model}
We consider a setting comprising of a central server and $M$ agents; the agents can communicate only via the server. Each agent $i \in [M]$ interacts with the same linear bandit model characterized by an 
unknown parameter $\theta_*$ that belongs to a known compact set $\Theta \subset \mathbb{R}^{d}$. We assume  $\Vert \theta_* \Vert \leq 1.$  The set of actions $\mathcal{A}$ for each agent is given by $K$ distinct vectors in $\mathbb{R}^{d}$, i.e.,  $\mathcal{A}=\{a_1, \ldots, a_K\}$, where $K$ is a finite, positive integer. We will assume throughout that $\Vert a_k \Vert \leq 1, \forall a_k \in \mc{A}.$ Based on all the information acquired by an agent up to time  $t-1$, it takes an action $a_{i,t} \in \mathcal{A}$ at time $t$, and receives a reward $y_{i,t}$ given by the following observation model:
\begin{equation}
y_{i,t}=\langle \theta_*, a_{i,t} \rangle + \eta_{i,t}.
\label{eqn:Obs_model}
\end{equation}
Here, $\{\eta_{i,t}\}$ is a sequence of independent Gaussian random variables with zero mean and unit variance. Thus far, we have essentially described a  distributed/multi-agent linear stochastic bandit model. Departing from this standard model, we focus on a setting where a fraction $\alpha \in [0, 1/2)$ of the agents are adversarial; the adversarial set is denoted by $\mathcal{B}$, where $|\mathcal{B}| = \alpha M$. In particular, we consider a \textit{worst-case} attack model, where each adversarial agent $i\in \mathcal{B}$ is assumed to have complete knowledge of the system, and is allowed to act arbitrarily. Under this attack model, an agent $i\in \mathcal{B}$ can transmit arbitrarily corrupted messages to the central server. 
Our performance measure of interest is the following group regret metric $R_T$ defined w.r.t. the non-adversarial agents:
\begin{equation}
    R_T = \mathbb{E} \left[\sum_{i\in [M] \setminus \mathcal{B}} \sum_{t=1}^{T} \langle \theta_*, a_*-a_{i,t} \rangle \right],
\label{eqn:regret}
    \end{equation}
where $a_*  = \argmax_{a\in \mathcal{A}} \langle \theta_*, a \rangle$ is the optimal arm, and $T$ is the time horizon.\footnote{For ease of exposition, we assume that there is an unique optimal arm.} We will work under a regime where the horizon $T$ is large, satisfying $T\geq Md$. The goal of the good (non-adversarial) agents is to collaborate via the server and play a sequence of actions that minimize the group regret $R_T$. Let us now make a few key observations. In principle, each good agent can choose to act independently throughout (i.e., not talk to the server at all), and achieve $\tilde{O}(\sqrt{dT})$ regret by playing a standard bandit algorithm. Clearly, the group regret $R_T$ would scale as $\tilde{O}\left((1-\alpha) M \sqrt{dT}\right)$ in such a case. In the absence of adversaries however, one can achieve a significantly better group regret bound of $\tilde{O}\left( \sqrt{MdT}\right)$ via collaboration, i.e., the regret per good agent can be reduced by a factor of $\sqrt{M}$ relative to the case when it acts independently (see, for instance, \cite{dMAB3} and \cite{dubeyNIPS20}). Our specific interest in this paper is to investigate whether, and to what extent, one can retain the benefits of collaboration despite the worst-case attack model described above. Said differently, we ask the following question: 

\textit{Can we improve upon the trivial per agent regret bound of $\tilde{O}(\sqrt{dT})$ in the presence of adversarial agents?}

Throughout the rest of the paper, we will answer the above question in the affirmative by deriving novel robust algorithms for several structured  bandit models, and then establishing near-optimal regret bounds for each such model.
\newpage
\section{Robust Collaborative Phased Elimination Algorithm for Linear Stochastic Bandits}
\label{sec:algo}
In this section, we develop a robust phased elimination algorithm that achieves the near-optimal regret bound of  $\tilde{O}\left(\left(\alpha + \sqrt{{1}/{M}}\right)\sqrt{dT}\right)$ per good agent. This is non-trivial as we must account for the worst-case attack model described in Section \ref{sec:model}. To highlight the challenges that we need to overcome, consider the following scenario. During the initial stages of the learning process, when the arms in $\mathcal{A}$ have not been adequately sampled by the agents, even a good agent $i$ may have ``poor estimates" of the true payoffs associated with each arm, i.e., the variance associated with such estimates may be large. This statistical uncertainty can be exploited by the adversarial agents to their benefit. In particular, we need to devise an approach that can distinguish between benign stochastic perturbations (due to the noise in our model) and deliberate adversarial behavior. In what follows, we describe our proposed algorithm -  \texttt{Robust Collaborative Phased Elimination for Linear Bandits (RCLB)} - that precisely does so in a principled way.

\textbf{Description of \texttt{RCLB} (Algorithm \ref{algo:RCPLB}).} The \texttt{RCLB} algorithm we propose is inspired by the phased elimination algorithm in \cite[Chapter 22]{tor}, but features some key differences due to the distributed and adversarial nature of our problem. The algorithm proceeds in epochs/phases, and in each phase $\ell$, the server maintains an active candidate set $\mathcal{A}_{\ell}$ of potential optimal arms. The exploration of arms in $\mathcal{A}_{\ell}$ is distributed among the agents. Upon such exploration, each agent $i$ reports back a local estimate $\hat{\theta}^{(\ell)}_i$ of the unknown parameter $\theta_*$; adversarial agents can transmit arbitrarily corrupted messages at this stage. Using the local parameter estimates $\{\hat{\theta}^{(\ell)}_i\}_{i\in [M]}$, the server then constructs (i) a \textit{robust} estimate of the true mean payoff $\langle \theta_*, a \rangle$ for each active arm $a\in \mathcal{A}_{\ell}$ (line 8 of Algorithm  \ref{algo:RCPLB}), and (ii) an \textit{inflated} confidence interval that captures both statistical and adversarial uncertainties associated with such robust mean estimates. This step is crucial to our scheme and requires a lot of care as we explain shortly. With the robust mean payoffs and associated confidence intervals in hand, the server eliminates arms with rewards far away from that of the optimal arm $a_*$ (line 9 of Algorithm  \ref{algo:RCPLB}). The remaining arms enter the  active arm-set for phase $\ell+1$, namely $\mathcal{A}_{\ell+1}$. We now elaborate on the above ideas. 

\begin{algorithm}[t]
\caption{Robust Collaborative Phased Elimination for Linear Bandits   (\texttt{RCLB})}
\label{algo:RCPLB}  
 \begin{algorithmic}[1] 
\Statex \hspace{-5mm} \textbf{Input:} Action set $\mathcal{A}=\{a_1, \ldots, a_K\}$, confidence parameter $\delta$, and corruption fraction $\alpha.$
\Statex \hspace{-5mm} \textbf{Initialize:} $\ell=1$ and $\mathcal{A}_1=\mathcal{A}.$
\State Let $ V_{\ell}(\pi) \triangleq \sum_{a\in\mathcal{A}_{\ell}} \pi(a) aa'$ and $g_{\ell}(\pi) \triangleq \max_{a\in\mathcal{A}_{\ell}} {\Vert a \Vert^2}_{{V_{\ell}(\pi)}^{-1}}.$ Server solves an approximate G-optimal design problem to compute a distribution $\pi_{\ell}$ over $\mathcal{A}_{\ell}$ such that $g_{\ell}(\pi_{\ell}) \leq 2d$ and  $|\textrm{Supp}(\pi_{\ell})| \leq 48d \log\log d.$ 
\State For each $a\in\mathcal{A}_{\ell}$, server computes $m^{(\ell)}_a$ via  Eq. \eqref{eqn:arm_pulls}, and broadcasts $\{m^{(\ell)}_a\}_{a\in\mathcal{A}_{\ell}}$ to all agents. 
\For {$i\in [M]\setminus \mathcal{B}$} 
\State For each arm $a\in \mathcal{A}_{\ell}$, pull it $m^{(\ell)}_a$ times. Let $r^{(\ell)}_{i,a}$ be the average of the rewards observed by agent $i$ for arm $a$ during phase $\ell$. 
\State Compute local estimate $\hat{\theta}^{(\ell)}_i$ of $\theta_*$ as follows.\footnotemark
\begin{equation}
    \hat{\theta}^{(\ell)}_i = \tilde{V}^{-1}_{\ell} Y_{i,\ell}, \hspace{1.5mm} \textrm {where} \hspace{1.5mm} \tilde{V}_{\ell} = \hspace{-3mm}  \sum_{a\in\textrm{Supp}(\pi_{\ell})} \hspace{-3mm} m^{(\ell)}_a aa' \hspace{1mm} ; \hspace{1.5mm} Y_{i,\ell} = \hspace{-5mm} \sum_{a\in\textrm{Supp}(\pi_{\ell})} \hspace{-5mm} m^{(\ell)}_a r^{(\ell)}_{i,a} a. 
\label{eqn:local_estimate}
\end{equation}
\State Transmit $\hat{\theta}^{(\ell)}_i$ to server.  Adversarial agents can transmit arbitrary vectors at this stage. 
\EndFor
\State Server computes robust mean pay-offs for each active arm: for each $a \in \mathcal{A}_{\ell}$, estimate $\mu^{(\ell)}_{a}$ as 
$$ \mu^{(\ell)}_{a}=\texttt{Median}\left( \{\langle \hat{\theta}^{(\ell)}_i, a \rangle, i \in [M]\} \right). $$
\State Define robust confidence threshold $\gamma_{\ell} \triangleq \sqrt{2} C \left(1+\alpha\sqrt{M} \right) \epsilon_{\ell}$, where $C$ is as in Lemma \ref{lemma:rob_conf_main}.  Server performs phased elimination with the robust means and  $\gamma_{\ell}$ to update active arm set:
\begin{equation}
    \mathcal{A}_{\ell+1} = \{a\in\mathcal{A}_{\ell}: \max_{b\in\mathcal{A}_{\ell}} \mu^{(\ell)}_b - \mu^{(\ell)}_a \leq 2 \gamma_{\ell}\}. 
\label{eqn:phased_el}
\end{equation}
\State $\ell = \ell+1$ and \textbf{Goto} line 1. 
\end{algorithmic}
 \end{algorithm}
 
$\bullet$ \textit{Optimal Experimental Design.} To minimize the regret incurred in each phase $\ell$, we need to minimize the number of arm-pulls made to arms in $\mathcal{A}_{\ell}$. To this end, we will appeal to a well-known concept from statistics known as optimal experimental design. The concept is as follows. Let $\pi:\mathcal{A} \rightarrow [0, 1]$ be a distribution on $\mathcal{A}$ such that $\sum_{a\in\mathcal{A}} \pi(a)=1$. Now define
\begin{equation}
    V(\pi) \triangleq \sum_{a\in\mathcal{A}} \pi(a) aa'; \hspace{2mm} g(\pi) \triangleq \max_{a\in\mathcal{A}} {\Vert a \Vert^2}_{{V(\pi)}^{-1}}.
\label{eqn:design}
\end{equation}
The so-called \textit{G-optimal design problem} seeks to find a distribution (design) $\pi$ that minimizes $g$. When $\mathcal{A}$ is compact, and $\textrm{span}(\mathcal{A})=d$, the Kiefer-Wolfowitz theorem \cite[Theorem 21.1]{tor} tells us that there exists an optimal design $\pi_*$ that minimizes $g$, such that $g(\pi_*)=d$. Moreover, the distribution $\pi_*$ has support bounded above as $|\textrm{Supp}(\pi_*)| \leq d(d+1)/2$. In essence, sampling each arm $a\in\mathcal{A}$ in proportion to  $\pi_*(a)$ minimizes the number of samples/arm-pulls needed to achieve a desired level of precision in the  estimates of the arm means $\langle \theta_*, a \rangle$, $a\in \mathcal{A}$. For our purpose, we only need to solve an approximate G-optimal design problem: using the Frank-Wolfe algorithm and an appropriate initialization, one can in fact find an approximate optimal design $\pi$ such that $g(\pi) \leq 2d$, and $|\textrm{Supp}(\pi)| \leq 48 d \log \log d$ \cite[Chapter 3]{todd}, \cite[Chapter 21]{tor}. Accordingly, the server computes such an approximate optimal distribution $\pi_{\ell}$ over $\mathcal{A}_{\ell}$ in each epoch $\ell$ (line 1 of Algo.   \ref{algo:RCPLB}).\footnotetext{Throughout, we assume that $\tilde{V}^{-1}_{\ell}$ is invertible in each epoch $\ell$; this is the case when $\mathcal{A}_{\ell}$ spans $\mathbb{R}^{d}$. If $\mathcal{A}_{\ell}$ does not span $\mathbb{R}^{d}$, we can consider the lower dimensional subspace given by $\textrm{span}(\mathcal{A}_{\ell})$ and construct an approximate G-optimal design that is supported on $\tilde{O}(m)$ points, where $m=\textrm{dim}(\textrm{span}(\mathcal{A}_{\ell}))$.} 

$\bullet$ \textit{Construction of Robust Arm-Payoff Estimates and Confidence Intervals.} In each phase $\ell$, the server computes $T^{(\ell)}_a$ and $m^{(\ell)}_a$ for every $a\in\mathcal{A}_{\ell}$ as follows:
\begin{equation}
    T^{(\ell)}_a= \ceil*{\frac{\pi_{\ell}(a) d}{\epsilon^2_{\ell}} \log\left(\frac{1}{\delta_{\ell}}\right)} \hspace{2mm} \textrm{and} \hspace{2mm} m^{(\ell)}_a = \ceil*{\frac{T^{(\ell)}_a}{M}}, 
\label{eqn:arm_pulls}
\end{equation}
where $\pi_{\ell}(a)$ is obtained from the approximate $G$-optimal design problem, $\epsilon_{\ell}=2^{-\ell}$, $\delta_{\ell}=\bar{\delta}/{(K\ell^2)}$, and $\bar{\delta}$ is a design variable to be chosen later. The idea is to explore an arm $a$  $T^{(\ell)}_a$ times to estimate $ \langle \theta_*, a \rangle$ up to a precision that scales linearly with $\epsilon_{\ell}$; in later phases, we require progressively finer precision (hence, $\epsilon_{\ell}$ decays exponentially with $\ell$). The task of exploration is distributed among the agents, with each agent $i$ being assigned $m^{(\ell)}_a$ arm-pulls for every $a\in \mathcal{A}_{\ell}$. Using the rewards that it observes during phase $\ell$, each good agent $i\in [M] \setminus \mathcal{B}$ computes a local estimate $\hat{\theta}^{(\ell)}_i$ of $\theta_*$, and transmits it to the server (lines 3-7 of Algorithm  \ref{algo:RCPLB}). The key question now is as follows: \textit{How should the server use the local estimates $\{\hat{\theta}^{(\ell)}_i\}_{i\in [M]}$}? Let us consider two natural strategies.
\newpage
\textbf{Candidate Strategies.} One option could be to use the local estimates $\{\hat{\theta}^{(\ell)}_i\}_{i\in [M]}$ along with a high-dimensional robust mean estimation algorithm to compute a robust estimate of $\theta_*$. Yet another strategy could be for the server to query the raw observations (i.e., the $y_{i,t}$'s) from the agents, use an univariate robust mean estimation algorithm (e.g., trimmed mean or median) to generate a ``clean" version of each observation, and then use these clean observations to compute a robust estimate of $\theta_*$. Although feasible, each of the above strategies can unfortunately lead to an additional $\sqrt{d}$ factor in the regret bound; we discuss this point in detail in  Appendix \ref{app:alternate}. The main message we want to convey here is that certain natural candidate solutions can lead to sub-optimal regret bounds. This, in turn, highlights the importance of our proposed approach.

\textbf{Main Ideas.} Our main insight is the following: to achieve near-optimal regret bounds, one need not go through the route of first computing a robust estimate of $\theta_*$. As our analysis will soon reveal, it suffices to instead compute robust estimates of the arm pay-offs $\{\langle \theta_*, a \rangle \}_{a\in\mathcal{A}_{\ell}}$ \textit{directly} by employing the estimator in line 8 of Algorithm \ref{algo:RCPLB}. The key statistical property that we exploit here is that for each $a\in \mc{A}_{\ell}$ and $i\in [M]\setminus \mc{B}$, the quantity $\langle \hat{\theta}^{(\ell)}_i, a \rangle$ is conditionally-Gaussian with mean $\langle \theta_*, a \rangle$. This observation informs the choice of the median operator in line 8 of Algorithm \ref{algo:RCPLB}. Our next task is to compute appropriate confidence intervals for the robust arm-mean-estimates in order to eliminate sub-optimal arms. This is a delicate task as such confidence intervals need to account for both statistical uncertainties \textit{and}  adversarial perturbations. Indeed, if the confidence intervals are too tight, then they can lead to elimination of the optimal arm $a_*$; if they are too loose, then they can lead to large regret. The confidence threshold $\gamma_{\ell}$ in line 9 of Algorithm \ref{algo:RCPLB} strikes just the right balance; the choice of such intervals is justified in Lemma \ref{lemma:rob_conf_main}. 

This completes the description of \texttt{RCLB}; in the next section, we will characterize its performance. 

\subsection{Analysis of \texttt{RCLB}}
\label{sec:results_RCPLB}
In this section, we state and discuss our main result concerning the performance of \texttt{RCLB}. 

\begin{mytheo}[label=thm:RCPLB]{Performance of \texttt{RCLB}}{}
Suppose the corruption fraction satisfies $\alpha < 1/2$. Given any $\delta \in (0,1)$, set the design parameter $\delta'=\delta/(10K)$. Then,  \texttt{RCLB}  guarantees that with probability at least $1-\delta$, the following holds for each good agent $i\in [M]\setminus\mathcal{B}$: 
\begin{equation}
    \sum_{t=1}^{T} \langle \theta_*, a_*-a_{i,t} \rangle = \tilde{O}\left(\left(\alpha + \sqrt{\frac{1}{M}} \right) \sqrt{dT} \right).
\label{eqn:bound_RCPLB}
\end{equation}
\end{mytheo}

The following is an immediate corollary on the group regret $R_T$.
\begin{corollary}
\label{corr:group_reg}
(\textbf{Bound on Group Regret})
Under the conditions of Theorem \ref{thm:RCPLB}, we have:
\begin{equation}
    R_T = \tilde{O}\left(\left(\alpha M + \sqrt{M}\right) \sqrt{dT} \right).
\end{equation}
\end{corollary}

\textbf{Main Takeaways.} From Theorem \ref{thm:RCPLB}, we note that \texttt{RCLB} guarantees sublinear regret despite the presence of adversarial agents. More importantly, the regret bound in Eq. \eqref{eqn:bound_RCPLB} has optimal dependence on the model-dimension $d$, the horizon $T$, and also on the number of agents $M$ when $\alpha =0$ (i.e., in the absence of adversaries). \textit{When $\alpha$ is small, Eq. \eqref{eqn:bound_RCPLB} reveals that one can indeed retain the benefits of collaboration, and improve upon the trivial per agent regret of $\tilde{O}(\sqrt{dT})$ - this is one of the main messages of our paper.} Interestingly, our result in Theorem \ref{thm:RCPLB} mirrors that of a similar flavor for distributed stochastic optimization under attacks: given $M$ machines, $\alpha$ fraction of which are corrupt, the authors in \cite{yin} showed that no algorithm can achieve statistical error lower than $\tilde{\Omega}\left(\alpha/\sqrt{T}+1/\sqrt{MT}\right)$ for strongly convex loss functions; here, $T$ is the number of samples on each machine. \textit{As far as we are aware, this is the first work to establish an analogous result for collaborative linear stochastic bandits}. Inspired by the lower bound in  \cite{yin}, one may ask: \textit{Is the additive $\alpha \sqrt{T}$ term in Eq. \eqref{eqn:bound_RCPLB} unavoidable?} We will provide a definitive answer to this question in Section \ref{sec:lower_bounds}. In what follows, we provide some intuition as to why \texttt{RCLB} works. 

\textbf{Proof Outline of Theorem \ref{thm:RCPLB}.} The first main step in our analysis of Theorem \ref{thm:RCPLB} is to provide guarantees on the estimates $\{\mu^{(\ell)}_a\}_{a\in \mc{A}_{\ell}}$ computed in line 8 of \texttt{RCLB}. This is achieved in the following result. 

\begin{mylem}[label=lemma:rob_conf_main]{Robust Confidence Intervals for \texttt{RCLB}}{}
Fix any epoch $\ell$. There exists an universal constant $C >0$ such that for each active arm $b \in \mc{A}_{\ell}$, the following holds with probability at least $1-\delta_{\ell}$:
\begin{equation}
    |\mu^{(\ell)}_b - \langle \theta_*, b \rangle| \leq \gamma_{\ell}, \hspace{2mm} \textrm{where} \hspace{2mm} \gamma_{\ell}=\sqrt{2} C \left(1+\alpha\sqrt{M} \right) \epsilon_{\ell}.
\end{equation}
\end{mylem}

Equipped with the above result, we argue that with high-probability, (i) the optimal arm $a_*$ is never eliminated by \texttt{RCLB} (Lemma \ref{lemma:good_arm_RCPLB} in Appendix   \ref{app:proof_RCPLB}); and (ii) in each epoch $\ell$, an active arm in $\mc{A}_{\ell}$ can contribute to at most $O(\gamma_{\ell})$ per-time-step regret (Lemma \ref{lemma:bad_arm_RCPLB} in Appendix   \ref{app:proof_RCPLB}). Putting these pieces together in a careful manner yields the desired result; we defer a detailed proof of Theorem \ref{thm:RCPLB} to Appendix \ref{app:proof_RCPLB}. 

In the next section, we will derive a lower bound that provides fundamental insights into the impact of the adversarial agents for the multi-agent sequential decision-making problem considered in this paper. In particular, this result will indicate that our bound in Theorem \ref{thm:RCPLB} is near-optimal. We close this section with a few remarks.

\begin{remark} \textbf{(Infinite Action Sets)} Note that our result in Theorem \ref{thm:RCPLB} is for the case when the action set $\mc{A}$ is finite. One can extend this result to the infinite action setting (provided that $\mc{A}$ is a compact set) using fairly standard covering arguments. Deriving an analogue of Theorem \ref{thm:RCPLB} for the case when the action set is finite, but time-varying, requires much more work. We will focus on this topic in Section \ref{sec:context}.
\end{remark}

\begin{remark} (\textbf{Communication Complexity of \texttt{RCLB}}) It is not hard to see that the number of epochs/phases in \texttt{RCLB} is $O(\log(MT))$; for a precise reasoning, one can refer to Appendix \ref{app:proof_RCPLB}. Since communication between the server and the agents  occurs only once in every epoch, we note that the communication complexity of \texttt{RCLB} scales logarithmically with the horizon $T$. Thus, \texttt{RCLB} not only leads to near-optimal regret bounds in the face of worst-case adversarial attacks, it is also communication-efficient by design. This is an important point to take note of as communication-efficiency is a key consideration in large-scale computing paradigms such as federated learning.

\end{remark}

\section{A Lower Bound and Fundamental Limits}
\label{sec:lower_bounds}
In this section, we assess the optimality of the regret bound obtained in Theorem \ref{thm:RCPLB}. To do so, we consider a slightly different attack model: we assume that each of the agents is adversarial with probability $\alpha$, independently of the other agents. Thus, the expected fraction of adversaries is $\alpha$. Next, to provide a clean argument, we will focus on a class of policies $\Pi$ where at each time-step $t$, the server assigns the \textit{same} action to every agent, i.e., $a_{i,t}=a_t, \forall i\in [M]$. We note that all the algorithms developed in this paper adhere to policies in $\Pi$. Moreover, policies within the class $\Pi$ yield the minimax optimal per-agent regret bound of order $\tilde{\Theta}(\sqrt{dT}/\sqrt{M})$ when $\alpha=0$. Since the standard multi-armed bandit setting \cite{auer} is a special case of the structured linear bandit setting considered here, a lower bound for the former implies one for the latter.\footnote{Indeed, when the action set for the linear bandit problem corresponds to the standard orthonormal unit vectors, we recover the unstructured multi-armed bandit setting.} With this in mind, let us denote by $\mc{E}^{(K)}_{\mc{N}}(1)$ the class of multi-armed bandits with $K$ arms, where the reward distribution of each arm is Gaussian with unit variance. An instance $\nu_{\mu} \in \mc{E}^{(K)}_{\mc{N}}(1)$ is characterized by the mean vector $\mu \in \mathbb{R}^{K}$ associated with the $K$ arms. Let us also denote by  $R^{(s)}_T(\nu_{\mu})$ the expected cumulative regret of the server (which is the same as that of a good agent $i\in [M]\setminus \mc{B}$) when it interacts with the instance $\nu_{\mu}$. We can now state the following result which establishes a fundamental lower bound for our problem. 

\begin{mytheo}[label=thm:lower_bnd]{Fundamental Limit}{}
 Given any policy in $\Pi$, there exist two distinct instances $\nu_{\mu}, \nu_{\mu'} \in \mc{E}^{(2)}_{\mc{N}}(1)$, and an universal constant $c > 1$, such that 
\begin{equation}
    \max\{R^{(s)}_T(\nu_{\mu}), R^{(s)}_T(\nu_{\mu'})\} \geq c \alpha \sqrt{T},
\end{equation}
irrespective of the number of agents $M$.
\end{mytheo}

\textbf{Main Takeaways.} Observe from Eq.~\eqref{eqn:bound_RCPLB} that even when $M $ is arbitrarily large, the additive $\alpha \sqrt{T}$ term due to the adversaries remains unaffected. This leads to the following natural question: \textit{Is this term truly unavoidable or just an artifact of our analysis?} Theorem \ref{thm:lower_bnd} settles this question by revealing a fundamental performance limit: \textit{every policy} in $\Pi$ has to suffer the additive $\alpha \sqrt{T}$ regret, regardless of the number of agents. Thus, taken together, Theorems \ref{thm:RCPLB} and \ref{thm:lower_bnd}  provide the first set of tight, near-optimal regret guarantees for the setting considered in this paper. We consider this to be a significant contribution of our work. 

One question that remains open is the following. While our analysis reveals that the effect of adversarial corruptions will get manifested in an \textit{unavoidable} additive $\alpha \sqrt{T}$ term, it is not clear whether such a term should be multiplied by the dimension-dependent quantity  $\sqrt{d}$ (as in Eq.~\eqref{eqn:bound_RCPLB}).  Specifically, the construction that we employ in the proof of Theorem \ref{thm:lower_bnd} relies on a 2-armed bandit instance for the unstructured multi-armed bandit problem. As such, our proof does not shed any light on dimension dependence. We leave further investigations along this line for future work.

In what follows, we briefly sketch the main idea behind the proof of Theorem \ref{thm:lower_bnd}.

\textbf{Proof Idea for Theorem \ref{thm:lower_bnd}}. For our setting, the standard techniques to prove lower bounds for non-adversarial bandits do not directly apply. The lower-bound proofs used for reward-corruption models \cite{kapoor}, and attacks with a fixed budget \cite{bogunovic2}, are not applicable either. This motivates us to use a new proof technique that combines information-theoretic arguments in \cite{bubeck} with ideas from the robust mean estimation literature \cite{chenrobust,lai}. Specifically, for a two-armed bandit setting, we carefully construct two instances and attack strategies such that the joint distribution of rewards seen by the server is \textit{identical} for both instances, regardless of the number of agents.  Moreover, the instances are constructed such that (i) the optimal arm in one instance is sub-optimal for the other; and (ii) the per-time-step regret for selecting a sub-optimal arm in either instance is $\Omega(\alpha/\sqrt{T})$. For a detailed proof, we refer the reader to Appendix \ref{app:proof_lower_bnd}. 

Having established tight bounds for the linear bandit model in Section \ref{sec:model}, in the sequel, we will show how our algorithmic ideas and results can be significantly extended to more general settings. 

\section{Extension to Generalized Linear Models with Adversaries}
\label{sec:gen_lin}
In this section, we will show how to achieve a regret bound akin to that in Theorem \ref{thm:RCPLB} for a setting where the observation model is no longer restricted to be linear in the model parameter $\theta$. Instead, we will consider the non-linear observation model shown below \cite{GLM1,GLM2}, known as the generalized linear model (GLM):
\begin{equation}
    y_{i,t}=\mu\left(\langle \theta_*, a_{i,t} \rangle \right)+\eta_{i,t},
\label{eqn:gen_obs_model}
\end{equation}
where $\mu:\mathbb{R} \rightarrow \mathbb{R}$ is a continuously differentiable function typically referred to as the (inverse) \textit{link} function, and $\eta_{i,t} \sim \mc{N}(0, 1)$ is as before. Our goal is to now  control the following notion of regret: 
\begin{equation}
    R^{\textrm{GLM}}_T = \mathbb{E} \left[\sum_{i\in [M] \setminus \mathcal{B}} \sum_{t=1}^{T} \left(\mu \left(\langle \theta_*, a_*\rangle \right) - \mu \left( \langle \theta_*, a_{i,t} \rangle \right) \right)  \right],
\label{eqn:GLM_regret}
    \end{equation}
where $a_* = \argmax_{a\in \mc{A}} \mu(\langle \theta_*, a \rangle)$. The main technical challenge relative to the setting considered in Section \ref{sec:model} pertains to the construction of the robust confidence intervals. In particular, the non-linearity of the map $\mu(\cdot)$ makes it hard to apply the technique employed in line 8 of \texttt{RCLB}, necessitating a different approach that we describe next. We start with the  following standard assumption that is typical in the literature on generalized linear models \cite{GLM1,GLM2}.

\begin{assumption} \label{ass:link} The function $\mu:\mathbb{R} \rightarrow \mathbb{R}$ is continuously differentiable, Lipschitz with constant $k_2 \geq 1$, and such that $$k_1=\min\{1,\inf_{\theta\in\Theta, a \in \mc{A}} \dot{\mu}(\langle \theta, a \rangle) \} > 0.$$
Here, $\dot{\mu}(\cdot)$ is used to represent the derivative of $\mu(\cdot)$.
\end{assumption}

 Next, for any $\theta\in \mathbb{R}^{d}$, we define  $h_{\ell}(\theta) \triangleq \sum_{a\in\textrm{Supp}(\pi_{\ell})}  m^{(\ell)}_a \mu(\langle \theta,a \rangle) a$, where $m^{(\ell)}_a$ is as in Eq.~\eqref{eqn:arm_pulls}. We now describe a variant of \texttt{RCLB} - dubbed \texttt{RC-GLM} - for the generalized linear bandit model considered in this section. Notably, unlike the standard bandit algorithms \cite{GLM1,GLM2} for GLM's that build on \texttt{LinUCB}, \texttt{RC-GLM} is based on phased elimination.  

\textbf{Description of \texttt{RC-GLM}.} Our algorithm uses as a sub-routine the recently proposed Iteratively Reweighted Mean Estimator for computing a robust estimate of the mean of high-dimensional Gaussian random variables with adversarial outliers \cite{Dalal}. Specifically, suppose we are given $M$ $d$-dimensional samples $x_1, \ldots, x_M$, such that $(1-\alpha)M$ of these samples are drawn i.i.d. from $\mc{N}(v, \Sigma)$, where $v \in \mathbb{R}^d$ is an unknown mean vector, and $\Sigma \in \mathbb{R}^{d\times d}$ is a known covariance matrix. The remaining $\alpha M$ samples are adversarial outliers and can be arbitrary. The estimator in \cite{Dalal} takes as input the $M$ samples, the corruption fraction $\alpha \in (0,1)$, and the covariance matrix $\Sigma$. It then outputs an estimate $\hat{v}$ of $v$ such that with high probability, $\Vert \hat{v} - v \Vert = \tilde{O} \left( \Vert \Sigma \Vert^{1/2}_2 \left( \sqrt{d/M} + \alpha \sqrt{\log(1/\alpha)} \right) \right).$ Importantly, the estimator in \cite{Dalal} runs in polynomial-time, and is minimax-rate-optimal. Let us now see how this estimator - described in Appendix \ref{app:gen_lin} - can be applied to our setting. 

We only describe the key differences of \texttt{RC-GLM} relative to \texttt{RCLB} here, and defer a detailed description of \texttt{RC-GLM} to Appendix \ref{app:gen_lin}. To get around the difficulty posed by the non-linear link function, our main idea is to first compute a robust estimate $\hat{\theta}^{(\ell)}$ of $\theta_*$ at the server, and then use it to develop a phased elimination strategy. To that end, instead of computing a local estimate $\hat{\theta}^{(\ell)}_i$ as in \texttt{RCLB}, each good agent $i\in [M]\setminus \mc{B}$ transmits $Y_{i,\ell}$ to the server, and the server computes a vector $X_{\ell}$ as follows: $X_{\ell}=\texttt{ITW}(\{\tilde{V}^{-1/2}_{\ell} Y_{i,\ell}, i \in [M]\})$. Here, $\tilde{V}^{-1/2}_{\ell}$ and $Y_{i,\ell}$ are as in Eq.~\eqref{eqn:local_estimate}, and we used $\texttt{ITW}(\cdot)$ to denote the output of the robust estimator in \cite{Dalal}. Our key observation here is that for each good agent $i$, $\tilde{V}^{-1/2}_{\ell} Y_{i,\ell}$ is a $d$-dimensional Gaussian random variable with mean $\tilde{V}^{-1/2}_{\ell} h_{\ell}(\theta_*)$, and covariance matrix $\Sigma= I_d$. This observation justifies the use of the robust high-dimensional Gaussian mean estimator in \cite{Dalal}. The fact that $\Sigma=I_d$ is crucial in our algorithm design as the error-bound in \cite{Dalal} scales with the 2-norm of $\Sigma$. Essentially, the above steps enable us to extract a statistic $X_{\ell}$ that captures information about the agents' observations during epoch $\ell$. Using this statistic, the server next computes an estimate $\hat{\theta}^{(\ell)}$ of $\theta_*$ by solving  $h_{\ell}(\hat{\theta}^{(\ell)}) = \tilde{V}^{1/2}_{\ell} X_{\ell}$,\footnote{We argue in Appendix \ref{app:gen_lin} that this equation admits a unique solution.} and employs the following phased elimination strategy:
$$ \mathcal{A}_{\ell+1} = \{a\in\mathcal{A}_{\ell}: \max_{b\in\mathcal{A}_{\ell}} \mu(\langle \hat{\theta}^{(\ell)}, b \rangle) - \mu(\langle \hat{\theta}^{(\ell)}, a \rangle) \leq 2 \bar{\gamma}_{\ell}\}, \hspace{2mm} \textrm{where} $$
$$ \bar{\gamma}_{\ell} = \bar{C} \frac{k_2}{k_1}  \left(\sqrt{d}+\alpha\sqrt{M \log(1/\alpha)} \right) \epsilon_{\ell}. 
$$
Here, $\bar{C}$ is an universal constant known to the server. Deriving an analogue of Lemma \ref{lemma:rob_conf_main} to compute the robust confidence threshold $\bar{\gamma}_{\ell}$ (as shown above) requires some work. This is achieved by exploiting the regularity properties of the link function in tandem with the confidence bounds in \cite{Dalal}; we refer the reader to Appendix \ref{app:gen_lin} for details of such a derivation. 

The main result of this section is as follows. 

\begin{mytheo}[label=thm:GLM]{Performance of Algorithm \texttt{RC-GLM}}{}
 Suppose the corruption fraction satisfies $\alpha < (5-\sqrt{5})/10$. Given a confidence threshold $\delta \in (0,1)$, suppose $M$ satisfies
 $$ M > \log{\left( \frac{160K^2 T^2}{\delta}\right)}.$$ The  \texttt{RC-GLM} algorithm then  guarantees that with probability at least $1-\delta$, the following holds for each good agent $i\in [M]\setminus\mathcal{B}$: 
\begin{equation}
    \sum_{t=1}^{T} \left(\mu \left(\langle \theta_*, a_*\rangle \right) - \mu \left( \langle \theta_*, a_{i,t} \rangle \right) \right) = \tilde{O}\left(\left( \frac{k_2}{k_1}\right)\left(\alpha \sqrt{\log\left({1}/{\alpha}\right)} + \sqrt{\frac{d}{M}} \right) \sqrt{dT} \right).
\label{eqn:bound_GLM}
\end{equation}
\end{mytheo}
\begin{proof}
Please see Appendix \ref{app:gen_lin}.
\end{proof}

\textbf{Main Takeaways.} Theorem \ref{thm:GLM} significantly generalizes Theorem \ref{thm:RCPLB} and shows that even for general non-linear observation maps, one can continue to reap the benefits of collaboration in the presence of adversaries. In particular, Theorem \ref{thm:GLM} is the only result we are aware of that provides guarantees of  adversarial-robustness for generalized linear models. As we mentioned in the introduction, the main implication of this result is that our framework can be used for a broad class of problems, e.g., probit and logistic regression models. We point out that using Theorem \ref{thm:GLM}, one can derive a high-probability bound of $\tilde{O}\left(\left( {k_2}/{k_1}\right)\left(\alpha M \sqrt{\log\left({1}/{\alpha}\right)} + \sqrt{{dM}} \right) \sqrt{dT} \right)$
on the group regret $R^{\textrm{GLM}}_T$. The details of this analysis are similar to that for Corollary \ref{corr:group_reg}. 

It is instructive to compare the bound for \texttt{RCLB} in Eq.~\eqref{eqn:bound_RCPLB} with that for \texttt{RC-GLM} in Eq.~\eqref{eqn:bound_GLM}. Notably, in place of the $\sqrt{\frac{1}{M}}$ term in the former bound,  we have a $\sqrt{\frac{d}{M}}$ term in the latter. The additional $\sqrt{d}$ factor in Eq.~\eqref{eqn:bound_GLM} is essentially inherited from the error-rate guarantees of the robust mean estimator in \cite{Dalal}. It would be interesting to see if this bound can be tightened by eliminating the additional dependence on the model dimension $d$. We leave further investigations along this route for future work. 
\newpage
\section{Robust Collaborative Contextual Bandits with Adversaries}
\label{sec:context}
In this section, we will consider a collaborative contextual bandit setting where at each time-step $t\in [T]$, the server and the agents observe $K$ $d$-dimensional feature vectors, $\{x_{t,a} | a \in [K]\}$, with $\Vert x_{t,a} \Vert \leq 1, \forall a \in [K]$ and $\forall t \in [T]$. We assume that the adversarial agents have no control over the generation of the feature vectors. Associated with each arm $a \in [K]$, the stochastic reward observed by an agent $i\in [M]$ comes from the following observation model:
\begin{equation}
    y_{i,t}(a)= \langle \theta_*, x_{t,a} \rangle + \eta_{i,t}(a),
\label{eqn:obs_context}
\end{equation}
where $\{\eta_{i,t}(a)\}$ are drawn i.i.d. from $\mc{N}(0,1)$. At each time step $t$, a good agent $i\in [M] \setminus \mc{B}$ plays an action $a_{i,t} \in [K]$, and receives the corresponding reward $r_{i,t} \triangleq y_{i,t}(a_{i,t})$ based on the observation model in Eq.~\eqref{eqn:obs_context}. The main difference of the setting considered here relative to the one in Section \ref{sec:model} is that the feature vectors for each arm can change over time. As a result, the optimal action $a^*_t = \argmax_{a \in [K]} \langle \theta_*, x_{t,a} \rangle$ can change over time, making it particularly challenging to compete with a time-varying optimal action in the presence of adversaries. This dictates the need for a different algorithmic strategy compared to the one we developed in Section \ref{algo:RCPLB}. Before we develop such a strategy, let us first formally define the performance metric of interest to us in this setting: 
\begin{equation}
    R^{\textrm{Context}}_T = \mathbb{E} \left[\sum_{i\in [M] \setminus \mathcal{B}} \sum_{t=1}^{T} \langle \theta_*, x_{t,a^*_t}- x_{t,a_{i,t}}\rangle \right]. 
\label{eqn:regret_context}
\end{equation}
\begin{algorithm}[t]
\caption{Robust BaseLinUCB (at Server)}
\label{algo:Base}  
 \begin{algorithmic}[1] 
\Statex \hspace{-5mm} \textbf{Input:} Confidence parameter $\bar{\delta}$, corruption fraction $\alpha$, and index set $\psi_t \subseteq [t-1]$.
\State $A_t \gets \frac{I_d}{M} + \sum_{\tau\in \psi_t} x_{\tau, a_{\tau}} x'_{\tau, a_{\tau}}.$ 
\For {$i \in [M]$} \Comment{Compute local parameters for each agent}
\State $b_{i,t} \gets \sum_{\tau \in \psi_t} r_{i,\tau} x_{\tau,a_{\tau}}; \hat{\theta}_{i,t} \gets A^{-1}_t b_{i,t}$
\EndFor
\For {$a \in [K]$} \Comment{Compute robust estimates for each feature vector}
\State $\hat{r}_{t,a} \gets \texttt{Median}\left( \{\langle \hat{\theta}_{i,t}, x_{t,a} \rangle, i \in [M]\} \right); w_{t,a} \gets \left(\alpha + 2C \sqrt{\frac{\log(\frac{1}{\bar{\delta}})}{M}}\right) {\Vert x_{t,a} \Vert}_{A^{-1}_t}$, where $C$ is as in Lemma \ref{lemma:rob_conf_main}.
\EndFor
\end{algorithmic}
 \end{algorithm}
The main question we are interested in answering is the following: \textit{For the contextual bandit setting described above, can one still hope for benefits of collaboration in the presence of adversaries?} In what follows, we will answer this question in the affirmative by developing a variant of the \texttt{SUPLINREL} algorithm \cite{auer2,Chucontext}. 

\begin{algorithm}[t]
\caption{Robust Collaborative SupLinUCB for Contextual Bandits (at Server)}
\label{algo:SupLin}  
 \begin{algorithmic}[1] 
\Statex \hspace{-5mm} \textbf{Input:}  Confidence parameter $\delta$, corruption fraction $\alpha$, and horizon $T$. 
\State $S \gets \ceil*{\ln T}; \psi^{(s)}_1 \gets \emptyset,  \forall s \in [S].$ 
\For {$t \in [T]$} 
\State $s \gets 1$ and $\mc{A}_1 \gets [K]$.
 \Repeat 
 \State \hspace{-5mm}Use Algorithm \ref{algo:Base} with $\bar{\delta}=\delta/(KST)$, and index set $\psi^{(s)}_t$ to compute robust estimates $\{\hat{r}^{(s)}_{t,a}, w^{(s)}_{t,a}\}$ of the means and variances of the payoffs associated with each arm $a\in \mc{A}_s$. 
 \State \hspace{-5mm}If $w^{(s)}_{t,a} > 2^{-s}/\sqrt{M}$ for some $a \in \mc{A}_s$, then choose this arm, i.e., set $a_t = a$. Store the corresponding phase: $\psi^{(s)}_{t+1}=\psi^{(s)}_{t} \cup \{t\}, \psi^{(\ell)}_{t+1}=\psi^{(\ell)}_{t} \hspace{1mm} \forall \ell \neq s.$  
 \State \hspace{-5mm}Else if $w^{(s)}_{t,a} \leq  1/\sqrt{MT} \hspace{1.5mm} \forall a \in \mc{A}_s$, then select the arm with the \textit{highest robust upper confidence bound}: $a_t = \argmax_{a \in \mc{A}_s} \left(\hat{r}^{(s)}_{t,a}+w^{(s)}_{t,a} \right).$ Do not store this phase, i.e., $\psi^{(\ell)}_{t+1}=\psi^{(\ell)}_{t} \hspace{1mm} \forall \ell \in [S].$ 
\State \hspace{-5mm}Else if $w^{(s)}_{t,a} \leq  2^{-s}/\sqrt{M} \hspace{1.5mm} \forall a \in \mc{A}_s$, then update active arm-set as
$$ \mc{A}_{s+1} = \{a\in \mc{A}_s| \max_{b\in \mc{A}_s} \left(\hat{r}^{(s)}_{t,b}+w^{(s)}_{t,b} \right) - \left(\hat{r}^{(s)}_{t,a}+w^{(s)}_{t,a} \right) \leq 2^{(1-s)}/{\sqrt{M}} \}.$$
\State \hspace{-5mm}$s \gets s+1$. 
\Until {an action $a_t$ is chosen.}
\State Broadcast the chosen action $a_t$ to every agent (i.e., $a_{i,t}=a_t, \forall i \in [M]$), and receive  corresponding rewards $\{r_{i,t}\}_{i\in [M]}$. Adversarial agents can transmit arbitrary reward values.
\EndFor 
\end{algorithmic}
 \end{algorithm}
\textbf{Description of Algorithm \ref{algo:SupLin}.} At each time-step $t$, our proposed algorithm, namely Algorithm \ref{algo:SupLin}, scans through the set $\mc{A}$ of arms to determine a suitable action $a_t$. This scanning process (lines 4-10 of Algorithm  \ref{algo:SupLin}) is done at the server over $S$ phases. Corresponding to each phase $s\in [S]$, the server maintains a set $\psi^{(s)}_t$; the set $\psi^{(s)}_t$ stores all the time-steps in $[t-1]$ where an action is chosen in phase $s$ of the scanning process based on line 6 on Algorithm  \ref{algo:SupLin}. The scanning process itself relies on Algorithm \ref{algo:Base} as a sub-routine. Specifically, in each phase $s$, the server first invokes Algorithm \ref{algo:Base} to obtain a robust estimate $\hat{r}^{(s)}_{t,a}$ of $\langle \theta_*, x_{t,a} \rangle$ for  each arm $a\in \mc{A}_{s}$, along with an associated \textit{inflated} confidence width $w^{(s)}_{t,a}$ (line 5 of Algorithm  \ref{algo:SupLin}). If the confidence width is too large for a particular arm (as in line 6), then such an arm requires exploration and is accordingly chosen to be $a_t$. If, on the other hand, the confidence widths of all arms are sufficiently small (as in line 7), then $a_t$ is chosen to be the arm with the highest upper-confidence bound. Thus, we follow the principle of \textit{optimism in the face of uncertainty here, while exercising caution to account for the presence of adversaries (via the use of inflated confidence intervals)}. If the conditions in lines 6 and 7 both fail,  then the arms in $\mc{A}_s$ require further screening. Accordingly, we move to the next phase $s+1$, retaining only those arms that are sufficiently close to the optimal arm $a^*_t$; see line 8 of Algorithm \ref{algo:SupLin}. Our main innovation lies in (i) the construction of the robust arm-estimates in Algorithm \ref{algo:Base} that account for both statistical and adversarial behavior, and (ii) the careful use of such estimates in lines 6-8 of Algorithm \ref{algo:SupLin} to pick the action $a_t$. In particular, the construction of the robust arm-estimates in Algorithm \ref{algo:Base} is guided by the insights and results for the simpler setting in Section \ref{sec:algo} where the feature vectors do not change over time. 

We now state the main result of this section. 

\begin{mytheo}[label=thm:contextual]{Performance of Algorithm \ref{algo:SupLin}}{}
Suppose the corruption fraction satisfies $\alpha < 1/2$. Given any $\delta \in (0,1)$, Algorithm \ref{algo:SupLin} guarantees that with probability at least $1-\delta$, the following holds for each good agent $i\in [M]\setminus\mathcal{B}$: 
\begin{equation}
    \sum_{t=1}^{T} \langle \theta_*, x_{t,a^*_t}- x_{t,a_{i,t}} \rangle = \tilde{O}\left(\left(\alpha + \sqrt{\frac{1}{M}} \right) \sqrt{dT} \right).
\label{eqn:bound_SupLin}
\end{equation}
\end{mytheo}
\begin{proof}
Please see Appendix \ref{app:proof_contxt}.
\end{proof}
\textbf{Main Takeaways.} For the contextual bandit model considered here, the single-agent minimax optimal regret in the absence of adversaries is $\tilde{\Theta}(\sqrt{dT})$ \cite{auer2}. In light of the lower bound in Theorem \ref{thm:lower_bnd}, we see that Theorem \ref{thm:contextual} provides a near-optimal regret guarantee (as in Theorem \ref{thm:RCPLB}) for the collaborative contextual bandit setting with adversaries. 

\section{Experimental Results}
\label{app:experiments}
\begin{figure*}[htp] 
  \centering
  \begin{subfigure}[H]{0.4\linewidth}
    \centering
    \includegraphics[width=1.1\linewidth,height=0.9\linewidth]{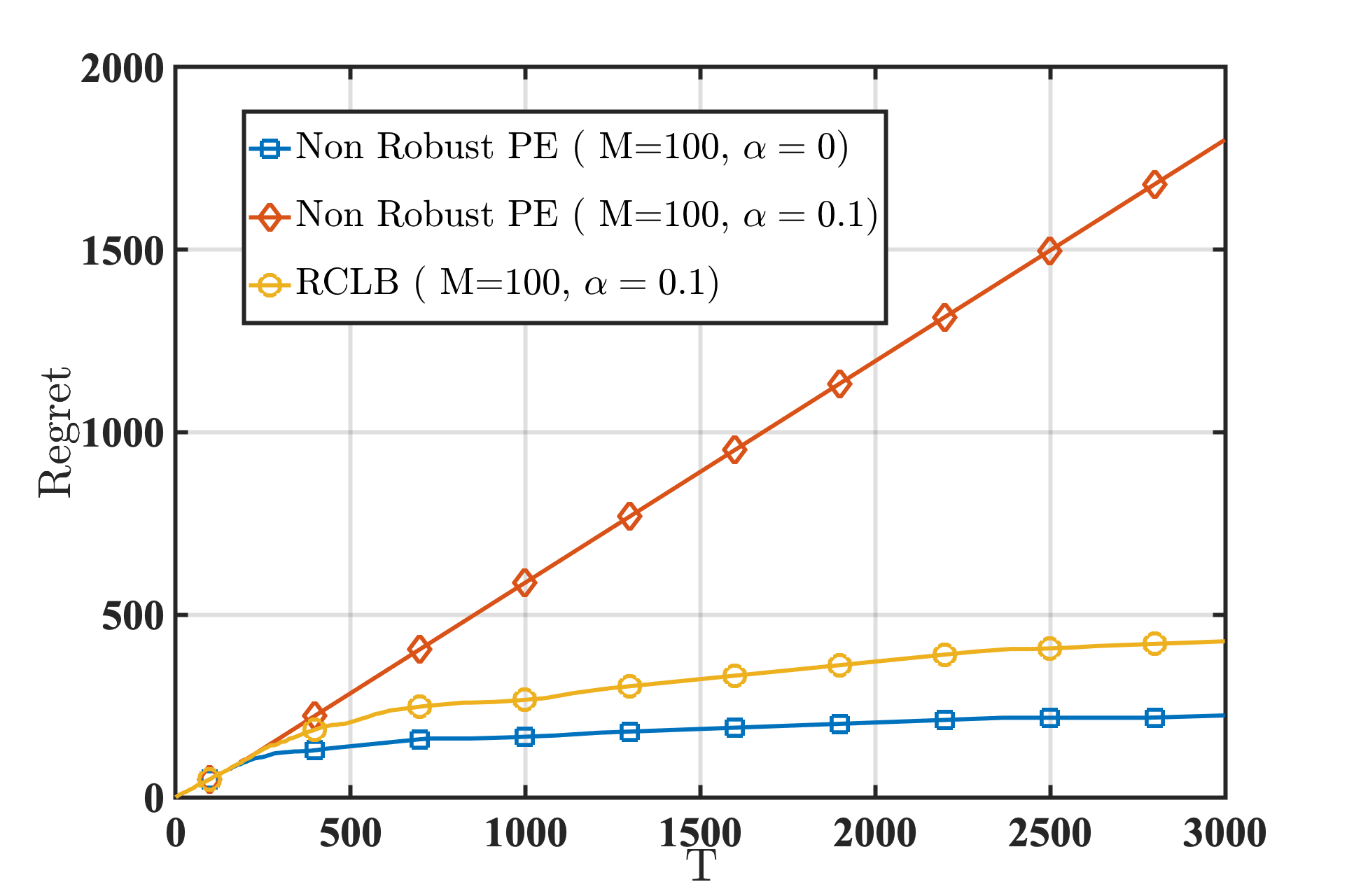} 
    \caption{} 
    \label{fig:simulation-1linear} 
  \end{subfigure}
  \quad
   \begin{subfigure}[H]{0.4\linewidth}
    \centering
    \includegraphics[width=1.1\linewidth,height=0.9\linewidth]{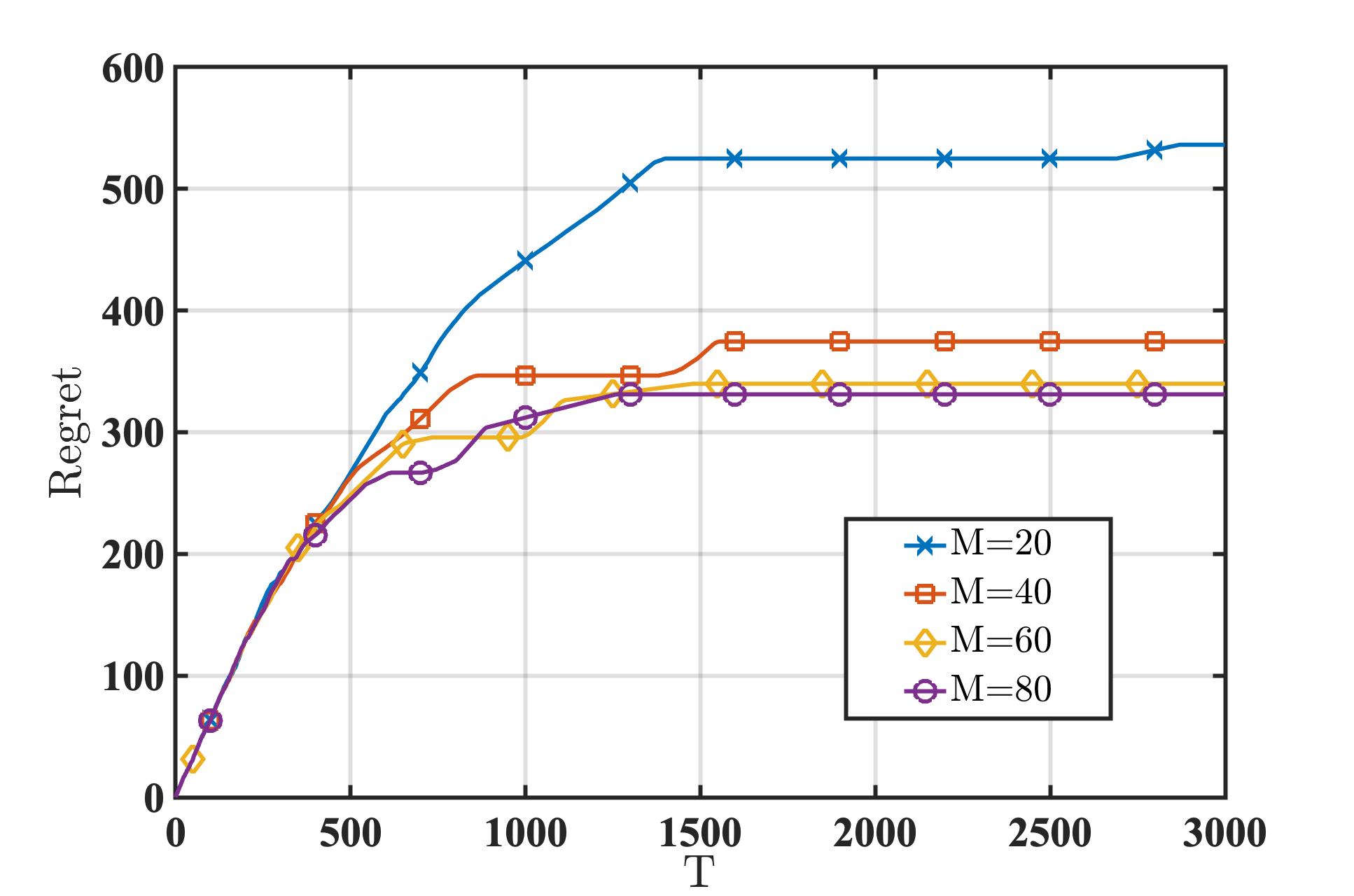}

    \caption{} 
    \label{fig:simulation-4linear} 
  \end{subfigure}
 \quad
   \begin{subfigure}[H]{0.4\linewidth}
    \centering
    \includegraphics[width=1.1\linewidth,height=0.9\linewidth]{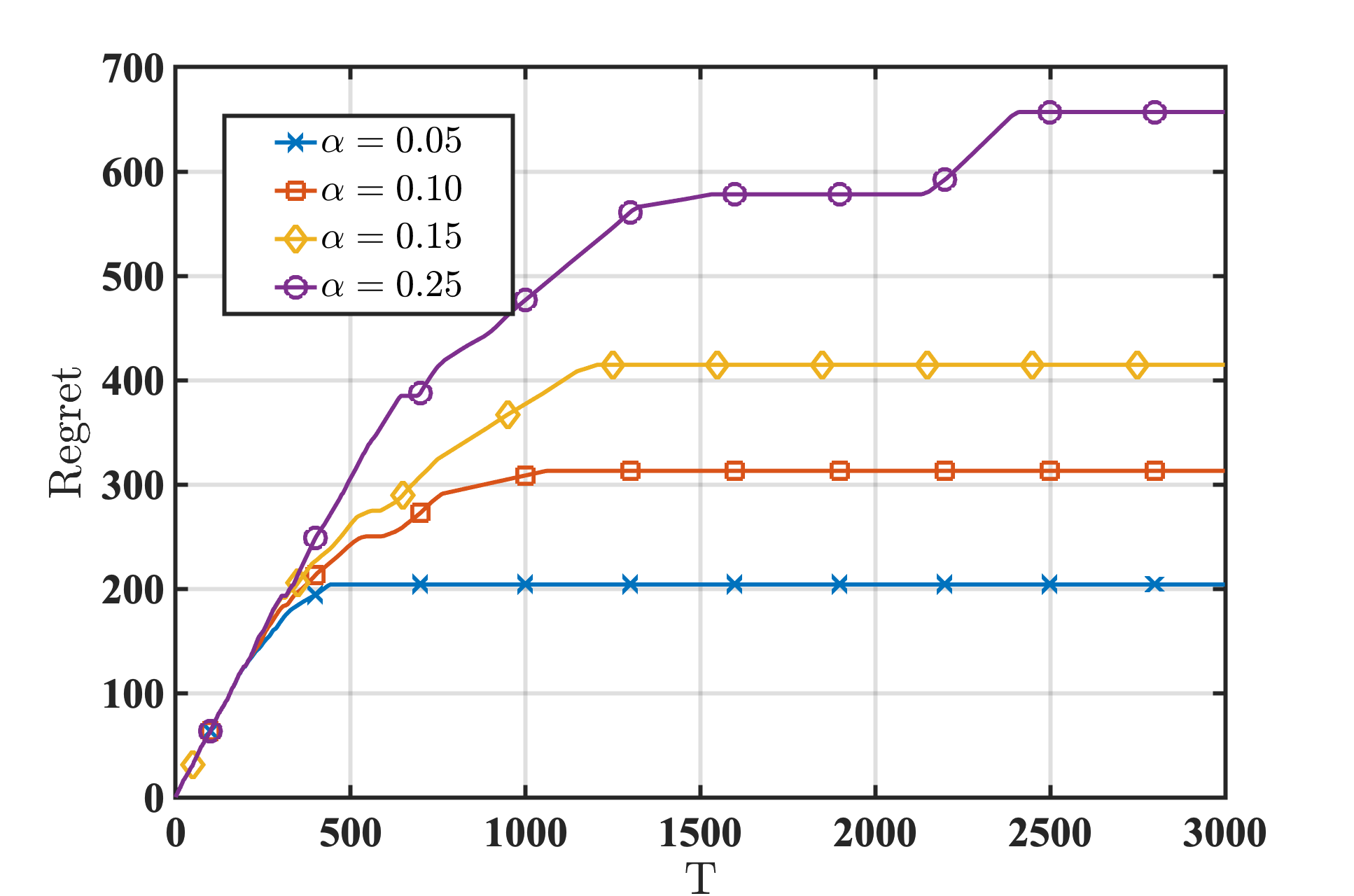}
    \caption{} 
    \label{fig:simulation-4linear2} 
  \end{subfigure} 
  \begin{subfigure}[H]{0.4\linewidth}
    \centering
    \includegraphics[width=1.1\linewidth,height=0.9\linewidth]{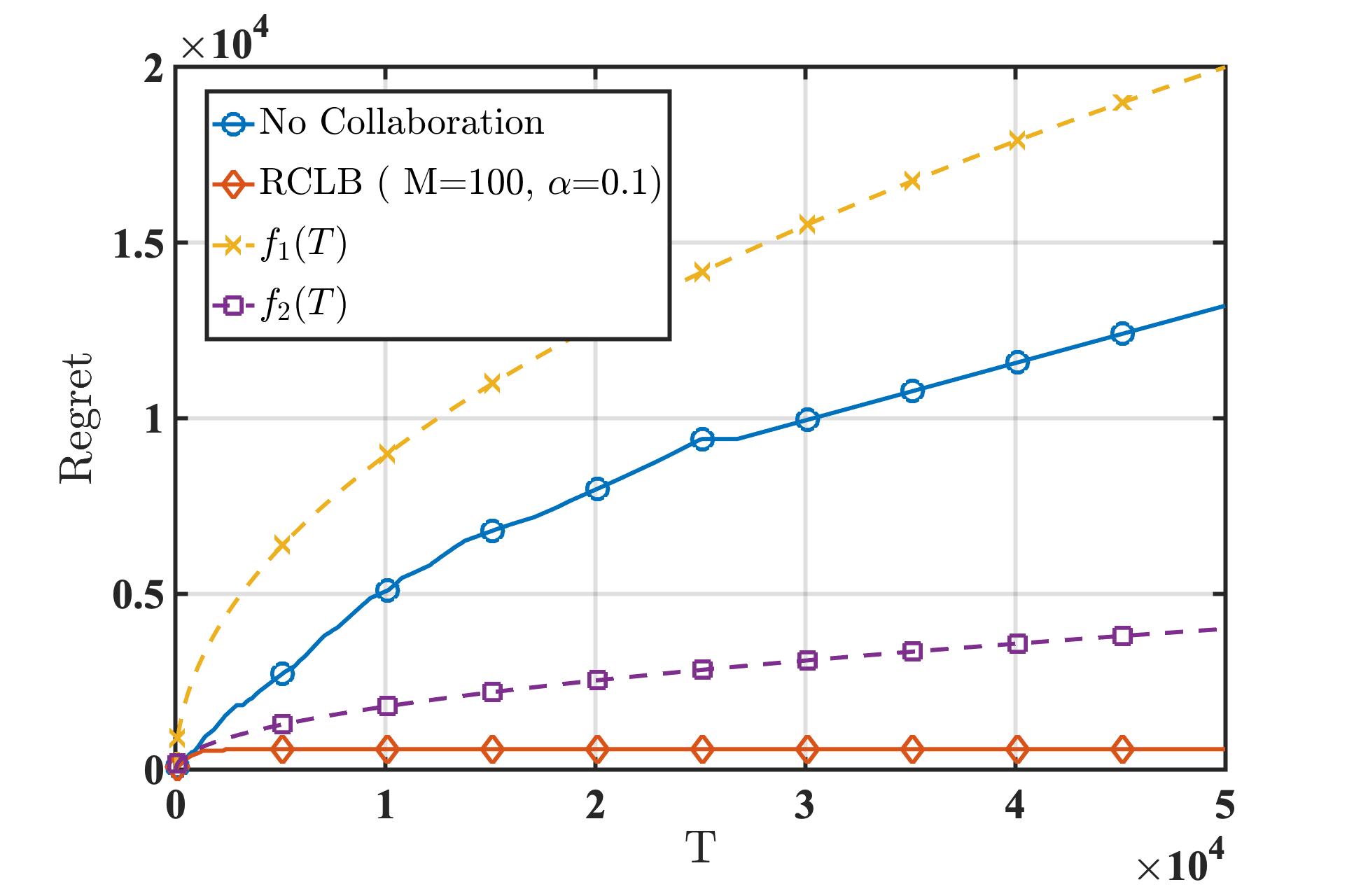} 
    \caption{} 
    \label{fig:simulation-2linear} 
  \end{subfigure} 
  \begin{subfigure}[H]{0.4\linewidth}
    \centering
    \includegraphics[width=1.1\linewidth,height=0.9\linewidth]{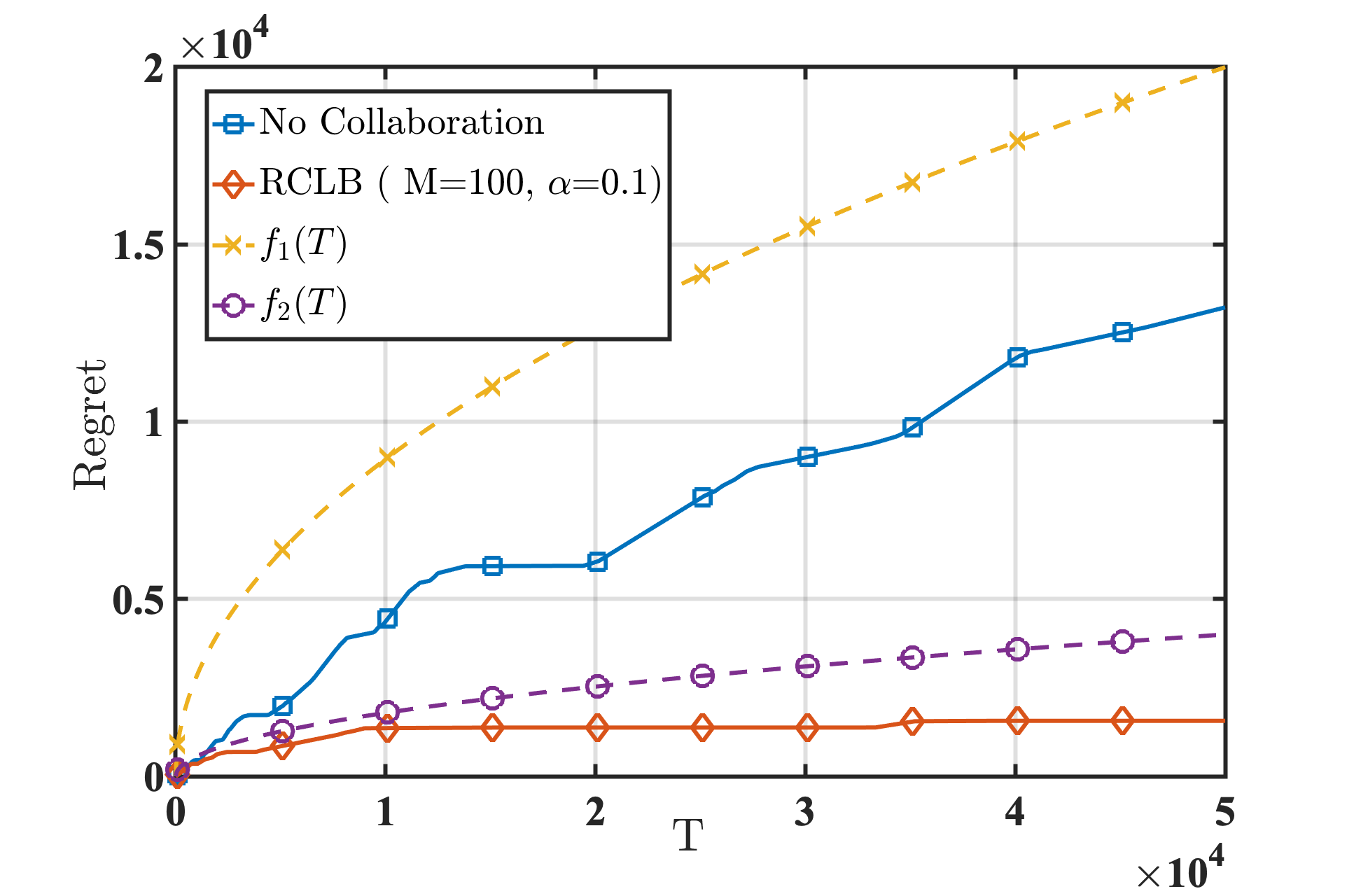} 
    \caption{} 
    \label{fig:simulation-3linear} 
  \end{subfigure}

  \vspace{-3mm}
  \caption{Plots of per-agent regret for the linear bandit experiment. (a) Comparison between \texttt{RCLB} and a vanilla non-robust phased elimination algorithm. 
  (b) Performance of \texttt{RCLB} for varying number of agents $M$, with $\alpha=0.1$. 
   (c) Performance of \texttt{RCLB} for varying corruption fraction $\alpha$, with $M=100$. For both (d) and (e), we set $\alpha=0.1$, $M=100$, and compare \texttt{RCLB} to a phased elimination algorithm where the agents do not collaborate. The arm-gap between the best and the second best arm is greater than $1/\sqrt{T}$ (resp., smaller than $1/\sqrt{T}$) in (d) (resp., (e)). For both (d) and (e), we also plot theoretical upper-bounds: $f_1(T)=40\sqrt{dT}$ and $f_2(T)=40(\alpha+\sqrt{{1}/{M}})\sqrt{dT}$.}
  
  \label{fig:simulation linear bandit} 
\end{figure*}
As far as we are aware, no existing algorithm provides theoretical guarantees for  multi-agent linear bandits with adversaries. Thus, in this section, we will provide various simulation results on synthetic data to corroborate the theory developed in our own work. We start by describing the experimental setup for the linear bandit setting.
\subsection{Experiments for the Linear Bandit Setting} 
\textbf{Linear Bandit Experimental Setup.} We generate 50 arms $a_1,a_2,\dots ,a_{50}\in \mathbb{R}^{d}$,  where $d=5$. Each arm $a_j, j\in [50]$, is generated by drawing each of the arm's coordinates i.i.d from the interval $[-{1}/{\sqrt{d}},{1}/{\sqrt{d}}]$. It thus follows that $\Vert a_j \Vert_2 \leq 1, \forall j \in [50]$. The model parameter $\theta_*$ is chosen to be a 5-dimensional vector with each entry equal to $1/{\sqrt{d}}$. The rewards are generated based on the observation model in Eq.~\eqref{eqn:Obs_model}. We now describe the attack model for the linear bandit setting. 

\textbf{Attack Model for Linear Bandit Setting.} The collective goal of the adversarial agents is to manipulate the server into selecting sub-optimal arms. To that end, each adversarial agent $i \in \mc{B}$ employs the simple strategy of reducing the rewards of the good arms and increasing the rewards of the bad arms. 
More precisely, in each epoch $\ell$,  upon pulling an arm $a \in \textrm{Supp}(\pi_{\ell})$ and observing the corresponding reward $y_a$, an adversarial agent $i$ does the following: for some scalar threshold $p >0$ and scalar bias term $\beta >0$, if $y_{a}>p\langle \theta_*, a_* \rangle$, then this reward is corrupted to $\tilde{y}_{a}=y_{a}-\beta$; and if
$y_{a}\leq p\langle \theta_*, a_*  \rangle$,  then the reward is corrupted to $\tilde{y}_{a}=y_{a}+\beta$. For this experiment, we fix $p=0.6$ and $\beta=5$. Agent $i\in\mc{B}$ then uses all the corrupted rewards in epoch $\ell$ to generate the local model estimate $\hat\theta_i^{(\ell)}$ that is transmitted to the server. 

\textbf{Discussion of Simulation Results.} Figure~\ref{fig:simulation linear bandit} provides a summary of our experimental results for the linear bandit setting. In Figure~\ref{fig:simulation linear bandit}(a), we compare our proposed algorithm \texttt{RCLB} to a vanilla distributed phased elimination (PE) algorithm that does not account for adversarial agents. Specifically, the latter is designed by replacing the median operation in line 8 of Algorithm \ref{algo:RCPLB} with a mean operation, and setting the threshold $\gamma_{\ell}$ in line 9 to be $\epsilon_{\ell}$. As can be seen from Figure~\ref{fig:simulation linear bandit}(a), in the absence of adversaries (i.e., when $\alpha=0$), the non-robust phased elimination algorithm guarantees sub-linear regret. However, even a small fraction $\alpha=0.1$ of adversaries causes the non-robust algorithm to incur linear regret. In contrast, \texttt{RCLB} continues to guarantee sub-linear regret bounds despite adversarial corruptions. Furthermore, the regret bounds of \texttt{RCLB} in the presence of a small fraction of adversarial agents is close to that of the non-robust phased elimination algorithm in the absence of adversaries. \textit{This goes on to establish the robustness of \texttt{RCLB}}.

Figure~\ref{fig:simulation linear bandit}(b) depicts the performance of \texttt{RCLB} for varying values $M\in\{20,40,60,80\}$ of the number of agents $M$; here, the corruption fraction $\alpha$ is fixed to $\alpha=0.1$. As we can see in this plot, increasing $M$ results in lower regret, \textit{indicating a clear benefit of collaboration despite the presence of adversaries}. In  Figure~\ref{fig:simulation linear bandit}(c), we plot the performance of \texttt{RCLB} for varying values $\alpha\in\{0.05,0.1,0.15,0.25\}$ of the corruption fraction $\alpha$; here, $M$ is set to a fixed value of $100$. As expected, increasing $\alpha$ leads to higher (albeit sub-linear) regret. Importantly, the trends observed in both Figure~\ref{fig:simulation linear bandit}(b) and Figure~\ref{fig:simulation linear bandit}(c) are consistent with the theoretical upper-bound of  ${O}((\alpha+{1}/{\sqrt{M}})\sqrt{dT})$ predicted by Theorem \ref{thm:RCPLB}. 

Note that in our setup, a trivial way to avoid adversarial corruption is for a good agent to avoid any interaction at all, and run a standard single-agent phased elimination algorithm. This would result in such an agent incurring $O(\sqrt{dT})$ regret. The purpose of 
Figure~\ref{fig:simulation linear bandit}(d)
and Figure~\ref{fig:simulation linear bandit}(e) is to drive home the point that $\texttt{RCLB}$ \textit{can lead to significant improvements over a trivial non-collaborative strategy.} To make this point precise, we compare  \texttt{RCLB} to a standard single-agent phased elimination algorithm that does not involve any collaboration. Both plots reveal that despite adversarial corruption, \texttt{RCLB} leads to considerably lower regret bounds as compared to the non-collaborative strategy. This highlights the importance of our proposed algorithm. We also plotted the theoretical upper bounds for the non-collaborative algorithm and $\texttt{RCLB}$ to verify the result in  Theorem~\ref{thm:RCPLB}. 

Finally, we study the effect of the ``hardness" of a given bandit instance, as captured by the arm-gap between the best and the second best arm. In Figure~\ref{fig:simulation linear bandit}(d), this gap is set to be larger than ${1}/{\sqrt{T}}$, resulting in smooth regret curves. In  Figure~\ref{fig:simulation linear bandit}(e), this gap is smaller than ${1}/{\sqrt{T}}$, causing the regret curves to exhibit sporadic jumps. 

\textbf{An Alternate Attack Model.} To further test the robustness of \texttt{RCLB}, we consider another attack model. In this attack, in each epoch $\ell$, every adversarial agent $i\in\mc{B}$ generates and transmits the following corrupted local model estimate to the server:

\begin{equation}
\hat{\theta}^{(\ell)}_i=-\frac{M}{|\mc{B}|}\theta_*-\frac{1}{| \mathcal{B}|}\sum_{j\in[M] \setminus \mathcal{B}}\hat{\theta}^{(\ell)}_j.
\label{eq:dif attack model}
\end{equation} 

\begin{figure}[t]
    \centering
    \includegraphics[width=0.5\linewidth]{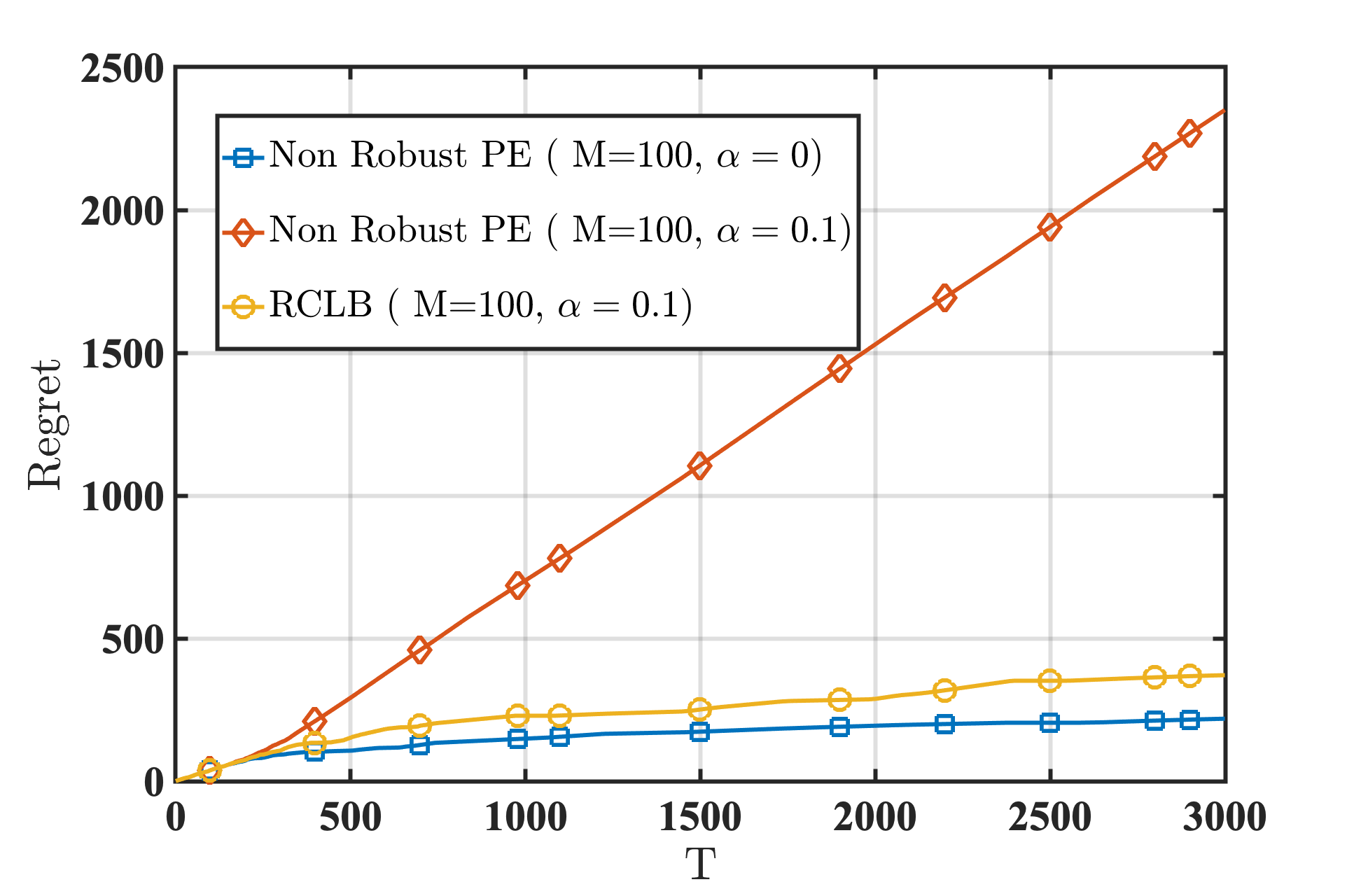}

    \caption{Performance of a vanilla non-robust distributed phased elimination algorithm vs.  \texttt{RCLB} for the attack model in  Eq.~\eqref{eq:dif attack model}.} 
    \label{fig:linear bandit other attack} 
  \end{figure}

The idea behind the above attack is to trick the server into thinking that the true model is $-\theta_*$, as opposed to $\theta_*$, by shifting the average of the agents' local model estimates towards $-\theta_*$. The motivation here is simple: if the server believes the true model to be $-\theta_*$, and attempts to maximize expected cumulative return, then it will be inclined to select the action/arm with the \textit{lowest mean payoff} w.r.t. the true model. As we can see from Figure~\ref{fig:linear bandit other attack}, the adversarial agents succeed in their cause when one employs a vanilla non-robust distributed phased elimination algorithm. However, our proposed approach $\texttt{RCLB}$ continues to remain immune to such attacks, and guarantees sub-linear regret as suggested by our theory. 

\subsection{Experiments for the Contextual Bandit Setting}

  \begin{figure*}[htp] 
  \centering
  \begin{subfigure}[H]{0.4\linewidth}
    \centering
    \includegraphics[width=1.1\linewidth,height=0.9\linewidth]{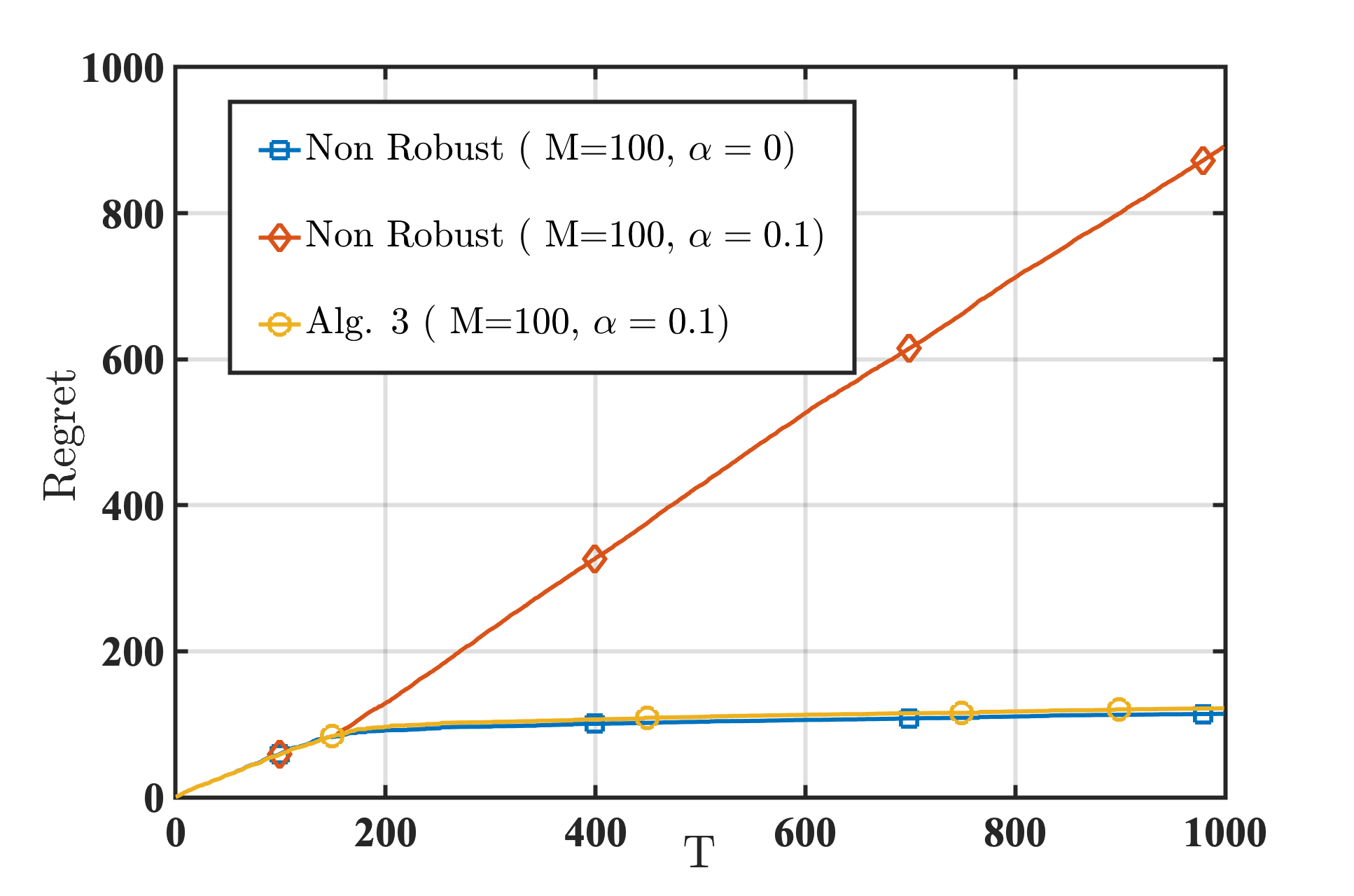} 
    \caption{} 
    \label{fig:simulation-1} 
  \end{subfigure}
  \quad
   \begin{subfigure}[H]{0.4\linewidth}
    \centering
    \includegraphics[width=1.1\linewidth,height=0.9\linewidth]{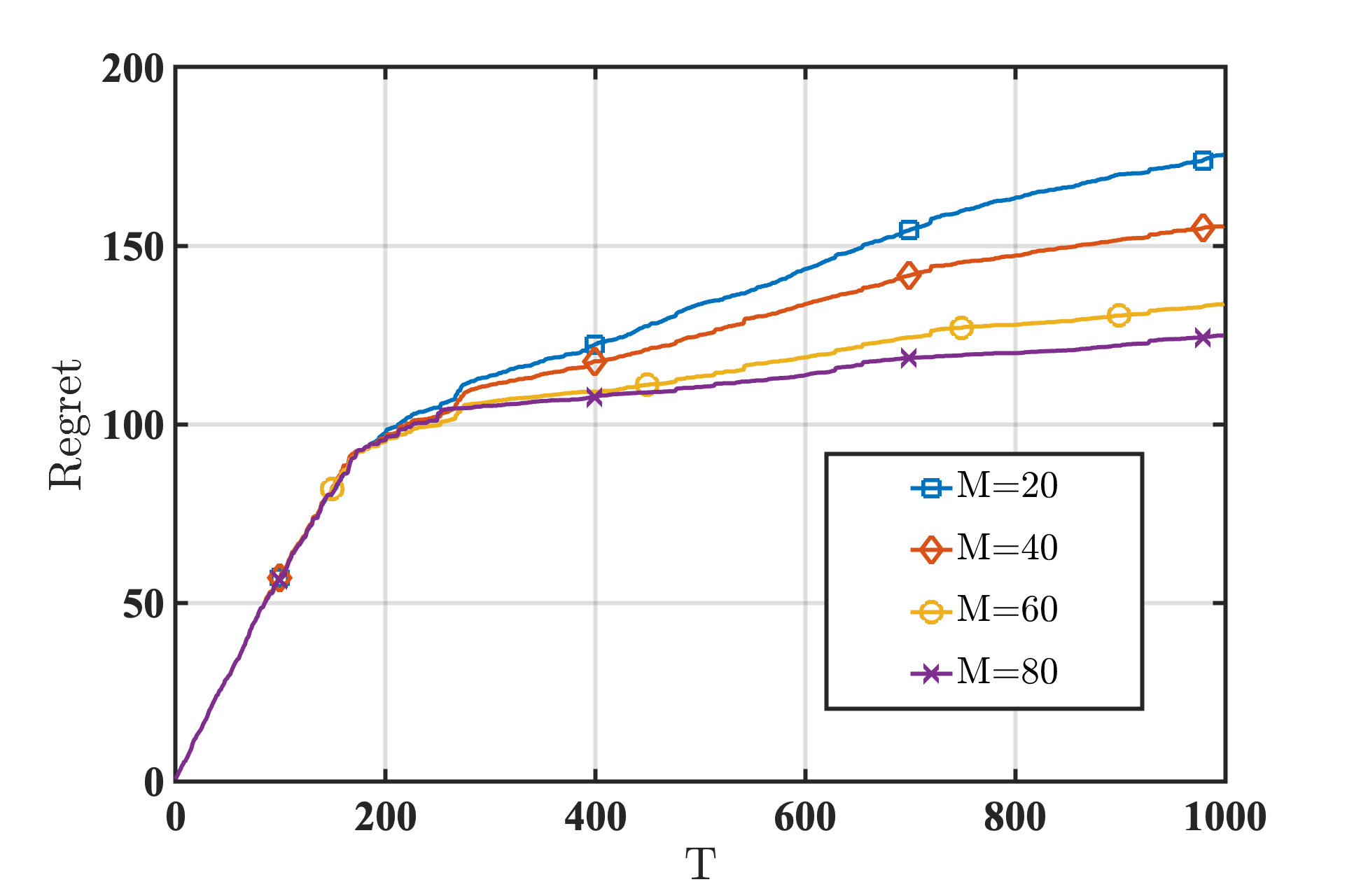}

    \caption{} 
    \label{fig:simulation-4.5} 
  \end{subfigure}
    \quad
   \begin{subfigure}[H]{0.4\linewidth}
    \centering
    \includegraphics[width=1.1\linewidth,height=0.9\linewidth]{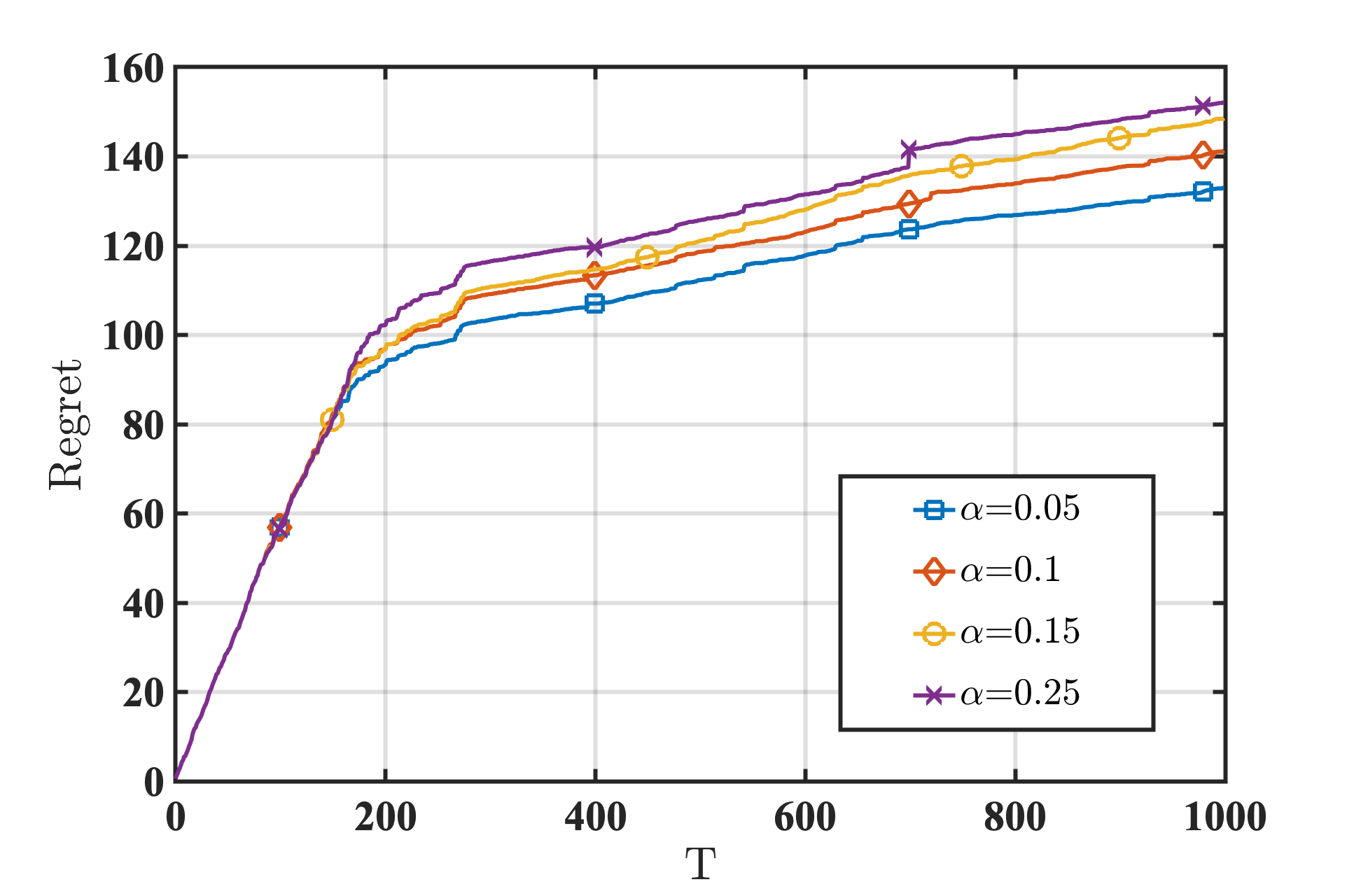}
    \caption{} 
    \label{fig:simulation-4} 
  \end{subfigure} 
  \quad
  \begin{subfigure}[H]{0.4\linewidth}
    \centering
    \includegraphics[width=1.1\linewidth,height=0.9\linewidth]{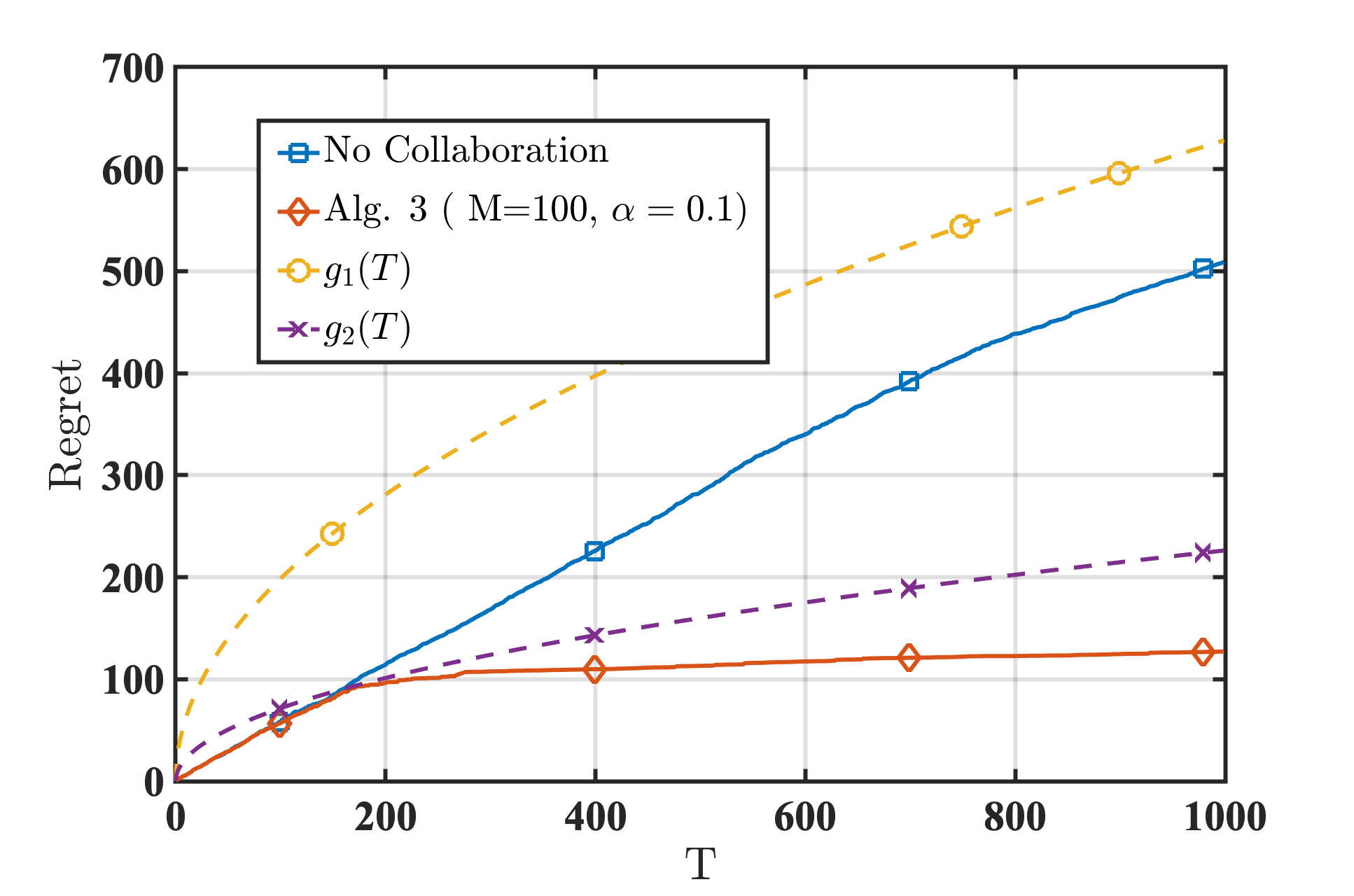} 
    \caption{} 
    \label{fig:simulation-2} 
  \end{subfigure}

  \caption{Plots of per-agent regret for the contextual bandit experiment. (a) Comparison between our proposed algorithm, namely Algorithm \ref{algo:SupLin}, and a vanilla non-robust distributed contextual bandit algorithm. 
  (b) Performance of Algorithm \ref{algo:SupLin} for varying number of agents $M$, with $\alpha=0.1$. 
   (c) Performance of Algorithm \ref{algo:SupLin} for varying corruption fraction $\alpha$, with $M=100$.
   (d) Comparison of Algorithm \ref{algo:SupLin} to a non-robust contextual bandit algorithm where the agents do not collaborate; here,  $\alpha=0.1$ and $M=100$. We also plotted theoretical upper bounds: $g_1(T)=3\sqrt{dT}$ and $g_2(T)=17\left(\alpha+\sqrt{\frac{1}{M}}\right)\sqrt{dT}$.}
  \label{fig:simulation contextual linear bandit} 
\end{figure*}
The goal of this section is to validate our proposed robust collaborative algorithm for the contextual bandit setting, namely Algorithm \ref{algo:SupLin}.

\textbf{Contextual Bandit Experimental Setup.} As in the linear bandit experiment, we set the number of arms $K$ to be $50$, the model dimension $d$ to be $5$, and the true parameter $\theta_*$ to be a $d$-dimensional vector with each entry equal to $1/\sqrt{d}$. At each time-step $t$, for each $a\in \mc{A}$, we generate the feature vector $x_{t,a}$ by drawing each of its entries i.i.d from the interval  $[-{1}/{\sqrt{d}},{1}/{\sqrt{d}}]$. The rewards are then generated based on the observation model in Eq.~\eqref{eqn:obs_context}.

\textbf{Attack Model for Contextual Bandit Setting.} We use an attack strategy similar in spirit to the first attack model for the linear bandit setting. Specifically, at each time-step $t$, each adversarial agent $i \in \mc{B}$ does the following: if $r_{i,t}>p\langle \theta_*, x_{t,a_t^{*}} \rangle$, then the attacker sets $\tilde{r}_{i,t}=r_{i,t}-\beta$; if $r_{i,t} < p \langle \theta_*, x_{t,a_t^{*}} \rangle$, then the attacker sets $\tilde{r}_{i,t}=r_{i,t}+\beta$. The corrupted reward $\tilde{r}_{i,t}$ is then sent to the server. In this experiment, we set $p=0.6$ and $\beta=5$.

\textbf{Discussion of Simulation Results.}  Figure~\ref{fig:simulation contextual linear bandit} illustrates the results for the contextual bandit experiment. In Figure~\ref{fig:simulation contextual linear bandit}(a), we compare our proposed algorithm, namely  Algorithm~\ref{algo:SupLin}, to a naive distributed implementation of Algorithm~\ref{algo:SupLin} that does not account for adversarial agents. Similar to what we observed in Figure~\ref{fig:simulation linear bandit}(a), while the non-robust algorithm incurs linear regret in the presence of adversaries, Algorithm \ref{algo:SupLin} continues to guarantee sub-linear regret bounds. The plots in Figures \ref{fig:simulation contextual linear bandit}(b)-(d) are analogous to the ones in Figures \ref{fig:simulation linear bandit}(b)-(d). In short, these plots once again indicate a clear benefit of collaboration (for small $\alpha$) in the presence of adversarial agents, thereby highlighting the importance of Algorithm \ref{algo:SupLin}, and validating Theorem \ref{thm:contextual}. 
 
\section{Conclusion and Future Work}
In this paper, we studied for the first time the problem of tackling adversarial agents in the context of a collaborative linear stochastic bandit setting. We introduced a simple, robust phased elimination algorithm called \texttt{RCLB} for this purpose, and proved that it guarantees sub-linear regret. In particular, the main message conveyed by our work is that when the fraction of adversarial agents is small, \texttt{RCLB} can lead to significant benefits of collaboration. We also proved a fundamental lower bound, thereby providing the first set of tight, near-optimal regret guarantees for our problem of interest. Finally, we showed how our algorithmic ideas and results can be significantly extended to more general structured bandit settings. 

There are several interesting future directions to pursue. First, we note that our lower-bound in Theorem \ref{thm:lower_bnd} is constructed for instances where the gap between the best and the second best arm is very small, namely, of order $\alpha/{\sqrt{T}}$. Such hard instances naturally favor the adversarial agents. However, given \textit{any} instance, what is the worst effect the adversarial agents can induce? This leads to the open problem of deriving \textit{instance-specific} lower bounds for the setting considered in this paper. Second, we considered a homogeneous setting where all agents interact with the \textit{same} bandit instance. However, in large-scale applications such as federated learning, accounting for heterogeneity is an emerging challenge \cite{sahu,khaled2,li,scaffold,mitralin}. Thus, it would be interesting to see how our results can be extended to scenarios where the agents interact with similar, but not necessarily identical environments. Third, the results in this paper do not cover the case where the action set is allowed to be infinite (but bounded) \textit{and} time-varying; developing robust algorithms for this case is still open.  Finally, we plan to investigate the more general reinforcement learning setting as future work.

\bibliographystyle{unsrt} 
\bibliography{refs}
\appendix 
\newpage
\section{Analysis of \texttt{RCLB}: Proof of Theorem \ref{thm:RCPLB}}
\label{app:proof_RCPLB}
In this section, we will prove Theorem \ref{thm:RCPLB}. We start with a standard  result from robust statistics on the guarantees afforded by the median operator for robust mean estimation of univariate Gaussian random variables; see, for instance, \cite{lai}. 

\begin{mylem}[label=lemma:median]{}{}
Consider a set $\mathcal{S}=\{x_1, \ldots, x_M\}$ of $M$ samples partitioned as $\mathcal{S}=\mathcal{S}_g \cup \mathcal{S}_b$, such that (i) all the samples in $\mathcal{S}_g$ are drawn i.i.d. from $\mathcal{N}(\mu,\sigma^2)$, where $\mu, \sigma^2 \in \mathbb{R}$; (ii) the samples in $\mathcal{S}_b$ are chosen by an adversary, and can be arbitrary; and (iii) $|\mathcal{S}_b| < \alpha |\mathcal{S}|$, where $\alpha < 1/2$. Let $\hat{\mu}=\texttt{Median}\left(\{x_i\}, i \in [M]\right).$ Given any $\delta \in (0,1)$, we then have that with probability at least $1-\delta$, 
\begin{equation}
    |\hat{\mu}-\mu| \leq C \left(\alpha + \sqrt{ \frac{ \log(\frac{1}{\delta})}{M}} \right) \sigma,
\label{eqn:Median_bound}
\end{equation}
where $C$ is a suitably large universal constant. 
\end{mylem}

The next key lemma - a restatement of Lemma \ref{lemma:rob_conf_main} in the main body of the paper - informs us about the quality of the robust mean payoffs computed in line 8 of Algorithm \ref{algo:RCPLB}. Before proceeding to prove this result, we define by $\mc{F}_{\ell}$ the $\sigma$-algebra generated by all the actions and rewards up to the beginning of epoch $\ell$. 

\begin{mylem}[label=lemma:robust_payoffs]{Robust Confidence Intervals for \texttt{RCLB}}{}
Fix any epoch $\ell$. For each active arm $b \in \mc{A}_{\ell}$, the following holds with probability at least $1-\delta_{\ell}$:
\begin{equation}
    |\mu^{(\ell)}_b - \langle \theta_*, b \rangle| \leq \gamma_{\ell}, \hspace{2mm} \textrm{where} \hspace{2mm} \gamma_{\ell}=\sqrt{2} C \left(1+\alpha\sqrt{M} \right) \epsilon_{\ell},
\end{equation}
where $C$ is as in Lemma \ref{lemma:median}. 
\end{mylem}
\begin{proof}
Fix an epoch $\ell$, an active arm $b\in\mc{A}_{\ell}$, and a good agent $i\in [M]\setminus \mathcal{B}$. We start by analyzing the statistics of the quantity $\langle \hat{\theta}^{(\ell)}_i,b \rangle$. From the definition of $\hat{\theta}^{(\ell)}_i$ and $\tilde{V}_{\ell}$ in Eq.\eqref{eqn:local_estimate}, we have
\begin{equation}
    \begin{aligned}
    \hat{\theta}^{(\ell)}_i &= \tilde{V}^{-1}_{\ell} Y_{i,\ell} \\
    & = \tilde{V}^{-1}_{\ell} \left( \sum_{a\in\textrm{Supp}(\pi_{\ell})} \hspace{-5mm}   m^{(\ell)}_a r^{(\ell)}_{i,a} a  \right)\\
    & = \tilde{V}^{-1}_{\ell} \left(\sum_{a\in\textrm{Supp}(\pi_{\ell})} \hspace{-5mm}   m^{(\ell)}_a \left( \langle  \theta_*, a \rangle + \bar{\eta}^{(\ell)}_{i,a} \right) a \right) \\
    & = \theta_* + \tilde{V}^{-1}_{\ell} \left( \sum_{a\in\textrm{Supp}(\pi_{\ell})} \hspace{-5mm}   m^{(\ell)}_a \bar{\eta}^{(\ell)}_{i,a} a  \right). 
    \end{aligned}
\label{eqn:unrolled_local_est}
\end{equation}
For the third equality above, we used the observation model \eqref{eqn:Obs_model}, and denoted by $\bar{\eta}^{(\ell)}_{i,a}$ the average of the noise terms associated with the rewards observed by agent $i$ during phase $\ell$ for arm $a$. From \eqref{eqn:unrolled_local_est}, we then have
\begin{equation}
  \langle \hat{\theta}^{(\ell)}_i, b \rangle = \langle \theta_*, b \rangle +  \sum_{a\in\textrm{Supp}(\pi_{\ell})} \hspace{-5mm}   m^{(\ell)}_a \bar{\eta}^{(\ell)}_{i,a} \langle \tilde{V}^{-1}_{\ell} a, b \rangle.
\end{equation}
Now conditioned on $\mc{F}_{\ell}$, the only randomness in the above equation corresponds to the noise terms $\{\bar{\eta}^{(\ell)}_{i,a}\}_{ a\in\textrm{Supp}(\pi_{\ell})}.$ Furthermore, based on our noise model, it is clear that $\bar{\eta}^{(\ell)}_{i,a} \sim \mc{N}(0,1/m^{(\ell)}_a)$ for each $a \in \textrm{Supp}(\pi_{\ell})$. It then follows that 
$$ \mathbb{E} \left[ \langle \hat{\theta}^{(\ell)}_i, b \rangle | \mc{F}_{\ell} \right] = \langle \theta_*, b \rangle . $$ 
We also have
\begin{equation}
    \begin{aligned}
    \mathbb{E} \left[ \left(\langle \hat{\theta}^{(\ell)}_i - \theta_*, b \rangle \right)^2| \mc{F}_{\ell} \right] & =  \sum_{a\in\textrm{Supp}(\pi_{\ell})} \hspace{-5mm}   m^{(\ell)}_a \left( \langle \tilde{V}^{-1}_{\ell} a, b \rangle \right)^2 \\
    & =  b' \tilde{V}^{-1}_{\ell} \left(\sum_{a\in\textrm{Supp}(\pi_{\ell})} \hspace{-5mm}   m^{(\ell)}_a a a' \right) \tilde{V}^{-1}_{\ell} b \\
    & =  {\Vert b \Vert}^2_{\tilde{V}^{-1}_{\ell}},
    \end{aligned}
\end{equation}
where we used the fact that the noise terms are independent across arms. We conclude that conditioned on $\mathcal{F}_{\ell}$, 
$$ \langle \hat{\theta}^{(\ell)}_i, b \rangle \sim \mathcal{N}\left(\langle \theta_*, b \rangle, {\Vert b \Vert}^2_{\tilde{V}^{-1}_{\ell}} \right). $$
In each epoch $\ell$, the server has access to a set $\mc{S}^{(\ell)}_b=\{ \langle \hat{\theta}^{(\ell)}_i, b \rangle \}_{i\in [M]}$, where the samples corresponding to agents in $[M]\setminus\mc{B}$ are independent and identically distributed as per the distribution above. Moreover, at most $\alpha \in [0,1/2)$ fraction of the samples in $\mc{S}^{(\ell)}_b$ are corrupted. Recalling that $\mu^{(\ell)}_{b}=\texttt{Median}\left( \{\langle \hat{\theta}^{(\ell)}_i, b \rangle, i \in [M]\} \right)$, and using Lemma \ref{lemma:median}, we immediately observe that conditioned on $\mc{F}_{\ell}$, with probability at least $1-\delta_{\ell}$,
\begin{equation}
    |\mu^{(\ell)}_{b}-\langle \theta_*, b \rangle | \leq C \left(\alpha + \sqrt{ \frac{ \log(\frac{1}{\delta_{\ell}})}{M}} \right) {\Vert b \Vert}_{\tilde{V}^{-1}_{\ell}} .
\label{eqn:rc_bnd1}
\end{equation}
We now proceed to bound the term ${\Vert b \Vert}_{\tilde{V}^{-1}_{\ell}}$. To that end, let us start by noting that
\begin{equation}
\begin{aligned}
    \tilde{V}_{\ell} &= \sum_{a\in \textrm{Supp}(\pi_{\ell})} m^{(\ell)}_a a a' \\
    & = \sum_{a\in \textrm{Supp}(\pi_{\ell})} \ceil*{ \frac{T^{(\ell)}_a}{M}} a a' \\
    & \succcurlyeq \frac{1}{M} \sum_{a\in \textrm{Supp}(\pi_{\ell})} \hspace{-5mm} {T^{(\ell)}_a} a a' \\
    & \succcurlyeq \frac{d}{M \epsilon^2_{\ell}} \log\left(\frac{1}{\delta_{\ell}}\right) \sum_{a\in \textrm{Supp}(\pi_{\ell})}  \hspace{-5mm} \pi_{\ell}(a) a a' \\
   & = \frac{d}{M \epsilon^2_{\ell}} \log\left(\frac{1}{\delta_{\ell}}\right) \sum_{a\in \mc{A}_{\ell}}  \pi_{\ell}(a) a a' \\
   & = \frac{d}{M \epsilon^2_{\ell}} \log\left(\frac{1}{\delta_{\ell}}\right) V_{\ell}(\pi_{\ell}).
\end{aligned}
\end{equation}
Thus, we have
$$ \tilde{V}^{-1}_{\ell} \preccurlyeq   \frac{M \epsilon^2_{\ell}}{d \log\left(\frac{1}{\delta_{\ell}}\right)}    V^{-1}_{\ell}(\pi_{\ell}).$$
Using the above bound, we proceed as follows.
\begin{equation}
\begin{aligned}
    {\Vert b \Vert}_{\tilde{V}^{-1}_{\ell}} &= \sqrt{b' \tilde{V}^{-1}_{\ell} b} \\
    & \leq \epsilon_{\ell} \sqrt{\frac{M}{d \log\left(\frac{1}{\delta_{\ell}}\right)}} \sqrt{b' V^{-1}_{\ell}(\pi_{\ell}) b} \\
   & \leq \epsilon_{\ell} \sqrt{\frac{M}{d \log\left(\frac{1}{\delta_{\ell}}\right)}} \sqrt{ \max_{a\in\mc{A}_{\ell}} {\Vert a \Vert}^2_{V^{-1}_{\ell}(\pi_{\ell})}} \\
   & \overset{(a)}= \epsilon_{\ell} \sqrt{\frac{M}{d \log\left(\frac{1}{\delta_{\ell}}\right)}} \sqrt{ g_{\ell}(\pi_{\ell})}\\
   & \overset{(b)}\leq \epsilon_{\ell} \sqrt{\frac{2M}{ \log\left(\frac{1}{\delta_{\ell}}\right)}}. 
\end{aligned}
\label{eqn:rc_bnd2}
\end{equation}
In the above steps, we used the definition of $g_{\ell}(\pi_{\ell})$ for (a); for (b), we used the fact that based on the approximate G-optimal design problem solved by the server in line 1 of \texttt{RCLB}, $g_{\ell}(\pi_{\ell}) \leq 2d$. Plugging the bound from \eqref{eqn:rc_bnd2} into \eqref{eqn:rc_bnd1}, and using the fact that $\log\left(\frac{1}{\delta_{\ell}}\right) \geq 1$, we have that
\begin{equation}
 \mathbb{P}\left(|\mu^{(\ell)}_{b}-\langle \theta_*, b \rangle |\geq \gamma_{\ell}| \mc{F}_{\ell} \right)\leq \delta_{\ell}.
\label{eqn:err_bnd}
\end{equation}
Consider the following event $\mc{E}_{\ell} \triangleq \{ |\mu^{(\ell)}_{b}-\langle \theta_*, b \rangle |\geq \gamma_{\ell} \}.$ Now observe that 
\begin{equation}
\begin{aligned}
    \mathbb{P}( \mc{E}_{\ell} ) &= \mathbb{E}[\mathbf{1}_{\mc{E}_{\ell}}] \\
   & = \mathbb{E}\left[ \mathbb{E}[\mathbf{1}_{\mc{E}_{\ell}}| \mc{F}_{\ell}] \right] \\
   & = \mathbb{E}\left[ \mathbb{P}(\mc{E}_{\ell}|\mc{F}_{\ell}) \right] \\
   & \leq \delta_{\ell},
\end{aligned}    
\label{eqn:filtrations}
\end{equation}
where we used $\mathbf{1}_{\mc{E}_{\ell}}$ to denote an indicator random variable associated with the event $\mc{E}_{\ell}$; also, for the last line, we used \eqref{eqn:err_bnd}. 
\end{proof}
In the following two results, we use the robust confidence intervals from Lemma \ref{lemma:robust_payoffs} to construct clean events that hold with high probability on which (i) the optimal arm $a_*$ is never eliminated (Lemma \ref{lemma:good_arm_RCPLB}); and (ii) any arm retained in epoch $\ell$ contributes at most $O(\gamma_{\ell})$ regret in each time-step within epoch $\ell$ (Lemma \ref{lemma:bad_arm_RCPLB}). To proceed, for each $a\in \mc{A}$, define the arm-gap $\Delta_a = \langle \theta_*, a_*-a \rangle$. 
\begin{mylem}[label=lemma:good_arm_RCPLB]{}{}
Define the event $\mc{G}_1 \triangleq \{a_* \in \mc{A}_{\ell}, \forall \ell \in [L]\}$, where $L$ is the total number of epochs. It then holds that $\mathbb{P}(\mc{G}_1) \geq 1-4\bar{\delta}$. 
\end{mylem}
\begin{proof}
Based on the arm-elimination criterion in line 9 of Algorithm \ref{algo:RCPLB}, it follows that $\{a_* \in \mc{A}_{\ell}, a_* \notin \mc{A}_{\ell+1}\} \implies \{\exists b \in \mc{A}_{\ell}: \mu^{(\ell)}_b - \mu^{(\ell)}_{a_*} > 2\gamma_{\ell}\}$. Now for any fixed $b\in\mc{A}_{\ell}$, we have
\begin{equation}
    \begin{aligned}
& \mu^{(\ell)}_b - \mu^{(\ell)}_{a_*} > 2\gamma_{\ell} \\
& \implies \left(\mu^{(\ell)}_b - \langle \theta_*,b \rangle \right) + \left( \langle \theta_*,a_* \rangle - \mu^{(\ell)}_{a_*} \right) > 2 \gamma_{\ell} + \Delta_b \\
& \implies \left(\mu^{(\ell)}_b - \langle \theta_*,b \rangle \right) + \left( \langle \theta_*,a_* \rangle - \mu^{(\ell)}_{a_*} \right) > 2 \gamma_{\ell}, \\
    \end{aligned}
\end{equation}
where for the second step, we used the fact that $\Delta_b \geq 0$. Thus, the event $\{\mu^{(\ell)}_b - \mu^{(\ell)}_{a_*} > 2\gamma_{\ell}\}$ implies the occurrence of either $\{ \mu^{(\ell)}_b - \langle \theta_*,b \rangle > \gamma_{\ell} \}$ or $\{ \langle \theta_*,a_* \rangle - \mu^{(\ell)}_{a_*} > \gamma_{\ell}\}$. From Lemma \ref{lemma:robust_payoffs}, we further know that the probability of each of these latter events is at most $\delta_{\ell}$. Putting these pieces together, and using an union bound, we have
\begin{equation}
    \begin{aligned}
    \mathbb{P}(\mc{G}^{c}_{1})&\leq \sum_{\ell\in [L]} \mathbb{P}(a^{*}\in \mathcal{A}_{\ell},a^{*}\not\in \mathcal{A}_{\ell+1})\\
   & \leq \sum_{\ell\in [L]} \mathbb{P}(\exists b \in \mc{A}_{\ell}: \mu^{(\ell)}_b - \mu^{(\ell)}_{a_*} > 2\gamma_{\ell}) \\ 
   &\leq 2K\sum_{\ell\in [L]} \delta_{\ell} 
    \\& = 2K\sum_{\ell\in [L]} \frac{\bar{\delta}}{K{\ell}^2}
    \\&\leq 2\sum_{\ell=1}^{\infty} \frac{\bar{\delta}}{\ell^2}
    \\&\leq 2\bar{\delta}\int_{x=1}^{\infty} \frac{1}{x^2}\, dx\leq 4\bar{\delta}. 
    \end{aligned}
\end{equation}
This completes the proof. 
\end{proof}
In our next result, we work towards bounding the regret incurred from playing each active arm in a given epoch. 
\begin{mylem}[label=lemma:bad_arm_RCPLB]{}{}  Consider any arm $a\in\mc{A}\setminus \{a_*\}$.  Let $\ell_{a}$ be defined as $\ell_{a} \triangleq \min\{\ell: \gamma_{\ell} < \frac{\Delta_a}{4}\}$. It then holds that $\mathbb{P}(a \in \mc{A}_{\ell_a+1}) \leq 6 \bar{\delta}.$ 
\end{mylem}
\begin{proof}
Let us start by observing that
\begin{equation}
\begin{aligned}
\mathbb{P}(a\in \mathcal{A}_{\ell_a+1})&=\mathbb{P}(a\in \mathcal{A}_{\ell_a+1},a^{*}\in \mathcal{A}_{\ell_a})+\mathbb{P}(a\in \mathcal{A}_{\ell_a+1},a^{*}\not\in \mathcal{A}_{\ell_a})
\\&\leq \mathbb{P}(a\in \mathcal{A}_{\ell_a+1},a^{*}\in \mathcal{A}_{\ell_a})+\mathbb{P}(a^{*}\not\in \mathcal{A}_{\ell_a})
\\&\leq \mathbb{P}(a\in \mathcal{A}_{\ell_a},a\in \mathcal{A}_{\ell_a+1},a^{*}\in \mathcal{A}_{\ell_a})+4\bar{\delta},
\end{aligned} 
\label{eqn:prob_bnd_1}
\end{equation}
where for the last step, we used the fact that $\{a \in \mathcal{A}_{\ell_a+1} \} \implies \{a \in \mathcal{A}_{\ell_a} \}$, and Lemma \ref{lemma:good_arm_RCPLB}. 
Now, to bound $\mathbb{P}(a\in \mathcal{A}_{\ell_a},a\in \mathcal{A}_{\ell_a+1},a^{*}\in \mathcal{A}_{\ell_a})$, we note based on line 9 of \texttt{RCLB} that 
\begin{equation}
\begin{aligned}
    \mathbb{P}(a\in \mathcal{A}_{\ell_a},a\in \mathcal{A}_{\ell_a+1},a^{*}\in \mathcal{A}_{\ell_a}) &\leq \mathbb{P}\left( \max_{b\in\mc{A}_{\ell_a}} \mu^{(\ell_a)}_b - \mu^{(\ell_a)}_a \leq 2 \gamma_{\ell_a}     \right) \\
& \leq \mathbb{P}\left( \mu^{(\ell_a)}_{a_*} - \mu^{(\ell_a)}_a \leq 2 \gamma_{\ell_a} \right)\\
& \leq \mathbb{P}\left( \left(\mu^{(\ell_a)}_a - \langle \theta_*,a \rangle \right) + \left( \langle \theta_*,a_* \rangle - \mu^{(\ell_a)}_{a_*} \right) > \Delta_a - 2 \gamma_{\ell_a} \right) \\
& \leq \mathbb{P}\left( \mu^{(\ell_a)}_a - \langle \theta_*,a \rangle > \gamma_{\ell_a} \right) + \mathbb{P}\left(  \langle \theta_*,a_* \rangle - \mu^{(\ell_a)}_{a_*}  > \gamma_{\ell_a} \right) \\
& \leq 2 \delta_{\ell_a},
\end{aligned}
\label{eqn:prob_bnd_2}
\end{equation}
where we used $\Delta_a > 4 \gamma_{\ell_a}$ for the second last step, and Lemma \ref{lemma:robust_payoffs} for the last step. Noting that $\delta_{\ell_a} \leq \bar{\delta}$, and combining the bounds in equations \eqref{eqn:prob_bnd_1} and \eqref{eqn:prob_bnd_2} leads to the claim of the lemma. 
\end{proof}
We are now in place to prove Theorem \ref{thm:RCPLB}. 
\begin{proof} (\textbf{Proof of Theorem \ref{thm:RCPLB}})
We start by constructing an appropriate clean event $\mc{E}$ for our subsequent analysis. Accordingly, let us define:
\begin{equation}
    \mc{E} = \{a_* \in \mc{A}_{\ell}, \forall \ell \in [L]\} \bigcap \{ \cap_{a\in \mc{A}\setminus \{a_*\}} \{a \notin \mc{A}_{\ell_a+1}\} \}. 
\end{equation}
Based on Lemmas \ref{lemma:good_arm_RCPLB} and \ref{lemma:bad_arm_RCPLB}, we then have
\begin{equation}
    \begin{aligned}
    \mathbb{P}(\mc{E}^{c}) & \leq 4 \bar{\delta} + \sum_{a \in \mc{A}\setminus\{a_*\}} \hspace{-3mm} \mathbb{P}(a \in \mc{A}_{\ell_a+1})\\
    & \leq 4 \bar{\delta} + 6 \bar{\delta} K\\
    & \leq 10 \bar{\delta} K =\delta,
    \end{aligned}
\end{equation}
as per the choice of $\bar{\delta}$ in Theorem \ref{thm:RCPLB}. Thus, $\mathbb{P}(\mc{E}) \geq 1-\delta$. Throughout the rest of the proof, we will condition on the clean event $\mc{E}$. Based on the definition of the event $\mc{E}$, it is easy to see that for any epoch $\ell \in [L]$, 
$a \in \mc{A}_{\ell} \implies \Delta_a \leq 8 \gamma_{\ell}.$ Using this key fact, we now proceed to bound the regret of any good agent $i\in [M]\setminus\mc{B}$. 
\begin{equation}
\begin{aligned}
\sum_{t=1}^{T}\langle\theta_*,a_{*}-a_{i,t}\rangle
&=\sum_{\ell=1}^{L}\sum_{a\in \textrm{Supp}(\pi_{\ell}) }m^{(\ell)}_a \langle\theta_*,a_{*}-a\rangle \\
    &=\sum_{\ell=1}^{L}\sum_{a\in \textrm{Supp}(\pi_{\ell}) }\ceil*{\frac{T^{(\ell)}_a}{M}} \Delta_a \\
    &\leq  \underbrace{\sum_{\ell=1}^{L}\sum_{a\in \textrm{Supp}(\pi_{\ell}) }\frac{T^{(\ell)}_a}{M} \Delta_a}_{T_1}+\underbrace{\sum_{\ell=1}^{L}\sum_{a\in \textrm{Supp}(\pi_{\ell}) }\Delta_a}_{T_2}. 
\end{aligned}
\end{equation}
We now bound $T_1$ and $T_2$ separately. For bounding $T_1$, we have:
\begin{equation}
\begin{aligned}
    T_1&=\sum_{\ell=1}^{L}\sum_{a\in \textrm{Supp}(\pi_{\ell}) }\frac{T^{(\ell)}_a}{M} \Delta_a \\ 
    &= \frac{d}{M} \sum_{\ell=1}^{L} \frac{1}{\epsilon^2_{\ell}} \log{\left( \frac{1}{\delta_{\ell}} \right)} \sum_{a\in \textrm{Supp}(\pi_{\ell}) } \hspace{-3mm} \pi_{\ell}(a) \Delta_a \\
   & \leq \frac{8d}{M} \sum_{\ell=1}^{L} \frac{1}{\epsilon^2_{\ell}} \log{\left( \frac{1}{\delta_{\ell}}\right)} \gamma_{\ell} \\
   &= \frac{8\sqrt{2}C (1+\alpha \sqrt{M}) d}{M} \sum_{\ell=1}^{L} \frac{1}{\epsilon_{\ell}} \log{\left( \frac{10K^2 \ell^2}{\delta}  \right)} \\
   & \leq \frac{8\sqrt{2}C (1+\alpha \sqrt{M}) d}{M} \log{\left( \frac{10K^2 L^2}{\delta}  \right)} \sum_{\ell=1}^{L} 2^{\ell}\\
   & = O\left(\frac{(1+\alpha \sqrt{M}) d}{M} \log{\left( \frac{10K^2 L^2}{\delta}  \right)} 2^L \right). 
\end{aligned}
    \label{eqn:boundonT1}
\end{equation}
In the third step above, we used $\sum_{a\in \textrm{Supp}(\pi_{\ell}) } \pi_{\ell}(a)=1$. 
We now need an upper-bound on the term $2^L$. To that end, notice that the length of the horizon $T$ is bounded below by the length of the last epoch, i.e., the $L$-th epoch. Moreover, the duration of the $L$-th epoch corresponds to the number of arm-pulls made by any single good agent during the $L$-th epoch. We thus have:
\begin{equation}
    \begin{aligned}
    T &\geq \sum_{a\in \textrm{Supp}(\pi_{L}) } \hspace{-2mm} \frac{T^{(L)}_a}{M} \\
    & \geq \frac{ 4^{L} d}{M} \sum_{a\in \textrm{Supp}(\pi_{L}) } \hspace{-2mm} \pi_{L}(a) \log{\left( \frac{1}{\delta_{L}}\right)} \\
    & = \frac{4^{L} d}{M} \log{\left( \frac{10K^2 L^2}{\delta}  \right)} \sum_{a\in \textrm{Supp}(\pi_{L}) } \hspace{-2mm} \pi_{L}(a) \\
    & = \frac{ 4^{L} d}{M} \log{\left( \frac{10K^2 L^2}{\delta}  \right)}.
    \end{aligned}
\end{equation}
 Thus, $2^L \leq \sqrt{MT/(d \log{\left( {10K^2 L^2}/{\delta}  \right)} }$. Plugging this bound in \eqref{eqn:boundonT1}, we obtain
$$ T_1 = O\left(\left(\alpha + \frac{1}{\sqrt{M}}\right) \sqrt{\log\left( \frac{KT}{\delta}\right) dT} \right) = \tilde{O}\left(\left(\alpha + \frac{1}{\sqrt{M}}\right) \sqrt{dT} \right).$$ 
As for the term $T_2$, we have
\begin{equation}
\begin{aligned}
    T_2 &= \sum_{\ell=1}^{L}\sum_{a\in \textrm{Supp}(\pi_{\ell}) }\hspace{-3mm} \Delta_a \\
    & \leq 8 \sum_{\ell=1}^{L} \gamma_{\ell}  |\textrm{Supp}(\pi_{\ell})| \\
    & \overset{(a)}\leq 384 \sqrt{2}C d \log \log d \left( 1+ \alpha \sqrt{M} \right) \sum_{\ell=1}^{L} 2^{-\ell} \\
    &= O \left(d \log \log d \left( 1+ \alpha \sqrt{M} \right) \right) \\
    &\overset{(b)} =\tilde{O}\left(\left(\alpha + \frac{1}{\sqrt{M}}\right) \sqrt{dT} \right). 
\end{aligned} 
\end{equation}
In the above steps, for (a), recall from line 1 of \texttt{RCLB} that $|\textrm{Supp}(\pi_{\ell})| \leq 48d \log \log d$ based on the approximate G-optimal design computation. For (b), we used the fact that by assumption, $T \geq Md$. Combining the bounds on $T_1$ and $T_2$, and recalling that $\mathbb{P}(\mc{E}) \geq 1-\delta$, we have that with probability at least $1-\delta$, 
\begin{equation}
    \sum_{t=1}^{T}\langle\theta_*,a_{*}-a_{i,t}\rangle = O\left(\left(\alpha + \frac{1}{\sqrt{M}}\right) \sqrt{\log\left( \frac{KT}{\delta}\right) dT} \right) = \tilde{O}\left(\left(\alpha + \frac{1}{\sqrt{M}}\right) \sqrt{dT} \right).
\label{eqn:reg_bnd}
\end{equation}
This concludes the proof. 
\end{proof}

We now provide a proof for Corollary \ref{corr:group_reg}.

\begin{proof} (\textbf{Proof of Corollary \ref{corr:group_reg}}) Recall from the proof of Theorem \ref{thm:RCPLB} that there exists a clean event $\mc{E}$ of measure at least $1-\delta$ on which the regret of every good agent is bounded above as per Eq.~\eqref{eqn:reg_bnd}. Let $\tilde{C}$ be an upper bound on the maximum instantaneous regret, i.e., 
$$ \max_{a\in \mc{A}} \langle \theta_*, a_* -a \rangle \leq \tilde{C}.$$ 

Since $\Vert \theta_* \Vert \leq 1$, and $\Vert a_k \Vert \leq 1, \forall a_k \in \mc{A}$, an upper bound of $\tilde{C} =2$ works for our case. Now set $\delta = \frac{1}{MT}$ and observe that:
\begin{equation}
\begin{aligned}
R_T &= \mathbb{E} \left[\sum_{i\in [M] \setminus \mathcal{B}} \sum_{t=1}^{T} \langle \theta_*, a_*-a_{i,t} \rangle \right]\\
 &=\mathbb{E} \left[\sum_{i\in [M] \setminus \mathcal{B}} \sum_{t=1}^{T} \langle \theta_*, a_*-a_{i,t} \rangle \Big| \mc{E} \right] \mathbb{P}(\mc{E}) + \mathbb{E} \left[\sum_{i\in [M] \setminus \mathcal{B}} \sum_{t=1}^{T} \langle \theta_*, a_*-a_{i,t} \rangle \Big| \mc{E}^c \right] \mathbb{P}(\mc{E}^c)\\
 & \leq C_1 \left(\alpha M + \sqrt{M}\right) \sqrt{\log\left( {KMT}\right) dT} + \tilde{C}MT \times \frac{1}{MT}\\
 & = O\left(\left(\alpha M + \sqrt{M}\right) \sqrt{\log\left( {KMT}\right) dT} \right)\\
 & = \tilde{O}\left(\left(\alpha M + \sqrt{M}\right) \sqrt{dT} \right).
\end{aligned}
\end{equation}
In the above steps, $C_1$ is a suitably large universal constant. 
\end{proof}
\newpage
\section{Lower Bound Analysis: Proof of Theorem \ref{thm:lower_bnd}}
\label{app:proof_lower_bnd}
In this section, we will prove Theorem \ref{thm:lower_bnd}. Before diving into the technical details, we remind the reader that we consider a slightly different adversarial model from the one considered throughout the paper. In this modified model, with probability $\alpha$, each agent is adversarial independently of the other agents. We will consider a class of policies $\Pi$ where at each time-step, the same action (decided by the server) is played by every agent. Thus, the regret incurred by any individual agent is the same as the regret incurred by the server. Finally, to prove the lower bound, we will focus on a class of $2$-armed bandits where the reward distribution of each arm is Gaussian with unit variance; such a class of bandits is succinctly denoted by $\mc{E}^{(2)}_{\mc{N}}(1)$. 

We will have occasion to use the following result \cite[Theorem 14.2]{tor}. 
\begin{mylem}[label=lemma:Breta]{Bretagnolle-Huber Inequality}{} Let $P$ and $Q$ be two probability measures on the same measurable space $(\Omega, \mc{F})$, and let $A \in \mc{F}$ be any arbitrary event. Then, $$ P(A)+Q(A^c) \geq \frac{1}{2} \exp\left(-\textrm{KL}(P,Q)\right), $$
where $A^c$ is the complement of the event $A$, and  $\textrm{KL}(P,Q)$ is the Kullback-Leibler distance between $P$ and $Q$. 
\end{mylem}

Our proof comprises of two main steps. First, we construct two \textit{distinct} bandit instances within the class $\mc{E}^{(2)}_{\mc{N}}(1)$ such that the two instances - although different - appear \textit{identical} to the server. Second, we devise an attack strategy and argue that regardless of the policy played by the server, it will end up suffering a regret of $\Omega(\alpha\sqrt{T})$ upon  interacting with at least one of the two instances; here, $T$ is the horizon for our problem. We now elaborate on these two steps. 

$\bullet$ \textbf{Step 1. Construction of the two bandit instances.} We first take a detour and describe an idea that is typically used to prove lower bounds for the robust mean estimation literature \cite{chenrobust,lai}. It will soon be apparent how such an idea can be exploited to construct the two bandit instances for our problem. We show that there are two univariate Gaussian distributions $P_1 = \mc{N}(\mu_1, 1)$, $P_2=\mc{N}(\mu_2, 1)$, and two $T$-dimensional distributions $Q_1, Q_2$, such that $\Vert \mu_1 - \mu_2 \Vert_2 = \Omega(\alpha/\sqrt{T})$, and:
\begin{equation}
    (1-\alpha) P^T_1 + \alpha Q_1 = (1-\alpha) P^T_2 + \alpha Q_2,
\label{eqn:identical}
\end{equation}
where $P^T_1$ (resp., $P^T_2$) is the joint distribution of $T$ i.i.d. samples drawn from $P_1$ (resp., $P_2$). Clearly, $P^T_1$ (resp., $P^T_2$) is equivalent to a $T$-dimensional Gaussian distribution $\mc{N}(\boldsymbol{\mu}_1, I_T)$ (resp., $\mc{N}(\boldsymbol{\mu}_2, I_T)$), where $\boldsymbol{\mu}_1$ (resp., $\boldsymbol{\mu}_2$) is a $T$-dimensional vector with each entry equal to $\mu_1$ (resp., $\mu_2$). Let $\phi_1$ be the p.d.f. of $P^T_1$ and $\phi_2$ be the p.d.f. of $P^T_2$. Next, let $\mu_1$ and $\mu_2$ be chosen such that the total variation distance $\delta(P^T_1, P^T_2)$ between $P^T_1$ and $P^T_2$ is $$ \frac{1}{2} \int \Vert \phi_1 -\phi_2 \Vert_1 dx = \frac{\alpha}{1-\alpha}.$$

Let $Q_1$ be the distribution with p.d.f. $\frac{1-\alpha}{\alpha} (\phi_2 - \phi_1) \mathbf{1}_{\phi_2 \geq \phi_1}$, and $Q_2$ be the distribution with p.d.f. $\frac{1-\alpha}{\alpha} (\phi_1 - \phi_2) \mathbf{1}_{\phi_1 \geq \phi_2}$. With such a construction of $Q_1$ and $Q_2$, one can verify that the equality in Eq.~\eqref{eqn:identical} is satisfied; see, for instance, the arguments in Appendix E of \cite{chenrobust}. Now from Pinsker's inequality, we know that:
$$ \sqrt{ \frac{1}{2} \textrm{KL}(P^T_1, P^T_2) } \geq \delta(P^T_1, P^T_2) = \frac{\alpha}{1-\alpha}, $$
where we used $\textrm{KL}(P^T_1, P^T_2)$ to denote the Kullback-Leibler distance between $P^T_1$ and $P^T_2$. We conclude that:
$$ \frac{1}{2} \Vert \boldsymbol{\mu}_1 - \boldsymbol{\mu}_2 \Vert^2_2 = \textrm{KL}(P^T_1, P^T_2) \geq 2  { \left( \frac{\alpha}{1-\alpha} \right) }^2.$$
This, in turn, implies
$$ \sqrt{T} \left(\vert \mu_1 - \mu_2 \vert\right) = \Vert \boldsymbol{\mu}_1 - \boldsymbol{\mu}_2 \Vert_2 \geq  \frac{2\alpha}{1-\alpha} \geq 2 \alpha.$$ 
We have thus shown that there exist $\mu_1, \mu_2 \in \mathbb{R}$, satisfying $\vert \mu_1 - \mu_2 \vert \geq 2\alpha/ \sqrt{T}$, such that with $P_1 = \mc{N}(\mu_1, 1)$, $P_2=\mc{N}(\mu_2, 1)$, one can satisfy Eq.~\eqref{eqn:identical} with appropriately chosen $T$-dimensional distributions $Q_1$ and $Q_2$. With these ideas in place, we now return to our bandit problem. 

Without loss of generality, suppose $\mu_2 > \mu_1$, where $\mu_2$ and $\mu_1$ are as in the construction described above. Let us construct two bandit instances $\nu$ and $\nu'$, each involving two arms, i.e., $\mc{A}=\{a_1, a_2\}$. Let $r^{(\nu)}_{a_k}$ and $r^{(\nu')}_{a_k}$ denote the reward distribution of arm $a_k, k\in\{1,2\}$, in instance $\nu$ and instance $\nu'$, respectively. These reward distributions are chosen as follows.
\begin{equation}
    \begin{aligned}
    & \textrm{Instance} \hspace{1mm} \nu: r^{(\nu)}_{a_1} \sim  \mc{N}\left(\frac{\mu_1 + \mu_2}{2}, 1\right); \hspace{2mm} r^{(\nu)}_{a_2} \sim \mc{N}(\mu_1, 1)\\
    &\textrm{Instance} \hspace{1mm} \nu': r^{(\nu')}_{a_1} \sim \mc{N}\left(\frac{\mu_1 + \mu_2}{2}, 1\right); \hspace{2mm} r^{(\nu')}_{a_2} \sim  \mc{N}(\mu_2, 1).\\
    \end{aligned}
\end{equation}
Thus, the distribution of the first arm is the same in both instances. However, as $\mu_2 > \mu_1$, the first arm is the best arm in the first instance while the opposite is true for the second instance. The attack strategy for the adversarial agents will be dictated by the distributions $Q_1$ and $Q_2$ in a manner to be described shortly.

$\bullet$ \textbf{Step 2. The attack strategy and regret analysis.} Inspired by the argument in the proof of \cite[Theorem 5]{bubeck}, we consider a full information setting where the server has access to $t$ reward samples from \textit{each} arm from \textit{each} agent at time-step $t$. Since this full information setting is simpler than the bandit setting, a lower bound for the former implies one for the latter. 

Here is the attack strategy. Suppose an agent is adversarial (which happens with probability $\alpha$). In either instance, for arm $1$, it reports samples from the true distribution for arm $1$ corresponding to that instance. In other words, the reward samples for arm 1 are not corrupted by the agent. As for arm 2, in instance $\nu$ (resp., $\nu'$), the first $t$ reward samples (where $t\in [T]$) corresponding to $a_2$ are generated from $Q^t_1$ (resp., $Q^t_2$) by the adversarial agent. Here, $Q^t_1$ (resp., $Q^t_2$) is the marginal of the $T$-dimensional distribution $Q_1$ (resp., $Q_2$) corresponding to the first $t$ components. To sum up, in instance $\nu$, the joint distribution $D^{(\nu)}_{a_1}$ of rewards for $a_1$ over the horizon $T$, as seen by the server from any given agent, is the $T$-dimensional Gaussian distribution $\mc{N}(\bar{\boldsymbol{\mu}}, I_T)$, where $\bar{\boldsymbol{\mu}}$ is a $T$-dimensional vector with each component equal to $\frac{\mu_1+\mu_2}{2}$. Based on our discussion above, $D^{(\nu)}_{a_1}=D^{(\nu')}_{a_1}$. Let $D^{(\nu)}_{a_2}$ and $D^{(\nu')}_{a_2}$ have analogous meanings for arm $a_2$. Then, we have:
\begin{equation}
    D^{(\nu)}_{a_2}  = (1-\alpha) P^T_1 + \alpha Q_1; \hspace{2mm}
    D^{(\nu')}_{a_2} = (1-\alpha) P^T_2 + \alpha Q_2. 
\end{equation}
In light of Eq.~\eqref{eqn:identical}, however, we have $D^{(\nu)}_{a_2} = D^{(\nu')}_{a_2}$. Essentially, what we have established is the following: \textit{the joint distribution of rewards for  each arm over the horizon $T$, as seen by the server from each agent, is identical for both instances}.  

In what follows, given any two distributions $D_1$ and $D_2$, let $D_1 \otimes D_2$ represent their product distribution. Since the rewards across arms are independent, the joint distribution of rewards for both arms is given by $D^{(\nu)} \triangleq D^{(\nu)}_{a_1} \otimes D^{(\nu)}_{a_2}$ in instance $\nu$, and $D^{(\nu')} \triangleq D^{(\nu')}_{a_1} \otimes D^{(\nu')}_{a_2}$ in instance $\nu'$. Since rewards across agents are independent, the joint distributions of rewards from \textit{all} agents, as seen by the server in each of the two instances, are given by:
$$ D^{(\nu)}_M \triangleq \underbrace{D^{(\nu)} \otimes D^{(\nu)} \otimes \cdots \otimes D^{(\nu)}}_{M\textrm{-fold product distribution}}; \hspace{2mm} D^{(\nu')}_M \triangleq \underbrace{D^{(\nu')} \otimes D^{(\nu')} \otimes \cdots \otimes D^{(\nu')}}_{M\textrm{-fold product distribution}}.$$

Let $\mathbb{E}_{\nu}[\cdot]$ (resp., $\mathbb{E}_{\nu'}[\cdot]$) represent the expectation operation w.r.t. the measure $D^{(\nu)}_M$ (resp., $D^{(\nu')}_M$). Let us use $\mathbb{P}_{\nu}$ as a shorthand for $D^{(\nu)}_M$, and $\mathbb{P}_{\nu'}$ as a shorthand for $D^{(\nu')}_M$. Furthermore, let $n_k(T)$ be the random variable representing the total number of times arm $a_k, k\in\{1,2\}$, is chosen by the server over the horizon $T$. Finally, recall that $R^{(s)}_T(\nu)$ (resp., $R^{(s)}_T(\nu')$) is the regret incurred by the server upon interaction with instance $\nu$ (resp., instance $\nu'$). 

In instance $\nu$, each time arm 2 is chosen by the server, it incurs an instantaneous regret of $(\mu_2 - \mu_1)/{2}$. We thus have:
\begin{equation}
    R^{(s)}_T(\nu) = \left(\frac{\mu_2-\mu_1}{2} \right)\left( T - \mathbb{E}_{\nu}[n_1(T)] \right) \geq \frac{(\mu_2 - \mu_1)T}{4}  \mathbb{P}_{\nu}\left(n_1(T) \leq \frac{T}{2} \right). 
\label{eqn:regret_nu}
\end{equation}
To see why the latter inequality is true, observe:
\begin{equation}
    \begin{aligned}
    \mathbb{E}_{\nu}[n_1(T)] &= \mathbb{E}_{\nu}\left[n_1(T) | n_1(T) \leq \frac{T}{2}\right] \mathbb{P}_{\nu}\left(n_1(T) \leq \frac{T}{2} \right) +  \mathbb{E}_{\nu}\left[n_1(T) | n_1(T) > \frac{T}{2}\right] \mathbb{P}_{\nu}\left(n_1(T) > \frac{T}{2} \right)\\
     & \leq  \frac{T}{2} \mathbb{P}_{\nu}\left(n_1(T) \leq \frac{T}{2} \right) + T \left(1 - \mathbb{P}_{\nu}\left(n_1(T) \leq \frac{T}{2} \right) \right)\\
     & = T- \frac{T}{2} \mathbb{P}_{\nu}\left(n_1(T) \leq \frac{T}{2} \right).
     \end{aligned}
\end{equation}

In instance $\nu'$, each time arm 1 is chosen by the server, it incurs an instantaneous regret of $(\mu_2 - \mu_1)/{2}$. We thus have: 
\begin{equation}
    R^{(s)}_T(\nu') = \left(\frac{\mu_2-\mu_1}{2} \right)  \mathbb{E}_{\nu'}[n_1(T)]  \geq \frac{(\mu_2 - \mu_1)T}{4}  \mathbb{P}_{\nu'}\left(n_1(T) > \frac{T}{2} \right). 
\label{eqn:regret_nu'}
\end{equation}
Combining Eq.~\eqref{eqn:regret_nu} and Eq.~\eqref{eqn:regret_nu'} yields:
\begin{equation}
\begin{aligned}
    \max\{R^{(s)}_T(\nu), R^{(s)}_T(\nu')\} &\geq \frac{1}{2} \left(R^{(s)}_T(\nu)+R^{(s)}_T(\nu')\right)\\
    &\geq \frac{(\mu_2 - \mu_1)T}{8} \left( \mathbb{P}_{\nu}\left(n_1(T) \leq \frac{T}{2} \right) + \mathbb{P}_{\nu'}\left(n_1(T) > \frac{T}{2} \right)  \right)\\ 
    &\overset{(a)}\geq \frac{(\mu_2 - \mu_1)T}{16} \exp{\left(-\textrm{KL}\left(\mathbb{P}_{\nu}, \mathbb{P}_{\nu'} \right)\right)} \\ 
    &\overset{(b)}\geq \frac{\alpha \sqrt{T}}{8} \exp{\left(-\textrm{KL}\left(\mathbb{P}_{\nu}, \mathbb{P}_{\nu'} \right)\right)}\\
    &\overset{(c)}=\frac{\alpha \sqrt{T}}{8}.
\end{aligned}
\label{eqn:result_lb}
\end{equation}
In the above steps, we used the Bretagnolle-Huber inequality (namely, Lemma \ref{lemma:Breta}) for (a); for (b), we used the fact that $\mu_2$ and $\mu_1$ were chosen in Step 1 to satisfy $\mu_2 - \mu_1 \geq 2\alpha/\sqrt{T}$; and for (c), we used $\textrm{KL}\left(\mathbb{P}_{\nu}, \mathbb{P}_{\nu'} \right)=0$. To see why $\textrm{KL}\left(\mathbb{P}_{\nu}, \mathbb{P}_{\nu'} \right)=0$, we use the chain-rule for relative entropies to obtain:
\begin{equation}
\begin{aligned}
   \textrm{KL}\left(\mathbb{P}_{\nu}, \mathbb{P}_{\nu'} \right) & = \textrm{KL}\left(D^{(\nu)}_M, D^{(\nu')}_M \right)\\
   & = M \left(\textrm{KL}\left(D^{(\nu)}, D^{(\nu')} \right) \right) \\
   & = M \left( \textrm{KL}\left(D^{(\nu)}_{a_1}, D^{(\nu')}_{a_1} \right) + \textrm{KL}\left(D^{(\nu)}_{a_2}, D^{(\nu')}_{a_2} \right) \right)\\
   &= 0,
\end{aligned}
\end{equation}
where the last step is a consequence of the fact that $D^{(\nu)}_{a_1}=D^{(\nu')}_{a_1}$, and $D^{(\nu)}_{a_2}=D^{(\nu')}_{a_2}$. The claim of Theorem \ref{thm:lower_bnd} follows from noting that the resulting lower bound in Eq.~\eqref{eqn:result_lb} holds regardless of the number of agents $M$. 
\hspace{137mm}$\blacksquare$

\newpage
\begin{algorithm}[H]
\caption{Robust Collaborative Phased Elimination for Generalized Linear Bandits  (\texttt{RC-GLM})}
\label{algo:RC-GLM}  
 \begin{algorithmic}[1] 
\Statex \hspace{-5mm} \textbf{Input:} Action set $\mathcal{A}=\{a_1, \ldots, a_K\}$, confidence parameter $\delta$, and corruption fraction $\alpha.$
\Statex \hspace{-5mm} \textbf{Initialize:} $\ell=1$ and $\mathcal{A}_1=\mathcal{A}.$
\State Let $ V_{\ell}(\pi) \triangleq \sum_{a\in\mathcal{A}_{\ell}} \pi(a) aa'$ and $g_{\ell}(\pi) \triangleq \max_{a\in\mathcal{A}_{\ell}} {\Vert a \Vert^2}_{{V_{\ell}(\pi)}^{-1}}.$ Server solves an approximate G-optimal design problem to compute a distribution $\pi_{\ell}$ over $\mathcal{A}_{\ell}$ such that $g_{\ell}(\pi_{\ell}) \leq 2d$ and  $|\textrm{Supp}(\pi_{\ell})| \leq 48d \log\log d.$ 
\State For each $a\in\mathcal{A}_{\ell}$, server computes $m^{(\ell)}_a$ via  Eq. \eqref{eqn:arm_pulls}, and broadcasts $\{m^{(\ell)}_a\}_{a\in\mathcal{A}_{\ell}}$ to all agents. 
\For {$i\in [M]\setminus \mathcal{B}$} 
\State For each arm $a\in \mathcal{A}_{\ell}$, pull it $m^{(\ell)}_a$ times. Let $r^{(\ell)}_{i,a}$ be the average of the rewards observed by agent $i$ for arm $a$ during phase $\ell$. 
\State Compute $\tilde{V}_{\ell}$ and $Y_{i,\ell}$ as  follows.
\begin{equation}
     \tilde{V}_{\ell} = \hspace{-3mm}  \sum_{a\in\textrm{Supp}(\pi_{\ell})} \hspace{-3mm} m^{(\ell)}_a aa' \hspace{1mm} ; \hspace{1.5mm} Y_{i,\ell} = \hspace{-5mm} \sum_{a\in\textrm{Supp}(\pi_{\ell})} \hspace{-5mm} m^{(\ell)}_a r^{(\ell)}_{i,a} a.
\nonumber
\end{equation}
\State Transmit $Y_{i,\ell}$ to server.  Adversarial agents can transmit arbitrary vectors at this stage.
\EndFor
\State Server computes a statistic $X_{\ell}$ as follows: 
$$ X_{\ell}=\texttt{ITW}(\{\tilde{V}^{-1/2}_{\ell} Y_{i,\ell}, i \in [M]\}),$$ 
where \texttt{ITW}($\cdot$) is the output of the Iteratively Reweighted Mean Estimator from \cite{Dalal}. 

\State Server computes a robust estimate $\hat{\theta}^{(\ell)}$ of $\theta_*$ by solving:
$$ h_{\ell}(\hat{\theta}^{(\ell)}) = \tilde{V}^{1/2}_{\ell} X_{\ell}.$$ 
\State Define robust confidence threshold $\bar{\gamma}_{\ell} = 4 C_1 (k_2/k_1)  \left(\sqrt{d}+\alpha\sqrt{M \log(1/\alpha)} \right) \epsilon_{\ell}$, where $\epsilon_{\ell} = 2^{-\ell}$, and $C_1$ is as in Lemma \ref{lemma:Dalal}. Server performs phased elimination with the estimate $\hat{\theta}^{(\ell)}$ and confidence threshold $\bar{\gamma}_{\ell}$ to update active arm set:
\begin{equation}
    \mathcal{A}_{\ell+1} = \{a\in\mathcal{A}_{\ell}: \max_{b\in\mathcal{A}_{\ell}} \mu(\langle \hat{\theta}^{(\ell)}, b \rangle) - \mu(\langle \hat{\theta}^{(\ell)}, a \rangle) \leq 2 \bar{\gamma}_{\ell}\}. 
\label{eqn:phased_el_GLM}
\end{equation}
\State $\ell = \ell+1$ and \textbf{Goto} line 1.
\end{algorithmic}
 \end{algorithm}

\section{Algorithms and Analysis for the Generalized Linear Bandit Model}
\label{app:gen_lin}
In this section, we first provide a detailed outline of the \texttt{RC-GLM} algorithm introduced in Section \ref{sec:gen_lin}; see Algorithm \ref{algo:RC-GLM}. We then proceed to analyze \texttt{RC-GLM}, and provide a proof for Theorem \ref{thm:GLM}. Finally, since \texttt{RC-GLM} uses the iteratively reweighted mean estimator from \cite{Dalal} as a sub-routine, we also provide a description of this estimator to keep the paper self-contained; this description, however, is deferred to the end of the section. We start by reminding the reader that the non-linear observation model of interest to us in this section is as follows:
\begin{equation}
    y_{i,t}=\mu\left(\langle \theta_*, a_{i,t} \rangle \right)+\eta_{i,t},
\end{equation}
where $\mu:\mathbb{R} \rightarrow \mathbb{R}$ is the link function. We also recall the definition of $h_{\ell}(\theta)$:
$$ h_{\ell}(\theta) \triangleq \sum_{a\in\textrm{Supp}(\pi_{\ell})}  m^{(\ell)}_a \mu(\langle \theta,a \rangle) a, \forall \theta \in \Theta.$$  

From comparing \texttt{RC-GLM} (Algorithm \ref{algo:RC-GLM}) to \texttt{RCLB} (Algorithm \ref{algo:RCPLB}), we note that while both algorithms share the same general structure, the key difference between the two stems from the manner in which the robust confidence thresholds are computed. In particular, to tackle the difficulty posed by the non-linearity of the observation map, we first compute a robust estimate $\hat{\theta}^{(\ell)}$ of $\theta_*$ at the server - a route that we avoided in \texttt{RCLB} - and then use such an estimate to devise a phased elimination rule. 

\subsection{Proof of Theorem \ref{thm:GLM}}
In this section, we prove Theorem \ref{thm:GLM}. The crux of the analysis lies in deriving a robust confidence bound akin to that in Lemma \ref{lemma:rob_conf_main}. To work towards such a result, we need to first go through a few intermediate steps:  

\textbf{Step 1.} Prove that conditioned on $\mc{F}_{\ell}$, for each $i\in [M]\setminus\mc{B}$, $\tilde{V}^{-1/2}_{\ell} Y_{i,\ell}$ is a $d$-dimensional Gaussian random variable with mean $\tilde{V}^{-1/2}_{\ell} h_{\ell}(\theta_*)$, and covariance matrix $\Sigma=I_d$.\footnote{Recall that $\mc{F}_{\ell}$ is the $\sigma$-algebra generated by all the actions and rewards up to the beginning of epoch $\ell$.} 

\textbf{Step 2.} Use the result from Step 1, along with the error-bounds of the iteratively reweighted Gaussian mean estimator from \cite{Dalal}, to derive a high-probability error-bound on $\Vert X_{\ell} - \tilde{V}^{-1/2}_{\ell} h_{\ell}(\theta_*) \Vert$.

\textbf{Step 3.} Exploit regularity properties of the link function in tandem with the bounds from Step 2, to derive high-probability error bounds on $\vert \mu(\langle \hat{\theta}^{(\ell)},a \rangle) - \mu(\langle \theta_*,a \rangle) \vert$, for each $a\in \mc{A}_{\ell}$. 

We now proceed to formally establish each of the above steps, starting with step 1.

\begin{mylem}[label=lemma:prop_GLM]{}{} For each epoch $\ell$, and each good agent $i\in [M]\setminus \mc{B}$, it holds that:
\begin{equation}
    \mathbb{E}\left[\tilde{V}^{-1/2}_{\ell} Y_{i,\ell}| \mc{F}_{\ell}\right] = \tilde{V}^{-1/2}_{\ell} h_{\ell}(\theta_*); \hspace{2mm} \textrm{and}
\label{eqn:GLM_mean}
\end{equation}
\begin{equation}
    \mathbb{E}\left[\left(\tilde{V}^{-1/2}_{\ell} Y_{i,\ell} - \tilde{V}^{-1/2}_{\ell} h_{\ell}(\theta_*) \right)\left(\tilde{V}^{-1/2}_{\ell} Y_{i,\ell} - \tilde{V}^{-1/2}_{\ell} h_{\ell}(\theta_*) \right)'| \mc{F}_{\ell}\right] = I_d. 
\label{eqn:GLM_var}
\end{equation}
\end{mylem}
\begin{proof}
Fix an epoch $\ell$, and a good agent $i\in [M]\setminus \mc{B}$. We start by observing that:
\begin{equation}
    \begin{aligned}
Y_{i,\ell}
    & =   \sum_{a\in\textrm{Supp}(\pi_{\ell})} \hspace{-5mm}   m^{(\ell)}_a r^{(\ell)}_{i,a} a \\
    & =  \sum_{a\in\textrm{Supp}(\pi_{\ell})} \hspace{-5mm}   m^{(\ell)}_a \left( \mu \left(\langle  \theta_*, a \rangle \right) + \bar{\eta}^{(\ell)}_{i,a} \right) a  \\
    & = h_{\ell}(\theta_*) +  \sum_{a\in\textrm{Supp}(\pi_{\ell})} \hspace{-5mm}   m^{(\ell)}_a \bar{\eta}^{(\ell)}_{i,a} a,
    \end{aligned}
\label{eqn:unrolled_GLM}
\end{equation}
where for the last step, we used the definition of $h_{\ell}(\theta_*)$. Just as in the proof of Theorem \ref{thm:RCPLB}, we have used  $\bar{\eta}^{(\ell)}_{i,a}$ to denote the average of the noise terms associated with the rewards observed by agent $i$ during phase $\ell$ for arm $a$. We thus have:
$$ \tilde{V}^{-1/2}_{\ell} Y_{i,\ell} = \tilde{V}^{-1/2}_{\ell} h_{\ell}(\theta_*) +  \tilde{V}^{-1/2}_{\ell} \left(\sum_{a\in\textrm{Supp}(\pi_{\ell})} \hspace{-5mm}   m^{(\ell)}_a \bar{\eta}^{(\ell)}_{i,a} a \right). $$
Now conditioned on $\mc{F}_{\ell}$, the only randomness in the above equation stems from the noise terms $\{\bar{\eta}^{(\ell)}_{i,a}\}, a \in \textrm{Supp}(\pi_{\ell})$, that are each zero-mean. The claim in Eq.~\eqref{eqn:GLM_mean} thus follows. 

Based on Eq.~\eqref{eqn:unrolled_GLM}, we have:
\begin{equation}
    \begin{aligned}
    \mathbb{E}\left[\left(Y_{i,\ell} -  h_{\ell}(\theta_*) \right)\left(Y_{i,\ell} -  h_{\ell}(\theta_*) \right)'| \mc{F}_{\ell}\right] &= \mathbb{E}\left[\left( \sum_{a\in\textrm{Supp}(\pi_{\ell})} \hspace{-5mm}   m^{(\ell)}_a \bar{\eta}^{(\ell)}_{i,a} a \right)\left( \sum_{a\in\textrm{Supp}(\pi_{\ell})} \hspace{-5mm}   m^{(\ell)}_a \bar{\eta}^{(\ell)}_{i,a} a \right)' \Big| \mc{F}_{\ell}\right] \\
      & \overset{(a)} =  \sum_{a\in\textrm{Supp}(\pi_{\ell})} \hspace{-5mm} {\left(m^{(\ell)}_a\right)}^2  \mathbb{E}\left[{\left(\bar{\eta}^{(\ell)}_{i,a}\right)}^2  \right] aa' \\
      & \overset{(b)} = \sum_{a\in\textrm{Supp}(\pi_{\ell})} \hspace{-5mm} {m^{(\ell)}_a} aa' \\
      & \overset{(c)} = \tilde{V}_{\ell}. 
    \end{aligned}
\label{eqn:VAR_com}
\end{equation}
In the above steps, (a) follows by observing that the noise terms are independent across different arms; hence, the expectation of each of the cross-terms vanish. For (b), we used the fact that $\bar{\eta}^{(\ell)}_{i,a}$ is the average of $m^{(\ell)}_a$ independent Gaussian noise terms, each with zero-mean and unit variance; hence, $\mathbb{E}\left[{\left(\bar{\eta}^{(\ell)}_{i,a}\right)}^2  \right] = 1/(m^{(\ell)}_a).$ For (c), we simply used the definition of $\tilde{V}_{\ell}$. In light of Eq.~\eqref{eqn:VAR_com}, it is easy to see why Eq.~\eqref{eqn:GLM_var} holds. 
\end{proof}

We now state - adapted to our notation - one of the main convergence guarantees from \cite{Dalal} for the iteratively reweighted mean estimator. 

\begin{mylem}[label=lemma:Dalal]{}{}
Suppose we are given $M$ $d$-dimensional samples $x_1, \ldots, x_M$, such that $(1-\alpha)M$ of these samples are drawn i.i.d. from $\mc{N}(v, \Sigma)$, where $v \in \mathbb{R}^d$ is an unknown mean vector, and $\Sigma \in \mathbb{R}^{d\times d}$ is a known covariance matrix. The remaining $\alpha M$ samples can be arbitrary. Let the corruption fraction $\alpha$ satisfy $\alpha < (5-\sqrt{5})/10$, and let $\delta \in (16 \exp{(-M)}, 1)$ be a given tolerance level. Then, with probability at least $1-\delta$, we have
\begin{equation}
    \Vert \hat{v}-v \Vert_2 \leq C_1 \Vert \Sigma \Vert^{1/2}_2 \left( \sqrt{ \frac{d+\log(16/\delta)}{M}} + \alpha \sqrt{\log{\frac{1}{\alpha}}} \right),
\end{equation}
where $C_1$ is a suitably large universal constant, and $\hat{v}$ is the output of the Iteratively Reweighted Mean Estimator, namely Algorithm 1 in \cite{Dalal}, when it takes as input the $M$ samples, the corruption fraction $\alpha$, and the covariance matrix $\Sigma$. 
\end{mylem}

Let us now see how the above bound can assist in our cause. Fix any epoch $\ell$, and recall that $\delta_{\ell}=\bar{\delta}/(K \ell^2)$, where $\bar{\delta}=\delta/(10K)$, and $\delta$ is the given confidence parameter. Suppose we want to derive an error-bound based on Lemma \ref{lemma:Dalal} that holds with probability at least $1-\delta_{\ell}$. For this to happen, we need $\delta_{\ell} > 16 \exp{(-M)}$. Since $\ell \leq T$, one can verify that the aforementioned condition is satisfied as long as $M$ is large enough in the following sense:
\begin{equation}
    M > \log{\left( \frac{160K^2 T^2}{\delta}\right)}. 
\label{eqn:cond_M}
\end{equation}
From now on, we assume that the above condition holds. Next, recall that $$X_{\ell}=\texttt{ITW}(\{\tilde{V}^{-1/2}_{\ell} Y_{i,\ell}, i \in [M]\}).$$ 

Based on Lemma \ref{lemma:prop_GLM}, Lemma \ref{lemma:Dalal}, and the same line of reasoning as used to arrive at Eq.~\eqref{eqn:filtrations}, we have that with probability at least $1-\delta_{\ell}$: 
\begin{equation}
    \Vert X_{\ell} - \tilde{V}^{-1/2}_{\ell} h_{\ell}(\theta_*) \Vert \leq  C_1  \left( \sqrt{ \frac{d+\log(16/\delta_{\ell})}{M}} + \alpha \sqrt{\log{\frac{1}{\alpha}}} \right).
\label{eqn:X_l_bound}
\end{equation}

We will call upon the above bound later in our analysis. For now, this ends step 2. As for step 3, we start with the following result. 

\begin{mylem}[label=lemma:MVT]{}{}
Consider any $\theta_1, \theta_2 \in \Theta$, and any epoch $\ell$. There exists a symmetric positive definite matrix $G_{\ell}(\theta_1;\theta_2)$ satisfying $k_1 \tilde{V}_{\ell} \preccurlyeq G_{\ell}(\theta_1;\theta_2)  \preccurlyeq k_2 \tilde{V}_{\ell}$, such that:
\begin{equation}
    h_{\ell}(\theta_1)-h_{\ell}(\theta_2)=G_{\ell}(\theta_1;\theta_2) (\theta_1-\theta_2). 
\end{equation}
\end{mylem}
\begin{proof}
For any $\theta\in\Theta$, let us denote by $\nabla h_{\ell}(\theta)$ the Jacobian matrix of $h_{\ell}(\cdot)$ at $\theta$. Such a matrix exists based on Assumption \ref{ass:link}. Now based on the mean value theorem, $\exists \alpha \in (0,1)$ such that
$$ h_{\ell}(\theta_1)-h_{\ell}(\theta_2)=\left(\nabla h_{\ell}\left(\alpha \theta_1 + (1-\alpha) \theta_2\right)\right)  (\theta_1-\theta_2). $$
Let $\bar{\theta}=\alpha \theta_1 + (1-\alpha) \theta_2$, and $G_{\ell}(\theta_1;\theta_2)=\nabla h_{\ell}(\bar{\theta})$. To complete the proof, we need to show that the matrix $G_{\ell}(\theta_1;\theta_2)$ defined above is symmetric, positive definite, and bounded above and below (in the Loewner sense) by scalar multiples of $\tilde{V}_{\ell}$. To that end, observe that:
\begin{equation}
    \begin{aligned}
    \nabla h_{\ell}(\bar{\theta}) & \overset{(a)}=  \sum_{a\in\textrm{Supp}(\pi_{\ell})} \hspace{-5mm}  m^{(\ell)}_a \dot{\mu}(\langle \bar{\theta},a \rangle) a a'\\
   & \overset{(b)} \succcurlyeq  k_1 \left(\sum_{a\in\textrm{Supp}(\pi_{\ell})} \hspace{-5mm}  m^{(\ell)}_a a a' \right)\\
    &= k_1 \tilde{V}_{\ell}.
    \end{aligned}
\end{equation}
In the above steps, we used the definition of $h_{\ell}(\cdot)$ for (a); and for (b), we used Assumption \ref{ass:link}. From the above steps, it is clear that $G_{\ell}(\theta_1;\theta_2)$ is symmetric. That it is also positive definite follows from the fact that $\tilde{V}_{\ell} \succ 0$. Finally, using the fact that $\mu(\cdot)$ is $k_2$-Lipschitz, and a similar line of reasoning, one can show that  $G_{\ell}(\theta_1;\theta_2) \preccurlyeq k_2 \tilde{V}_{\ell}$. This concludes the proof.   
\end{proof}

We now have all the pieces required to establish an analogue of Lemma \ref{lemma:rob_conf_main}. 

\begin{mylem}[label=lemma:GLM_conf_bnd]{Robust Confidence Intervals for \texttt{RC-GLM}}{} 
Suppose $\alpha < (5-\sqrt{5})/10$, and $M$ satisfies the condition in  Eq.~\eqref{eqn:cond_M}. Fix any epoch $\ell$. For each arm $a\in \mc{A}_{\ell}$, with probability at least $1-\delta_{\ell}$, it holds that:
\begin{equation}
    \vert \mu(\langle \hat{\theta}^{(\ell)},a \rangle) - \mu(\langle \theta_*,a \rangle) \vert \leq \bar{\gamma}_{\ell}, \hspace{2mm} \textrm{where} \hspace{2mm} \bar{\gamma}_{\ell} = 4 C_1 \frac{k_2}{k_1}  \left(\sqrt{d}+\alpha\sqrt{M \log(1/\alpha)} \right) \epsilon_{\ell},
\label{eqn:GLM_bnd}
\end{equation}
and $C_1$ is as in Lemma \ref{lemma:Dalal}.
\end{mylem}
\begin{proof}
Let us start by conditioning on the event of measure at least $1-\delta_{\ell}$ on which Eq.~\eqref{eqn:X_l_bound} holds. Invoking Lemma \ref{lemma:MVT}, we know that there exists a symmetrix positive definite matrix $G_{\ell}$ such that $k_1 \tilde{V}_{\ell} \preccurlyeq G_{\ell}  \preccurlyeq k_2 \tilde{V}_{\ell}$, and:\footnote{Here, we have dropped the dependence of $G_{\ell}$ on $\hat{\theta}^{(\ell)}$ and $\theta_*$ to lighten the notation.}
\begin{equation}
\begin{aligned}
    G_{\ell}\left( \hat{\theta}^{(\ell)}-\theta_* \right) & = h_{\ell}(\hat{\theta}^{(\ell)})-h_{\ell}(\theta_*)\\
    & = \tilde{V}^{1/2}_{\ell}X_{\ell} - h_{\ell}(\theta_*) \\
    & = \tilde{V}^{1/2}_{\ell} \left(X_{\ell} - \tilde{V}^{-1/2}_{\ell} h_{\ell}(\theta_*)\right).
\end{aligned}
\end{equation}
For the second step above, we used the fact that based on line 9 of \texttt{RC-GLM}, $h_{\ell}(\hat{\theta}^{(\ell)}) = \tilde{V}^{1/2}_{\ell} X_{\ell}$. Now fix any arm $a\in \mc{A}_{\ell}$, and observe that:
$$ \langle \hat{\theta}^{(\ell)}-\theta_*, a \rangle =  \Big \langle  G^{-1}_{\ell} \tilde{V}^{1/2}_{\ell} \left(X_{\ell} - \tilde{V}^{-1/2}_{\ell} h_{\ell}(\theta_*)\right), a \Big \rangle. $$
This, in turn, implies:
\begin{equation}
    \vert \langle \hat{\theta}^{(\ell)}-\theta_*, a \rangle \vert \leq \underbrace{\norm[\bigg]{G^{-1}_{\ell} \tilde{V}^{1/2}_{\ell} \left(X_{\ell} - \tilde{V}^{-1/2}_{\ell} h_{\ell}(\theta_*)\right)}_{\tilde{V}_{\ell}}}_{T_1} \norm[\big]{a}_{\tilde{V}^{-1}_{\ell}}.
\label{eqn:GLM_interim_1}
\end{equation}
We bound $T_1$ as follows.
\begin{equation}
\begin{aligned}
    T_1 &= \sqrt{ \left(X_{\ell} - \tilde{V}^{-1/2}_{\ell} h_{\ell}(\theta_*)\right)'  \tilde{V}^{1/2}_{\ell} G^{-1}_{\ell}  \tilde{V}_{\ell}  G^{-1}_{\ell} \tilde{V}^{1/2}_{\ell} \left(X_{\ell} - \tilde{V}^{-1/2}_{\ell} h_{\ell}(\theta_*)\right)} \\
    & \overset{(a)}\leq \frac{1}{\sqrt{k_1}} \sqrt{ \left(X_{\ell} - \tilde{V}^{-1/2}_{\ell} h_{\ell}(\theta_*)\right)'  \tilde{V}^{1/2}_{\ell} G^{-1}_{\ell}  \tilde{V}^{1/2}_{\ell} \left(X_{\ell} - \tilde{V}^{-1/2}_{\ell} h_{\ell}(\theta_*)\right)}\\ 
    & \overset{(b)}\leq \frac{1}{{k_1}} \sqrt{ \left(X_{\ell} - \tilde{V}^{-1/2}_{\ell} h_{\ell}(\theta_*)\right)' \left(X_{\ell} - \tilde{V}^{-1/2}_{\ell} h_{\ell}(\theta_*)\right)} \\
    & \overset{(c)} \leq \frac{C_1}{k_1}  \left( \sqrt{ \frac{d+\log(16/\delta_{\ell})}{M}} + \alpha \sqrt{\log{\frac{1}{\alpha}}} \right).
 \end{aligned}   
\label{eqn:GLM_interimbnd_2}
\end{equation}
For both (a) and (b) above, we used $k_1 \tilde{V}_{\ell} \preccurlyeq G_{\ell}$; for (c), we invoked the bound in Eq.~\eqref{eqn:X_l_bound}. Now recall from Eq.~\eqref{eqn:rc_bnd2} that 
$$   \norm[\big]{a}_{\tilde{V}^{-1}_{\ell}} \leq    \epsilon_{\ell} \sqrt{\frac{2M}{ \log\left(\frac{1}{\delta_{\ell}}\right)}}.$$
Combining the above bound with the ones in Eq.~\eqref{eqn:GLM_interim_1} and Eq.~\eqref{eqn:GLM_interimbnd_2}, and using $\log(1/\delta_{\ell}) \geq 1$,  we obtain
$$ \vert \langle \hat{\theta}^{(\ell)}-\theta_*, a \rangle \vert \leq \frac{\sqrt{2}C_1}{k_1} \left( \sqrt{ d+\frac{\log(16/\delta_{\ell})}{\log(1/\delta_{\ell})}} + \alpha \sqrt{M \log{\frac{1}{\alpha}}} \right) \epsilon_{\ell}.$$ 

Elementary calculations coupled with the fact that $\log(1/\delta_{\ell}) \geq 1$ yields:
$$ \frac{\log(16/\delta_{\ell})}{\log(1/\delta_{\ell})} \leq 4.$$ Putting all the pieces together, and simplifying, we arrive at the following bound:
\begin{equation}
    \vert \langle \hat{\theta}^{(\ell)}-\theta_*, a \rangle \vert \leq \frac{4 C_1}{k_1} \left( \sqrt{ d} + \alpha \sqrt{M \log{\frac{1}{\alpha}}} \right) \epsilon_{\ell}. 
\label{eqn:GLM_final}
\end{equation}
Using the fact that $\mu(\cdot)$ is $k_2$-Lipschitz then yields:
$$ \vert \mu(\langle \hat{\theta}^{(\ell)},a \rangle) - \mu(\langle \theta_*,a \rangle) \vert \leq k_2 \vert \langle \hat{\theta}^{(\ell)}-\theta_*, a \rangle \vert \leq 4 C_1\frac{ k_2}{k_1} \left( \sqrt{ d} + \alpha \sqrt{M \log{\frac{1}{\alpha}}} \right) \epsilon_{\ell},$$
which is the desired claim. This completes the proof.
\end{proof}

Equipped with the robust confidence bounds for \texttt{RC-GLM}, we can now complete the proof of Theorem \ref{thm:GLM}. 

\begin{proof} (\textbf{Proof of Theorem \ref{thm:GLM}}). Using essentially the same arguments as those used to prove Lemmas \ref{lemma:good_arm_RCPLB} and \ref{lemma:bad_arm_RCPLB}, we can prove that there exists a clean event, say $\mc{E}$, of measure at least $1-\delta$, such that on $\mc{E}$, the following hold: (i) $a_* \in \mc{A}_{\ell}, \forall \ell \in [L]$, where $L$ is the total number of epochs; and (ii) for any epoch $\ell \in [L], a \in \mc{A}_{\ell} \implies \tilde{\Delta}_a \leq 8\bar{\gamma}_{\ell}.$ Here, $\tilde{\Delta}_a = \mu(\langle \theta_*, a_*\rangle) - \mu(\langle \theta_*, a\rangle).$ Let us condition on this clean event $\mc{E}$. The remainder of the proof follows the same line of reasoning as that of Theorem \ref{thm:RCPLB}. For any good agent $i\in [M]\setminus\mc{B}$, we can bound the regret as follows. 

\begin{equation}
\begin{aligned}
\sum_{t=1}^{T} \left(\mu(\langle \theta_*, a_*\rangle) - \mu(\langle \theta_*, a_{i,t}\rangle) \right) 
&=\sum_{\ell=1}^{L}\sum_{a\in \textrm{Supp}(\pi_{\ell}) } \hspace{-2mm} m^{(\ell)}_a \left(\mu(\langle \theta_*, a_*\rangle) - \mu(\langle \theta_*, a\rangle) \right) \\
    &=\sum_{\ell=1}^{L}\sum_{a\in \textrm{Supp}(\pi_{\ell}) }\ceil*{\frac{T^{(\ell)}_a}{M}} \tilde{\Delta}_a \\
    &\leq  \underbrace{\sum_{\ell=1}^{L}\sum_{a\in \textrm{Supp}(\pi_{\ell}) } \hspace{-2mm} \frac{T^{(\ell)}_a}{M} \tilde{\Delta}_a}_{T_1}+\underbrace{\sum_{\ell=1}^{L}\sum_{a\in \textrm{Supp}(\pi_{\ell}) }\hspace{-2mm} \tilde{\Delta}_a}_{T_2}. 
\end{aligned}
\end{equation}
As in the proof of Theorem \ref{thm:RCPLB}, we bound $T_1$ and $T_2$ separately. For bounding $T_1$, we have:
\begin{equation}
\begin{aligned}
    T_1&=\sum_{\ell=1}^{L}\sum_{a\in \textrm{Supp}(\pi_{\ell}) }\hspace{-2mm} \frac{T^{(\ell)}_a}{M} \tilde{\Delta}_a \\ 
    &= \frac{d}{M} \sum_{\ell=1}^{L} \frac{1}{\epsilon^2_{\ell}} \log{\left( \frac{1}{\delta_{\ell}} \right)} \sum_{a\in \textrm{Supp}(\pi_{\ell}) } \hspace{-3mm} \pi_{\ell}(a) \tilde{\Delta}_a \\
   & \leq \frac{8d}{M} \sum_{\ell=1}^{L} \frac{1}{\epsilon^2_{\ell}} \log{\left( \frac{1}{\delta_{\ell}}\right)} \bar{\gamma}_{\ell} \\
   &= \frac{32C_1 d \left(\sqrt{d}+\alpha \sqrt{M\log(1/\alpha)}\right)}{M} \left(\frac{k_2}{k_1}\right)  \sum_{\ell=1}^{L} \frac{1}{\epsilon_{\ell}} \log{\left( \frac{10K^2 \ell^2}{\delta}  \right)} \\
   & \leq \frac{32C_1 d \left(\sqrt{d}+\alpha \sqrt{M\log(1/\alpha)}\right)}{M} \left(\frac{k_2}{k_1}\right)\log{\left( \frac{10K^2 L^2}{\delta}  \right)}\sum_{\ell=1}^{L} 2^{\ell}\\
   & = O\left(\left(\frac{k_2}{k_1}\right) \frac{d \left(\sqrt{d}+\alpha \sqrt{M\log(1/\alpha)}\right)}{M} \log{\left( \frac{10K^2 L^2}{\delta}  \right)} 2^L \right). 
\end{aligned}
\label{eqn:GLM_T1_bnd}
\end{equation}
Recall the following fact that we proved earlier for Theorem \ref{thm:RCPLB}: $$ 2^L \leq \sqrt{ \frac{MT}{d \log{\left( {10K^2 L^2}/{\delta}  \right)} }}.$$
Plugging this bound in Eq.~\eqref{eqn:GLM_T1_bnd}, we obtain:
$$ T_1 = O\left(\left(\frac{k_2}{k_1}\right) {\left(\alpha \sqrt{\log(1/\alpha)} + \sqrt{   \frac{d}{M}}\right)} \sqrt{\log{\left( \frac{KT}{\delta}  \right)} dT} \right).$$ 

One can upper-bound $T_2$ using the same bound as above using exactly the same steps as in the proof of Theorem \ref{thm:RCPLB}. Combining the bounds on $T_1$ and $T_2$ leads to the claim of Theorem \ref{thm:GLM}.
\end{proof}

$\bullet$ \textbf{Comments on solving for $\hat{\theta}^{(\ell)}$ in \texttt{RC-GLM}}. Recall that line 9 of Algorithm \texttt{RC-GLM} requires the server to solve for $\hat{\theta}^{(\ell)}$ based on the following equation:
\begin{equation}
 h_{\ell}(\hat{\theta}^{(\ell)}) = \tilde{V}^{1/2}_{\ell} X_{\ell}.
\label{eqn:injection}
\end{equation}

Based on Lemma \ref{lemma:MVT}, we have that for any $\theta_1, \theta_2 \in \Theta$ such that $\theta_1 \neq \theta_2$:
$$ \left(\theta_1 - \theta_2\right)' \left( h_{\ell}(\theta_1)-h_{\ell}(\theta_2)\right)=\left(\theta_1 - \theta_2\right)'G_{\ell}(\theta_1;\theta_2) \left(\theta_1 - \theta_2\right) > 0,$$
since $G_{\ell}(\theta_1;\theta_2)$ is positive-definite. Thus, the map $h_{\ell}: \mathbb{R}^{d} \rightarrow \mathbb{R}^{d}$ is injective, and $h_{\ell}^{-1}$ is well-defined. This, in turn, implies that Eq.~\eqref{eqn:injection} has a unique solution. 

\newpage
\subsection{The Iteratively Reweighted Mean Estimation Algorithm}
In this section, we briefly explain the main idea behind the Iteratively Reweighted Mean Estimation Algorithm in \cite{Dalal}. Suppose we are given an $\alpha$-corrupted set $\mc{X}$ of samples, namely, $M$ $d$-dimensional samples $x_1, \ldots, x_M$, such that $(1-\alpha)M$ of these samples are drawn i.i.d. from $\mc{N}(v, \Sigma)$, and the remaining $\alpha M$ samples are arbitrarily corrupted by an adversary. The goal is to recover the unknown mean vector $v \in \mathbb{R}^d$, given knowledge of the samples $\mc{X}$, the corruption fraction $\alpha$, and the covariance matrix $\Sigma.$ To see how this is done, let us define a couple of quantities for any pair of vectors $w \in [0,1]^d$ and $\mu\in \mathbb{R}^{d}$:
\begin{equation}
    \bar{x}_{w}=\sum_{i=1}^{M} w_i x_i; \hspace{2mm} G(w,\mu) = \lambda_{\max}\left(\sum_{i=1}^{M}w_i \left(x_i-\mu \right)\left(x_i-\mu \right)' - \Sigma\right).
\label{eqn:ITW}
\end{equation}

The basic idea is to find a weight vector $\hat{w}$ within the $d$-dimensional probability simplex such that the weighted average $\bar{x}_{\hat{w}}$ is close to the true mean $v$. Intuitively, a ``small" value of $G(\hat{w},\bar{x}_{\hat{w}})$ is an indicator of a good candidate for such a weight vector. This is essentially the strategy pursued in Algorithm \ref{alg:rob_est} where one iteratively updates the weight vectors, and the associated weighted averages, so as to minimize the function $G(\cdot, \cdot)$ defined in Eq.~\eqref{eqn:ITW}. At the termination of this algorithm, say after $N$ iterations, the goal is to output a weight vector $\hat{w}_N$ that mimics the ideal weight vector $w^*$ defined by: $w^*_j = \mathbf{1}(j\in \mc{I})/|\mc{I}|$, where $\mc{I}$ is the set of good samples (inliers). The steps of the Iteratively Reweighted Mean Estimator are outlined in Algorithm \ref{alg:rob_est}. The guarantees associated with this estimator are as stated in Lemma \ref{lemma:Dalal}. 

\begin{algorithm}[H]
\caption{\text{Iteratively Reweighted Mean Estimator} (\texttt{ITW})}
\label{alg:rob_est}
\begin{algorithmic}[1]
\State \textbf{Input:}  $\alpha$-corrupted set of $M$ samples $\mathcal{X}$, corruption fraction $\alpha$, and covariance matrix $\Sigma$. 
\State \textbf{Output:} Robust estimate of the mean $\hat{v}$.
\State \textbf{Initialize:} Compute  $\hat{v}_0$ as a minimizer of $\argmin\limits_{\mu} \sum\limits_{i=1}^{M}\|x_i-\mu\|$.
\State Let $N=0\vee \ceil*{\frac{\log(4 r_{\Sigma})-2 \log(\alpha(1-2\alpha))}{2 \log((1-2\alpha))-\log(\alpha)-\log(1-\alpha)}}.$ Here, $r_{\Sigma} = \textrm{Trace}(\Sigma)/\Vert \Sigma \Vert_2.$
\For{$k=1$ to $N$}
\State Compute current weights:
$$w \in \argmin_{(M-M\epsilon)\Vert w \Vert_{\infty} \leq 1}\lambda_{\max}\left(\sum_{i=1}^{M}w_i(x_i-\hat{v}_{k-1})(x_i-\hat{v}_{k-1})'-\Sigma\right)\vee 0.$$
\State Update the estimator:
$$\hat{v}_{k}=\sum_{i=1}^{M}w_ix_i.$$
\EndFor
\State \textbf{Return} $\hat{v}=\hat{v}_K$.
\end{algorithmic}%
\end{algorithm}

\newpage
\section{Analysis for the Contextual Bandit Setting: Proof of Theorem \ref{thm:contextual}}
\label{app:proof_contxt}
The proof of Theorem \ref{thm:contextual} proceeds in multiple steps. We start with an analysis of the Robust BaseLinUCB subroutine, namely Algorithm \ref{algo:Base}.

\begin{mylem}[label=lemma:context_bnd]{Bounds for Robust BaseLinUCB}{} 
Suppose the input index set $\Psi_t$ is constructed so that for fixed $x_{\tau,a_{\tau}}, \tau \in \Psi_t$, the rewards $\{r_{i,\tau}\}_{\tau\in \Psi_t}$ are independent random variables for each good agent $i\in [M]\setminus \mc{B}$. Then, for each $a\in[K]$, with probability at least $1-\bar{\delta}$, it holds that:
\begin{equation}
|\hat{r}_{t,a} - \langle \theta_*, x_{t,a} \rangle| \leq \left(\alpha + 2C \sqrt{ \frac{ \log{ \left( \frac{1}{\bar{\delta}} \right) } } {M} } \right) \Vert x_{t,a} \Vert_{A^{-1}_t},
\end{equation}
where $C$ is as in Lemma \ref{lemma:median}.
\end{mylem}
\begin{proof}
Fix any agent $i\in[M]\setminus\mc{B}$. Now from the expression for $\hat{\theta}_{i,t}$ in Algorithm \ref{algo:Base}, observe that:
\begin{equation}
    \begin{aligned}
    \hat{\theta}_{i,t} &= A^{-1}_t b_{i,t} \\
                       &= A^{-1}_t \left( \sum_{\tau \in \Psi_t} r_{i,\tau} x_{\tau,a_{\tau}} \right)\\
                       &= A^{-1}_t \left( \sum_{\tau \in \Psi_t} \left( \langle \theta_*, x_{\tau,a_{\tau}} \rangle +\eta_{i,\tau} \right) x_{\tau,a_{\tau}} \right)\\
                       &= 
                       A^{-1}_t \left( \sum_{\tau \in \Psi_t}  x_{\tau,a_{\tau}} x'_{\tau,a_{\tau}} \right) \theta_*+A^{-1}_t \left( \sum_{\tau \in \Psi_t}  \eta_{i,\tau}  x_{\tau,a_{\tau}} \right)\\
                       &=\left(I_d - \frac{A^{-1}_t}{M}\right)\theta_* + A^{-1}_t \left( \sum_{\tau \in \Psi_t}  \eta_{i,\tau}  x_{\tau,a_{\tau}} \right),
    \end{aligned}
\end{equation}
where we used $\eta_{i,\tau}$ as a shorthand for $\eta_{i,\tau}(a_{\tau})$. Now consider any $a\in [K]$. Using the above expression, we will now decompose the error in estimation of $\langle \theta_*, x_{t,a} \rangle$ into a bias term and a variance term:
\begin{equation}
    \langle \hat{\theta}_{i,t} - \theta_*, x_{t,a} \rangle = \underbrace{- \frac{1}{M} \langle A^{-1}_t \theta_*, x_{t,a} \rangle}_{ \textrm{bias term}} +  \underbrace{\sum_{\tau \in \Psi_t}   \langle A^{-1}_t x_{\tau,a_{\tau}}, x_{t,a} \rangle \eta_{i,\tau}}_{\textrm{variance term}}.
\label{eqn:bias_variance}
\end{equation}
For a fixed set of feature vectors, under our assumption that $\{r_{i,\tau}\}_{\tau\in \Psi_t}$ are independent random variables, the variance term is a sum of independent zero-mean Gaussian noise variables. Thus,  
$$ \mathbb{E}\left[ \sum_{\tau \in \Psi_t}   \langle A^{-1}_t x_{\tau,a_{\tau}}, x_{t,a} \rangle \eta_{i,\tau} \right]= \sum_{\tau \in \Psi_t}   \langle A^{-1}_t x_{\tau,a_{\tau}}, x_{t,a} \rangle \mathbb{E}\left[ \eta_{i,\tau} \right] = 0. $$
Furthermore, we have:
\begin{equation}
    \begin{aligned}
    \mathbb{E}\left[ \left(\sum_{\tau \in \Psi_t}   \langle A^{-1}_t x_{\tau,a_{\tau}}, x_{t,a} \rangle \eta_{i,\tau}\right)^2 \right] & = x'_{t,a} A^{-1}_t \left(\sum_{\tau \in \Psi_t}  \mathbb{E}\left[ \left( \eta^2_{i,\tau}   \right)  \right] x_{\tau,a_{\tau}} x'_{\tau,a_{\tau}}  \right)  A^{-1}_t x_{t,a}\\
    &= x'_{t,a} A^{-1}_t \left(\sum_{\tau \in \Psi_t}  x_{\tau,a_{\tau}} x'_{\tau,a_{\tau}}  \right)  A^{-1}_t x_{t,a} \\
    & \leq x'_{t,a} A^{-1}_t \underbrace{\left(\frac{I_d}{M} +  \sum_{\tau \in \Psi_t}  x_{\tau,a_{\tau}} x'_{\tau,a_{\tau}}  \right)}_{A_t}  A^{-1}_t x_{t,a}\\
    & = {\Vert x_{t,a} \Vert}^2_{A^{-1}_t}.
    \end{aligned}
\end{equation}
From the above arguments, we conclude that for each $i\in [M]\setminus\mc{B}$, 
$$ \langle \hat{\theta}_{i,t}, x_{t,a} \rangle \sim ~ \mc{N}\left( \langle \theta_* -  \frac{A^{-1}_t \theta_*}{M}, x_{t,a}\rangle, \sigma^2 \right), \hspace{2mm} \textrm{where} \hspace{2mm}  \sigma^2 \leq {\Vert x_{t,a} \Vert}^2_{A^{-1}_t}.$$
Since the noise samples are independent across agents, we also know that $\{ \langle \hat{\theta}_{i,t}, x_{t,a} \rangle \}_{i \in [M]\setminus \mc{B}}$ are independent. Recalling that $\hat{r}_{t,a} = \texttt{Median}\left( \{\langle \hat{\theta}_{i,t}, x_{t,a} \rangle, i \in [M]\} \right)$, and invoking Lemma \ref{lemma:median}, we then have that with probability at least $1-\bar{\delta}$, 
\begin{equation}
    \vert \hat{r}_{t,a} - \langle \theta_* -  \frac{A^{-1}_t \theta_*}{M}, x_{t,a}\rangle \vert \leq \left(\alpha + C \sqrt{ \frac{ \log{ \left( \frac{1}{\bar{\delta}} \right) } } {M} } \right) \Vert x_{t,a} \Vert_{A^{-1}_t}. 
\end{equation}
This immediately implies:
\begin{equation}
    \vert \hat{r}_{t,a} - \langle \theta_*, x_{t,a}\rangle \vert \leq \frac{1}{M} \vert \langle  A^{-1}_t \theta_*, x_{t,a}\rangle \vert + \left(\alpha + C \sqrt{ \frac{ \log{ \left( \frac{1}{\bar{\delta}} \right) } } {M} } \right) \Vert x_{t,a} \Vert_{A^{-1}_t}. 
\label{eqn:context_bnd1}
\end{equation}
It remains to bound the first term in the above display. To that end, we proceed as follows:
\begin{equation}
\begin{aligned}
    \frac{1}{M} \vert \langle  A^{-1}_t \theta_*, x_{t,a}\rangle \vert &= \frac{1}{M} \vert \langle  A^{-1}_t x_{t,a}, \theta_*\rangle \vert\\
    &\leq \frac{1}{M} \Vert A^{-1}_t x_{t,a} \Vert  \Vert \theta_* \Vert\\
    &\overset{(a)}\leq \frac{1}{\sqrt{M}} \sqrt{ x'_{t,a} A^{-1}_t \frac{I_d}{M} A^{-1}_t x_{t,a}} \\
    &\leq \frac{1}{\sqrt{M}} \sqrt{ x'_{t,a} A^{-1}_t \left(\frac{I_d}{M} +     \sum_{\tau \in \Psi_t}  x_{\tau,a_{\tau}} x'_{\tau,a_{\tau}}  \right) A^{-1}_t x_{t,a}} \\
    & = \frac{1}{\sqrt{M}} \Vert x_{t,a} \Vert_{A^{-1}_t} \\
    &\overset{(b)}\leq C \sqrt{ \frac{ \log{ \left( \frac{1}{\bar{\delta}} \right) } } {M} }  \Vert x_{t,a} \Vert_{A^{-1}_t},
\end{aligned}
\end{equation}
where for (a), we used the fact that $\Vert \theta_* \Vert \leq 1$ by assumption; and for (b), we used $C \geq 1, \log{ \left( \frac{1}{\bar{\delta}} \right) } \geq 1.$ Combining the above bound with the one in Eq.\eqref{eqn:context_bnd1} leads to the desired claim.
\end{proof}

For Lemma \ref{lemma:context_bnd} to hold, the crucial requirement is for the rewards corresponding to indices in $\Psi_t$ to be independent. Our next result shows that this is indeed the case. The proof of this lemma essentially follows the same arguments as that of \cite[Lemma 14]{auer2}; we reproduce these arguments here only for completeness. 
\begin{mylem}[label=ref:ind]{Independence of Samples}{} Fix any agent $i\in [M]\setminus\mc{B}.$ For each $s\in [S]$ and each $t\in [T]$, given any fixed sequence of feature vectors $\{x_{\tau,a_{\tau}}\}_{\tau \in \Psi^{(s)}_t}$, the rewards $\{r_{i,\tau}\}_{\tau \in \Psi^{(s)}_t}$ are independent random variables.
\end{mylem}
\begin{proof}
Let us start by observing that a time-step $t$ can be added to $\Psi^{(s)}_t$ only in line 6 of Algorithm \ref{algo:SupLin}. Thus, the event $\{t \in \Psi^{(s)}_t\}$ only depends on all prior phases $\cup_{\ell < s} \Psi^{(\ell)}_t$, and on the confidence width $w^{(s)}_{t,a}$. From the definition of $w^{(s)}_{t,a}$ in line 7 of Algorithm \ref{algo:Base}, we note that $w^{(s)}_{t,a}$ depends only on the feature vectors $x_{\tau, a_{\tau}}, \tau \in \Psi^{(s)}_t$, and $x_{t,a}$. Combining the above observations, it is easy to see that $\{t \in \Psi^{(s)}_t\}$ only depends on the feature vectors. Noting that the feature vector sequence is fixed, and cannot be controlled by the adversarial agents, leads to the claim of the lemma. 
\end{proof}

The next result tells us that with high-probability, at each time-step $t\in [T]$, (i) the best arm $a^*_t$ is retained in all stages of the screening process; and (ii) an active arm in phase $s$ can contribute to at most $8/(2^s \sqrt{M})$ instantaneous regret. 

\begin{mylem}[label=lemma:bounds_contex]{}{} With probability at least $1-\bar{\delta} K S T$, for any $t\in [T]$ and any  $s \in [S]$, the following hold:
\begin{enumerate}
    \item[(i)] $\vert \hat{r}^{(s)}_{t,a} - \langle \theta_*, x_{t,a} \rangle \vert \leq w^{(s)}_{t,a}, \forall a \in \mc{A}_s$.  
    \item[(ii)] $a^*_t \in \mc{A}_s.$
    \item[(iii)] $\langle \theta_*, x_{t,a^*_t} \rangle - \langle \theta_*, x_{t,a} \rangle \leq 8/(2^s \sqrt{M}), \forall a \in \mc{A}_s.$ 
\end{enumerate}
\end{mylem}
\begin{proof}
Part (i) of the result follows directly from Lemma \ref{lemma:context_bnd}, and an union bound over all time-steps, phases and arms. 

For part (ii), let us condition on the clean event, say $\mc{E}$, on which part (i) holds. From the rules of Algorithm \ref{algo:SupLin}, it holds trivially that $a^*_t \in \mc{A}_s$ for $s=1$. Now suppose there exists some phase $s > 1$ such that $a^*_t \in \mc{A}_{s-1}$, but $a^*_t \notin \mc{A}_s$. From line 8 in Algorithm \ref{algo:SupLin}, we must have $w^{(s-1)}_{t,a} \leq 2^{1-s}/\sqrt{M}, \forall a \in \mc{A}_{s-1}$. From the phased elimination strategy in line 8 of Algorithm \ref{algo:SupLin}, $a^*_t \notin \mc{A}_s$ implies the existence of some arm $a\in \mc{A}_s$ such that:
\begin{equation}
    \begin{aligned}
    & \left(\hat{r}^{(s-1)}_{t,a} + w^{(s-1)}_{t,a} \right) - \left(\hat{r}^{(s-1)}_{t,a^*_t} + w^{(s-1)}_{t,a^*_t} \right) > \frac{2^{2-s}}{\sqrt{M}} \\
    & \hspace{-7.5 mm} \overset{(a)}\implies \langle \theta_*, x_{t,a} \rangle + 2 w^{(s-1)}_{t,a} - \left(\hat{r}^{(s-1)}_{t,a^*_t} + w^{(s-1)}_{t,a^*_t} \right) > \frac{2^{2-s}}{\sqrt{M}} \\
    & \hspace{-7.5 mm} \overset{(b)}\implies \langle \theta_*, x_{t,a} - x_{t,a^*_t} \rangle + 2 w^{(s-1)}_{t,a} > \frac{2^{2-s}}{\sqrt{M}} \\
    & \hspace{-7.5 mm} \overset{(c)}\implies 2 w^{(s-1)}_{t,a} > \frac{2^{2-s}}{\sqrt{M}}, \\
    \end{aligned}
\end{equation}
which leads to a contradiction as $w^{(s-1)}_{t,a} \leq 2^{1-s}/\sqrt{M}$. In the above steps, both (a) and (b) follow from the defining property of the clean event $\mc{E}$; for (c), we used $\langle \theta_*, x_{t,a} - x_{t,a^*_t} \rangle \leq 0$ from the optimality of $a^*_t$. This completes the proof of part (ii). 

For part (iii), let us once again condition on the clean event $\mc{E}$ on which part (i) holds. Now $a\in \mc{A}_s \implies a \in \mc{A}_{s-1}$. We also know from part (ii) that $a^*_t \in \mc{A}_{s-1}$. The retention of arm $a$ in $\mc{A}_{s}$ implies (based on line 8 of Algorithm \ref{algo:SupLin}), 

\begin{equation}
    \begin{aligned}
    & \left(\hat{r}^{(s-1)}_{t,a^*_t} + w^{(s-1)}_{t,a^*_t} \right) - \left(\hat{r}^{(s-1)}_{t,a} + w^{(s-1)}_{t,a} \right) \leq \frac{2^{2-s}}{\sqrt{M}} \\
    & \hspace{-7.5 mm} \overset{(a)}\implies \langle \theta_*, x_{t,a^*_t} \rangle - \left(\hat{r}^{(s-1)}_{t,a} + w^{(s-1)}_{t,a} \right) \leq \frac{2^{2-s}}{\sqrt{M}} \\
    & \hspace{-7.5 mm} \overset{(b)}\implies \langle \theta_*, x_{t,a^*_t} - x_{t,a} \rangle \leq  \frac{2^{2-s}}{\sqrt{M}} + 2 w^{(s-1)}_{t,a} \\
    & \hspace{-7.5 mm} \overset{(c)}\implies \langle \theta_*, x_{t,a^*_t} - x_{t,a} \rangle \leq  \frac{8}{2^s\sqrt{M}}, \\
    \end{aligned}
\end{equation}
which is the desired claim. Here, for (a) and (b) we used the defining property of $\mc{E}$. As for (c), we used the fact that $w^{(s-1)}_{t,a} \leq 2^{1-s}/\sqrt{M}$.
\end{proof}

The final piece needed in the proof of Theorem \ref{thm:contextual} is a bound on $\vert \Psi^{(s)}_T \vert$ for each $s \in [S]$. To that end, we will make use of the elliptical potential lemma from \cite{abbasi}. 

\begin{mylem}[label=lemma:ellipt]{Elliptical Potential Lemma}{}
Let $\{X_t\}^{\infty}_{t=1}$ be a sequence in $\mathbb{R}^{d}$, $V$ a $d \times d$ positive definite matrix, and define $\bar{V}_t=V+\sum_{\tau=1}^{t} X_{\tau} X'_{\tau}$. If $\Vert X_t \Vert \leq L$ for all $t$, then we have that
$$ \sum_{t=1}^{T} \min\{1, {\Vert X_t \Vert}^2_{\bar{V}^{-1}_{t-1}}\} \leq 2 \log{ \frac{\det(\bar{V}_T)}{\det(V)}} \leq 2 (d \log{((\textrm{trace}(V)+TL^2)/d)} - \log(\det V)).
$$
\end{mylem}

We have the following result.

\begin{mylem}[label=lemma:cardinality]{Bound on $\vert \Psi^{(s)}_T\vert$}{}
Fix any $s \in [S]$. The following then holds:
\begin{equation}
    \vert \Psi^{(s)}_T \vert \leq 2^{s} \sqrt{M} \left(\alpha + 2C \sqrt{ \frac{ \log{ \left( \frac{1}{\bar{\delta}} \right) } } {M} } \right) \sqrt{2d |\Psi^{(s)}_T| \log(2M |\Psi^{(s)}_T|)}.
\end{equation}
\end{mylem}
\begin{proof}
Fix any phase $s\in [S]$. Now consider any time-step $t \in \Psi^{(s)}_T$. Since $t \in \Psi^{(s)}_T$, based on line 6 of Algorithm \ref{algo:SupLin}, it must be that:
\begin{equation}
    \frac{1}{2^s \sqrt{M}} < w^{(s)}_{t,a_t} = \left(\alpha + 2C \sqrt{ \frac{ \log{ \left( \frac{1}{\bar{\delta}} \right) } } {M} } \right) {\Vert x_{t,a_t} \Vert}_{A^{-1}_{s,t}}, \hspace{2mm} \textrm{where}
\label{eqn:ineq1}
\end{equation}
$$ A_{s,t} = \left(\frac{I_d}{M} +  \sum_{\tau \in \Psi^{(s)}_t}  x_{\tau,a_{\tau}} x'_{\tau,a_{\tau}}  \right). $$
Also, since $s \geq 1$, we have the following trivial inequality:
$$ \frac{1}{2^s \sqrt{M}} < \left(\alpha + 2C \sqrt{ \frac{ \log{ \left( \frac{1}{\bar{\delta}} \right) } } {M} } \right). $$
Combining the above inequality with the one in \eqref{eqn:ineq1}, we note that for each $t \in \Psi^{(s)}_T$, it holds that:
$$ \frac{1}{2^s \sqrt{M}} < \left(\alpha + 2C \sqrt{ \frac{ \log{ \left( \frac{1}{\bar{\delta}} \right) } } {M} } \right) \min\{1, {\Vert x_{t,a_t} \Vert}_{A^{-1}_{s,t}}\}. $$
Summing the above display over all indices in $\Psi^{(s)}_T$ yields:
\begin{equation}
    \begin{aligned}
 \frac{|\Psi^{(s)}_T|}{2^s \sqrt{M}} & < \left(\alpha + 2C \sqrt{ \frac{ \log{ \left( \frac{1}{\bar{\delta}} \right) } } {M} } \right) \sum_{t\in \Psi^{(s)}_T} \min\{1, {\Vert x_{t,a_t} \Vert}_{A^{-1}_{s,t}}\} \\
 & \overset{(a)} \leq \left(\alpha + 2C \sqrt{ \frac{ \log{ \left( \frac{1}{\bar{\delta}} \right) } } {M} } \right)\sqrt{ |\Psi^{(s)}_T| \sum_{t\in \Psi^{(s)}_T} \min\{1, {\Vert x_{t,a_t} \Vert}^2_{A^{-1}_{s,t}}\}} \\
 &\overset{(b)} \leq \left(\alpha + 2C \sqrt{ \frac{ \log{ \left( \frac{1}{\bar{\delta}} \right) } } {M} } \right)\sqrt{2 d |\Psi^{(s)}_T| \log{ \left(1+ \frac{M}{d} \vert \Psi^{(s)}_T \vert \right)}} \\
 & \leq \left(\alpha + 2C \sqrt{ \frac{ \log{ \left( \frac{1}{\bar{\delta}} \right) } } {M} } \right)\sqrt{2 d |\Psi^{(s)}_T| \log{ \left(2M\vert \Psi^{(s)}_T \vert \right)}},
 \end{aligned}
 \end{equation}
 where (a) follows from Jensen's inequality, and (b) follows from an application of Lemma \ref{lemma:ellipt}. Reorganizing the resulting inequality above leads to the desired claim. 
\end{proof}

We are now ready to prove Theorem \ref{thm:contextual}.

\begin{proof} (\textbf{Proof of Theorem \ref{thm:contextual}}) Since $S=\ceil*{\log T}$, we have $1/{(2^S \sqrt{M})} \leq 1/{(\sqrt{MT})}$. Thus, from the rules of Algorithm \ref{algo:SupLin}, it is apparent that at every time-step $t$, an action $a_t$ is \textit{always} chosen, either based on line 6, or on line 7. If we use $\Xi_T$ to store those time steps in $[T]$ where an action is chosen based on line 7 of Algorithm \ref{algo:SupLin}, then the above reasoning implies: $[T] = \Xi_T \cup \bigcup_{s\in [S]} \Psi^{(s)}_T$. 

Throughout the rest of the proof, we will  condition on the clean event on which items (i)-(iii) in Lemma \ref{lemma:bounds_contex} hold. We also recall that this clean event has measure at least $1-\bar{\delta} KST$. Now fix any good agent $i \in [M] \setminus \mc{B}$, and note that the same action $a_t$ is played by every good agent at time $t$. The cumulative regret for agent $i$ can thus be decomposed as follows:
\begin{equation}
    \sum_{t=1}^{T} \left( \langle \theta_*, x_{t,a^*_t} \rangle - \langle \theta_*, x_{t,a_{i,t}} \rangle \right) = \underbrace{\sum_{t\in \Xi_T}  \left( \langle \theta_*, x_{t,a^*_t} \rangle - \langle \theta_*, x_{t,a_{t}} \rangle \right)}_{T_1} + \underbrace{\sum_{s=1}^{S} \sum_{t\in \Psi^{(s)}_T} \left( \langle \theta_*, x_{t,a^*_t} \rangle - \langle \theta_*,x_{t,a_{t}} \rangle \right)}_{T_2}. 
\end{equation}
Let us first bound $T_2$ as follows.
\begin{equation}
    \begin{aligned}
    T_2 &= \sum_{s=1}^{S} \sum_{t\in \Psi^{(s)}_T} \left( \langle \theta_*, x_{t,a^*_t} \rangle - \langle \theta_*, x_{t,a_{t}} \rangle \right) \\
    & \overset{(a)} \leq \sum_{s=1}^{S} \sum_{t\in \Psi^{(s)}_T} \frac{8}{2^s \sqrt{M}}\\ 
    & = \sum_{s=1}^{S} \frac{8}{2^s \sqrt{M}} \vert \Psi^{(s)}_T \vert \\
    & \overset{(b)} \leq \sum_{s=1}^{S} 8 \left(\alpha + 2C \sqrt{ \frac{ \log{ \left( \frac{1}{\bar{\delta}} \right) } } {M} } \right) \sqrt{2d |\Psi^{(s)}_T| \log(2M |\Psi^{(s)}_T|)}\\
    & \overset{(c)} \leq 8 S \left(\alpha + 2C \sqrt{ \frac{ \log{ \left( \frac{1}{\bar{\delta}} \right) } } {M} } \right) \sqrt{2d T \log(2M T)} \\
    & \leq 8 (1+ \log(T)) \left(\alpha + 2C \sqrt{ \frac{ \log{ \left( \frac{1}{\bar{\delta}} \right) } } {M} } \right) \sqrt{2d T \log(2M T)},
    \end{aligned}
\end{equation}
where for (a), we used item (iii) of Lemma \ref{lemma:bounds_contex}; for (b), we invoked Lemma \ref{lemma:cardinality} to bound $|\Psi^{(s)}_T|$; and for (c), we used the trivial bound $|\Psi^{(s)}_T| \leq T$. With $\bar{\delta}=\delta/(KST)$, the bound on $T_2$ reads as follows:
\begin{equation}
    T_2 = O\left(\left(\alpha + 2C \sqrt{ \frac{ \log{ \left( \frac{KT}{{\delta}} \right) } } {M} } \right) \log(T) \sqrt{2d T \log(2M T)}  \right).
\label{eqn:T_2_bnd}
\end{equation}
Now let us turn to bounding $T_1$. Consider any time-step $t\in \Xi_T$ where the action $a_t$ is chosen based on line 7 of Algorithm \ref{algo:SupLin}. Since $a^*_t$ is never eliminated on the clean event (item (ii) of Lemma \ref{lemma:bounds_contex}), and since $a_t$ has the highest robust upper confidence bound among all active arms, it must be that:
\begin{equation}
\begin{aligned}
& \hat{r}^{(s)}_{t,a^*_t} + w^{(s)}_{t,a^*_t} \leq \hat{r}^{(s)}_{t,a_t} + w^{(s)}_{t,a_t} \\
& \hspace{-7.5mm} \implies \langle \theta_*, x_{t,a^*_t} - x_{t,a_t} \rangle \leq 2 w^{(s)}_{t,a_t}\\
& \hspace{-7.5mm} \implies \langle \theta_*, x_{t,a^*_t} - x_{t,a_t} \rangle \leq \frac{2}{\sqrt{MT}},
\end{aligned}
\end{equation}
where for the first implication, we used item (i) of Lemma \ref{lemma:bounds_contex}; and for the second, we used the fact for an action to be chosen based on line 7, it's confidence width must be bounded above by $1/(\sqrt{MT})$. We conclude that
$$ T_1 = \sum_{t\in \Xi_T}  \left( \langle \theta_*, x_{t,a^*_t} \rangle - \langle \theta_*, x_{t,a_t}  \rangle \right) = O\left( \sqrt{\frac{T}{M}}           \right). $$
\newpage
Combining the above bound on $T_1$ with that on $T_2$ in Eq.~\eqref{eqn:T_2_bnd}, we have that with probability at least $1-\delta$, the following is true for each good agent $i\in [M]\setminus \mc{B}$:
\begin{equation}
\begin{aligned}
\sum_{t=1}^{T} \left( \langle \theta_*, x_{t,a^*_t} \rangle - \langle \theta_*, x_{t,a_{t}} \rangle \right) &= O\left(\left(\alpha + 2C \sqrt{ \frac{ \log{ \left( \frac{KT}{{\delta}} \right) } } {M} } \right) \log(T) \sqrt{2d T \log(2M T)}  \right)\\
& = \tilde{O}\left(\left(\alpha + \sqrt{{1}/{M}} \right) \sqrt{dT} \right).
\end{aligned}
\end{equation}
This completes the proof.
\end{proof}
\newpage
\newpage 
\section{Alternate Strategies for Robust Collaborative Phased Elimination can lead to Sub-Optimal Regret Bounds.}
\label{app:alternate}
In Section \ref{sec:algo}, where we introduced \texttt{RCLB}, we alluded to the fact that certain natural candidate strategies may lead to sub-optimal regret bounds. In this section, we elaborate on this point. Note that our end goal is to come up with a phased elimination step akin to line 9 of \texttt{RCLB}. To achieve this, in every epoch $\ell$, we need estimates of $\langle \theta_*, a \rangle $ along with associated confidence intervals for each $a\in \mc{A}_{\ell}$. In what follows, we will consider two natural candidate strategies for the same, and demonstrate that they each lead to confidence bounds that are looser than the ones we derived in Lemma \ref{lemma:rob_conf_main}. As such, using such bounds in the phased elimination step will lead to regret guarantees that are sub-optimal in their dependence on the model dimension $d$. 

$\bullet$ \textbf{Candidate Strategy 1.} Suppose in every epoch $\ell$, the server collects the local model estimates $\{\hat{\theta}^{(\ell)}_i\}_{i\in [M]}$, and aims to first construct a robust estimate $\hat{\theta}^{(\ell)}$ of $\theta_*$. Subsequently, it uses $\langle \hat{\theta}^{(\ell)}, a \rangle$ as an estimate of $\langle \theta_*, a \rangle$ for each active arm $a\in \mc{A}_{\ell}$. To extract $\hat{\theta}^{(\ell)}$ from the local estimates $\{\hat{\theta}^{(\ell)}_i\}_{i\in [M]}$, we need a high-dimensional robust mean estimator. One natural candidate for this is the Iteratively Reweighted Mean Estimator from \cite{Dalal} since it leads to minimax-optimal error bounds. However, there is an immediate obstacle to directly applying the estimator from \cite{Dalal} on the model estimates $\{\hat{\theta}^{(\ell)}_i\}_{i\in [M]}$. This stems from the observation that although $\hat{\theta}^{(\ell)}_i$ is an unbiased estimate of $\theta_*$ for each good agent $i$, the covariance matrix associated with such an estimate may be \textit{ill-conditioned}. In particular, it is not hard to verify that for each $i\in [M]\setminus\mc{B}$:
$$ \mathbb{E}\left[ \left(\hat{\theta}^{(\ell)}_i - \theta_*\right) \left(\hat{\theta}^{(\ell)}_i - \theta_*\right)' \Big| \mc{F}_{\ell} \right] = \tilde{V}^{-1}_{\ell}. $$
Thus, if we were to construct $\hat{\theta}^{(\ell)}$ as 
$$\hat{\theta}^{(\ell)} =\texttt{ITW}( \{\hat{\theta}^{(\ell)}_i, \, i \in [M]\}),$$
then based on Lemma \ref{lemma:Dalal}, the error bound $\Vert \hat{\theta}^{(\ell)} - \theta_* \Vert$ would scale with $\Vert \tilde{V}^{-1}_{\ell} \Vert^{1/2}_2$.\footnote{Recall that we use $\texttt{ITW}(\cdot)$ to represent the output of the Iteratively Reweighted Mean Estimator from \cite{Dalal}, namely Algorithm \ref{alg:rob_est}.} This is undesirable as $\Vert \tilde{V}^{-1}_{\ell} \Vert^{1/2}_2$ can potentially take on a large value. To bypass this problem, we can use the same trick as we did for \texttt{RC-GLM}, and compute $\hat{\theta}^{(\ell)}$ as follows:
$$ \hat{\theta}^{(\ell)} = \tilde{V}^{-1/2}_{\ell} \left(\texttt{ITW}( \{ \tilde{V}^{1/2}_{\ell} \hat{\theta}^{(\ell)}_i, \, i \in [M]\}) \right). $$ 

The rationale behind the above approach is that the covariance matrix associated with $\tilde{V}^{1/2}_{\ell} \hat{\theta}^{(\ell)}_i, i \in [M] \setminus \mc{B}$, is $I_d$. Using Lemma \ref{lemma:Dalal}, and following similar arguments as used to arrive at Lemmas \ref{lemma:rob_conf_main} and \ref{lemma:GLM_conf_bnd}, one can show that for each $a\in \mc{A}_{\ell}$, with probability at least $1-\delta_{\ell}$, it holds that:
\begin{equation}
\vert \langle \hat{\theta}^{(\ell)},a \rangle - \langle \theta_*,a \rangle \vert = O\left( \left(\sqrt{d}+\alpha\sqrt{M \log(1/\alpha)} \right) \epsilon_{\ell}  \right).
\label{eqn:cand_sol_1}
\end{equation}
It is instructive to compare the above estimate on $\langle \theta_*,a \rangle$ with the estimate $\mu^{(\ell)}_a$ we used in \texttt{RCLB}. Specifically, recall from Lemma \ref{lemma:rob_conf_main} that for each $a\in \mc{A}_{\ell}$, with probability at least $1-\delta_{\ell}$, it holds that
$$ | \mu^{(\ell)}_a - \langle \theta_*,a \rangle | = O \left(\left(1+\alpha\sqrt{M} \right) \epsilon_{\ell} \right). $$ 

Comparing the above upper bound with the one in Eq.~\eqref{eqn:cand_sol_1}, we note that while the former is independent of the dimension $d$, the latter does exhibit a dependence via the $\sqrt{d}$ term. Now suppose we use the upper-bound from Eq.~\eqref{eqn:cand_sol_1} to construct a robust confidence threshold - say $\tilde{\gamma}_{\ell}$ - and use it to devise a phased elimination step as the one in line 9 of \texttt{RCLB}. Then, following the reasoning as that used to prove Theorem \ref{thm:RCPLB}, one can establish a per-agent regret bound of 
$$ \tilde{O}\left({\left(\alpha \sqrt{\log(1/\alpha)} + \sqrt{\frac{d}{M}}\right)} \sqrt{dT} \right), $$
which is unfortunately weaker than the guarantee we have in Theorem \ref{thm:RCPLB}.

$\bullet$ \textbf{Candidate Strategy 2.} The main idea is as follows. In every epoch $\ell$, the server queries each agent $i \in [M]$ to report their aggregate observation $r^{(\ell)}_{i,a}$ for each arm $a \in \textrm{Supp}(\pi_{\ell})$. Recall that $r^{(\ell)}_{i,a}$ is the average of the rewards for arm $a$ observed by agent $i$ during epoch $\ell$. The server next computes an aggregate ``clean" observation $\tilde{r}^{(\ell)}_{a}$ for each $a\in \textrm{Supp}(\pi_{\ell})$ as follows:
$$ \tilde{r}^{(\ell)}_{a} = \texttt{Median}\left(\{ r^{(\ell)}_{i,a}, i \in [M]\}       
\right). $$

It then uses these clean observations to compute an estimate $\hat{\theta}^{(\ell)}$ of $\theta_*$ as follows: 
$$ \hat{\theta}^{(\ell)} = \bar{V}^{-1}_{\ell} Y_{\ell}, \hspace{1.5mm} \textrm {where} \hspace{1.5mm} \bar{V}_{\ell} = \hspace{-3mm}  \sum_{a\in\textrm{Supp}(\pi_{\ell})} \hspace{-3mm} T^{(\ell)}_a aa' \hspace{1mm} ; \hspace{1.5mm} Y_{\ell} = \hspace{-5mm} \sum_{a\in\textrm{Supp}(\pi_{\ell})} \hspace{-3mm} T^{(\ell)}_a \tilde{r}^{(\ell)}_{a} a, $$
and 
$$T^{(\ell)}_a= \ceil*{\frac{\pi_{\ell}(a) d}{\epsilon^2_{\ell}} \log\left(\frac{1}{\delta_{\ell}}\right)}.$$

The quantity $\hat{\theta}^{(\ell)}$ obtained  above is now used to compute $\langle \hat{\theta}^{(\ell)}, a \rangle$ as an estimate  of the true mean payoff $\langle \theta_*, a \rangle$ of each arm $a \in \mc{A}_{\ell}$.\footnote{Note that the observations obtained from each agent $i$, namely $r^{(\ell)}_{i,a}, a \in \textrm{Supp}(\pi_{\ell})$, provide direct information about the mean payoffs of arms \textit{only} in $\textrm{Supp}(\pi_{\ell})$. However, for the phased elimination step, we need estimates of the mean payoffs of \textit{all arms in $\mc{A}_{\ell}$}, not just the ones in $\textrm{Supp}(\pi_{\ell}) \subseteq \mc{A}_{\ell}$. This is precisely why we need to go through an intermediate regression step to first compute an estimate of $\theta_*$.} As before, our goal is to bound $\vert \langle \hat{\theta}^{(\ell)}, a \rangle - \langle \theta_*, a \rangle \vert$ for each $a\in \mc{A}_{\ell}$. To that end, we start by noting that for each good agent $i$, $r^{(\ell)}_{i,a} \sim \mc{N}\left(\langle \theta_*, a \rangle, \frac{1}{m^{(\ell)}_a}\right).$ Invoking Lemma \ref{lemma:median} then tells us that with probability at least $1-\delta_{\ell}$, 
\begin{equation}
    \vert \tilde{r}^{(\ell)}_a - \langle \theta_*, a \rangle \vert \leq C \left(\alpha + \sqrt{ \frac{ \log(\frac{1}{\delta_{\ell}})}{M}} \right) \frac{1}{\sqrt{ m^{(\ell)}_a    } }  \leq C \left(\alpha \sqrt{M} + \sqrt{ { \log\left(\frac{1}{\delta_{\ell}}\right)}} \right) \frac{1}{\sqrt{ T^{(\ell)}_a    } }. 
\label{eqn:err_bnd_cand_2}
\end{equation}
For our subsequent discussion, let us condition on the event on which the above bound holds for every arm in $\mc{A}_{\ell}$. On this event, we can say that for each $a\in \mc{A}_{\ell}$, $\tilde{r}^{(\ell)}_a = \langle \theta_*, a \rangle + e^{(\ell)}_a$, where $e^{(\ell)}_a$ is an error term satisfying the bound in Eq.~\eqref{eqn:err_bnd_cand_2}. Now fix any $b\in \mc{A}_{\ell}$. Simple calculations reveal that: 
\begin{equation}
    \begin{aligned}
    \vert \langle \hat{\theta}^{(\ell)} - \theta_*, b \rangle \vert &= \Big|  \sum_{a\in\textrm{Supp}(\pi_{\ell})} \hspace{-3mm} T^{(\ell)}_a \langle \bar{V}^{-1}_{\ell}a, b \rangle e^{(\ell)}_a \Big| \\
    & \leq \sum_{a\in\textrm{Supp}(\pi_{\ell})} \hspace{-3mm} T^{(\ell)}_a \, \vert \langle \bar{V}^{-1}_{\ell}a, b \rangle \vert \vert e^{(\ell)}_a \vert \\
    & \overset{(a)} \leq C \left(\alpha \sqrt{M} + \sqrt{ { \log\left(\frac{1}{\delta_{\ell}}\right)}} \right) \underbrace{ \sum_{a\in\textrm{Supp}(\pi_{\ell})} \hspace{-3mm} \sqrt{T^{(\ell)}_a} \vert \langle \bar{V}^{-1}_{\ell}a, b \rangle \vert}_{T_1} \\
    & \overset{(b)} \leq C \left(\alpha \sqrt{M} + \sqrt{ { \log\left(\frac{1}{\delta_{\ell}}\right)}} \right) \sqrt{ \vert \textrm{Supp}(\pi_{\ell}) \vert \left(b' \bar{V}^{-1}_{\ell} \left(\sum_{a\in\textrm{Supp}(\pi_{\ell})} \hspace{-3mm} {T^{(\ell)}_a} a a' \right) \bar{V}^{-1}_{\ell} b \right)
 }\\
 & \leq C \left(\alpha \sqrt{M} + \sqrt{ { \log\left(\frac{1}{\delta_{\ell}}\right)}} \right) \sqrt{ \vert \textrm{Supp}(\pi_{\ell}) \vert } \, \Vert b \Vert_{\bar{V}^{-1}_{\ell}} \\
 & \overset{(c)} \leq C \left(\alpha \sqrt{M} + \sqrt{ { \log\left(\frac{1}{\delta_{\ell}}\right)}} \right) \sqrt{ 48d \log\log{d} } \, \Vert b \Vert_{\bar{V}^{-1}_{\ell}}\\
 & \overset{(d)}= \tilde{O}\left(\sqrt{d}\left(1+ \alpha\sqrt{M}\right) \epsilon_{\ell}\right).
    \end{aligned}
\label{eqn:cand_2_final_bnd}
\end{equation}

In the above steps, for (a) we used the bound from Eq.~\eqref{eqn:err_bnd_cand_2}; for (b), we used Jensen's inequality; for (c), we used the fact that $\vert \textrm{Supp}(\pi_{\ell}) \vert \leq 48d \log\log{d}$; and for (d), following a similar argument as in the proof of Lemma \ref{lemma:rob_conf_main}, we used that
$$ \Vert b \Vert_{\bar{V}^{-1}_{\ell}}  = O\left( \frac{\epsilon_{\ell}}{\sqrt{\log\left({1}/{\delta_{\ell}}\right)}}\right).$$

Comparing the bound in Eq.~\eqref{eqn:cand_2_final_bnd} with the one in Lemma \ref{lemma:rob_conf_main}, we once again note that while the latter bound is $d$-independent, the former has a clear dependence on $\sqrt{d}$. At the risk of sounding repetitive, if one were to employ the bound in Eq.~\eqref{eqn:cand_2_final_bnd} to construct a confidence threshold for phased elimination, and run through the same arguments as in the proof of Theorem \ref{thm:RCPLB}, one would end up with a per-agent regret bound of 
$$ \tilde{O}\left(\left(1+ \alpha\sqrt{M}\right)d\sqrt{T}\right).$$
Unlike the near-optimal guarantee we have in Theorem \ref{thm:RCPLB}, the above bound is clearly off by a factor of $\sqrt{d}$ from the optimal dependence on the model dimension $d$.
The looseness in the bound mainly stems from the following fact: the error terms $\{e^{(\ell)}_a\}_ {a \in \textrm{Supp}(\pi_{\ell})}$ are not necessarily sub-Gaussian random variables that are independent across arms. One can contrast this to the analysis in Lemma \ref{lemma:rob_conf_main}, where the noise terms $\{\bar{\eta}^{(\ell)}_{i,a}\}_{a \in \textrm{Supp}(\pi_{\ell})}$ were in fact Gaussian, and independent across arms. It is precisely the lack of nice statistical properties for the error terms $\{e^{(\ell)}_a\}_ {a \in \textrm{Supp}(\pi_{\ell})}$ that compels us to use Jensen's inequality to bound the term $T_1$ in Eq.~\eqref{eqn:cand_2_final_bnd}. At the moment, it is unclear to us whether one can come up with a tighter bound for this candidate strategy. 

\textbf{Main Takeaway.} The main message from this section is that deriving robust confidence intervals that lead to near-optimal bounds (such as the one in Theorem \ref{thm:RCPLB}) is non-trivial, and  requires a lot of care. In particular, the above discussion serves to highlight the significance of our algorithmic approach. 
\newpage
\end{document}